\documentclass[]{openmoss}

\usepackage{helvet}
\usepackage{booktabs}
\usepackage{multirow}
\usepackage{makecell}
\usepackage{array}
\usepackage{graphicx}
\usepackage{adjustbox}
\usepackage{threeparttable}
\usepackage[table]{xcolor}
\usepackage{ragged2e}

\usepackage{amsmath} 
\usepackage{natbib}
\usepackage{graphicx}
\usepackage{subcaption} 

\usepackage[toc,page,header]{appendix}
\usepackage[utf8]{inputenc} 
\usepackage[T1]{fontenc}    
\usepackage{hyperref}       
\usepackage{url}            
\usepackage{booktabs}       
\usepackage{amsfonts}       
\usepackage{nicefrac}       
\usepackage{microtype}      
\usepackage{wrapfig}
\usepackage{stfloats}

\usepackage{amssymb}        
\usepackage{titletoc}
\usepackage{minitoc}

\usepackage{array}
\usepackage{etoolbox}

\definecolor{lightblue}{RGB}{200, 230, 255}  
\definecolor{headerblue}{RGB}{150, 200, 255} 

\usepackage{pgfplots}
\usepackage{xcolor}
\usepackage{float}
\usepackage{comment}
\usepackage{multirow}
\usepackage{makecell}
\usepackage{siunitx}
\usepackage{tikz}
\usepackage{pgf-pie}

\usepackage{ragged2e}
\usepackage{tabularx}
\usepackage{caption}
\usepackage{enumitem}
\usepackage{pifont}
\usepackage[hang,flushmargin]{footmisc}

\usepackage{tcolorbox}
\tcbuselibrary{breakable}
\tcbuselibrary{skins}

\usepackage{tabularx}
\usepackage{listings}

\usepackage[edges]{forest}

\usepackage{threeparttable}
\usepackage{setspace}
\usepackage[scheme=plain]{ctex} 

\definecolor{MossCyan}{HTML}{82D9FF} 
\definecolor{MossBlue}{HTML}{82B1FF}

\definecolor{ForestGreen}{RGB}{34, 139, 34}
\definecolor{Red}{RGB}{255, 0, 0}

\definecolor{tickG}{rgb}{0.1, 0.588, 0.1}
\definecolor{crossR}{rgb}{0.588, 0.1, 0.1}

\definecolor{frenchblue}{rgb}{0.0, 0.45, 0.73}
\definecolor{babyblue}{rgb}{0.54, 0.81, 0.94}
\definecolor{classicrose}{rgb}{0.98, 0.8, 0.91}
\definecolor{beige}{rgb}{0.96, 0.96, 0.86}
\definecolor{forestgreen}{HTML}{2e7d43}

\definecolor{blue1}{HTML}{91BBE6}
\definecolor{blue2}{HTML}{3F90E0}
\definecolor{blue3}{HTML}{316FAD}

\definecolor{color1}{HTML}{FF9999}
\definecolor{color2}{HTML}{FF6666}
\definecolor{color3}{HTML}{FF3333}
\definecolor{color4}{HTML}{E60000}
\definecolor{color5}{HTML}{B30000}
\definecolor{color6}{HTML}{8CD98C}
\definecolor{color7}{HTML}{53c653}
\definecolor{color8}{HTML}{00B050}
\definecolor{color9}{HTML}{2d862d}
\definecolor{color10}{HTML}{206020}
\definecolor{color11}{HTML}{cca300}

\addto\extrasenglish{

}

\newtcolorbox{promptbox}[2][]{
    colback=white,
    coltext=black,
    arc=3mm,
    boxrule=0.5pt,
    colframe=black!60!white,
    title={#2},
    colbacktitle=black,
    coltitle=white,
    fonttitle=\bfseries,
    top=8pt,
    bottom=8pt,
    left=10pt,
    right=10pt,
    breakable,
    before upper={%
        \linespread{1}\selectfont
        \setlength{\parskip}{1ex plus 0.2ex minus 0.2ex}%
        \setlength{\parindent}{0pt}%
    },
    #1
}

\title{World Action Models: The Next Frontier in Embodied AI
}

\author{
Siyin Wang$^{1,2,*,\ddagger}$ \hspace{.3em}
Junhao Shi$^{1,2,*}$\hspace{.3em}
Zhaoyang Fu$^{1,*}$ \hspace{.1em}
Xinzhe He$^{1,*}$ \hspace{.1em}
Feihong Liu$^{1,*}$ \hspace{.1em}
\\
\textbf{
Chenchen Yang$^{1,2}$ \hspace{.1em}
Yikang Zhou$^{2}$ \hspace{.1em}
Zhaoye Fei$^{1}$ \hspace{.1em}
Jingjing Gong$^{2,}$ \hspace{.1em}
Jinlan Fu$^{1,}$ \hspace{.1em}
}
\\
\textbf{
Mike Zheng Shou$^{3}$ \hspace{.2em}
Xuanjing Huang$^{1,2}$ \hspace{.2em}
Xipeng Qiu$^{1,2}$\hspace{.2em}
Yu-Gang Jiang$^{1,\dagger}$ 
}
\\[1ex]
\texttt{
$^{1}$Fudan University   
$^{2}$Shanghai Innovation Institute   
$^{3}$National University of Singapore  
}
\\
{\fontsize{10}{12}\selectfont \texttt{$^*$Equal Contribution  $^{\ddagger}$Project Lead $^{\dagger}$Corresponding Author}}\\
[1ex]
}

\abstract{
Vision-Language-Action (VLA) models have achieved strong semantic generalization for embodied policy learning, yet they learn reactive observation-to-action mappings without explicitly modeling how the physical world evolves under intervention. A growing body of work addresses this limitation by integrating world models, predictive models of environment dynamics, into the action generation pipeline. We term this emerging paradigm \textbf{World Action Models (WAMs)}: embodied foundation models that unify predictive state modeling with action generation, targeting a joint distribution over future states and actions rather than actions alone. 
However, the literature remains fragmented across architectures, learning objectives, and application scenarios, lacking a unified conceptual framework. 
We formally define WAMs and disambiguate them from related concepts, and trace the foundations and early integration of VLA and world model research that gave rise to this paradigm. 
We organize existing methods into a structured taxonomy of Cascaded and Joint WAMs, with further subdivision by generation modality, conditioning mechanism, and action decoding strategy. 
We systematically analyze the data ecosystem fueling WAMs development, spanning robot teleoperation, portable human demonstrations, simulation, and internet-scale egocentric video, and synthesize emerging evaluation protocols organized around visual fidelity, physical commonsense, and action plausibility.
Overall, this survey provides the \textit{first} systematic account of the WAMs landscape, clarifies key architectural paradigms and their trade-offs, and identifies open challenges and future opportunities for this rapidly evolving field. 
}

\checkdata[Homepage]{\url{https://openmoss.github.io/Awesome-WAM}}
\checkdata[Github Repo]{\url{https://github.com/OpenMOSS/Awesome-WAM}}

\begin{document}
\maketitle

\begin{figure}[t]
    \centering
    \includegraphics[width=1\linewidth]{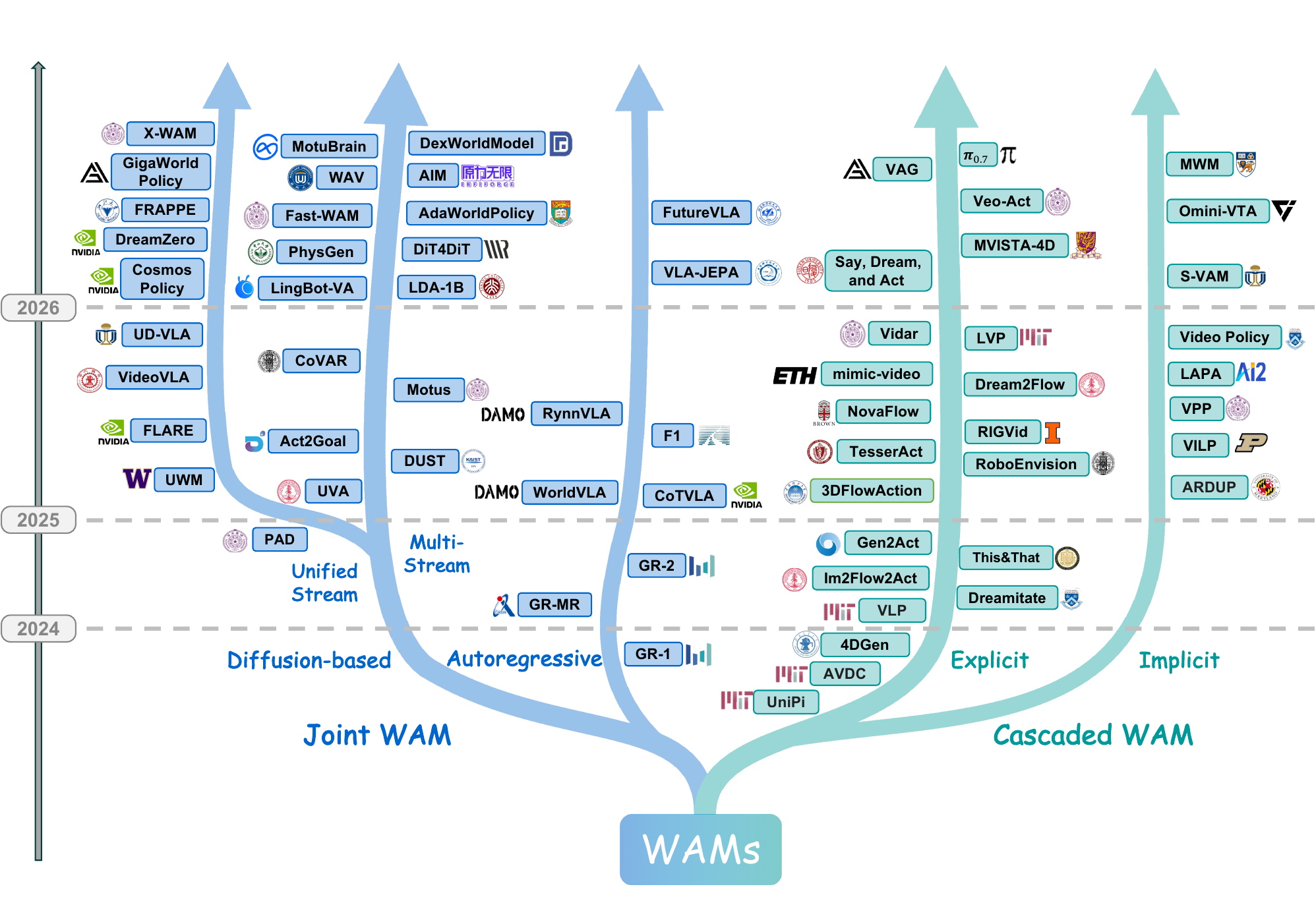}
    \caption{Temporal evolution and taxonomy of representative works on World Action Models (WAMs). The left branch illustrates the progression of \textbf{Joint WAM} architectures, which tightly couple world prediction and action generation, showing a divergence into \textit{Autoregressive} and \textit{Diffusion-based} representation schemes, with the continuous approach further bifurcating into Unified Stream and Multi-Stream backbones. The right branch summarizes the development of \textbf{Cascaded WAM} pipelines, where world modeling and action execution are primarily decoupled, evolving along \textit{Explicit} and \textit{Implicit} representation alignment trajectories. These structural strategies represent the field's dominant exploratory directions in architectural coupling rather than strictly sequential replacements.}
    \label{fig:Treemap}
\end{figure}

\section{Introduction}
\label{sec:intro}

The goal of building robots that perceive, reason, and act in unstructured physical environments has long driven embodied AI research. In recent years, the field has converged on a powerful paradigm: Vision-Language-Action (VLA) models that repurpose pretrained vision-language backbones as generalist robot policies. By formulating action generation as conditional token prediction atop internet-scale visual and linguistic representations, VLA models such as RT-2 \cite{RT-2}, OpenVLA \cite{OpenVLA}, and $\pi_0$ \cite{pi0} have demonstrated striking generalization—following novel language instructions, manipulating previously unseen objects, and transferring across robot embodiments with minimal finetuning. These results establish that the semantic understanding accumulated during large-scale vision-language pretraining can be effectively grounded into motor behavior, marking a qualitative advance over earlier task-specific controllers. The breadth of subsequent work building upon this formulation has consolidated VLA models as the dominant paradigm for generalist embodied policy learning.

However, standard VLA models do not explicitly model world dynamics—they learn direct observation-to-action mappings without predicting how the environment changes under intervention\cite{fei2025liberoplus}. This absence of predictive physical reasoning limits their generalization, where anticipating future states is essential.
Equipping embodied policy models with world modeling capabilities thus emerges as a natural direction \cite{D2PO}.
A growing body of recent work has begun integrating world models into the embodied policy pipeline. These approaches leverage predictive models of environment dynamics to provide agents with physical foresight—whether through video prediction as visual planning \cite{unipi,du2023vlp,ko2024avdc,yang2025roboenvision,gu2026saydreamact}, latent dynamics modeling for policy conditioning \cite{hu2024vpp,pai2025mimicvideo,liang2025videopolicy,yan2026svam,LAPA}, or joint state-action generation within unified architectures \cite{CosmosPolicy,DreamZero,LingBotVA,Motus,UWM,PAD,PhysGen}. This emerging direction has rapidly gained momentum, producing a diverse and expanding landscape of methods.

We formalize this family of approaches under the term \textbf{World Action Models (WAMs)}: embodied foundation models that unify predictive state modeling with action generation, targeting a joint distribution $p(o', a \mid o, l)$ over future states and actions rather than actions alone. Existing WAM methods can be broadly organized into two architectural categories: (1) \textit{Cascaded WAM} explicitly factorizes the objective, formally $p(o', a \mid o, l) = p(a \mid o', o, l)p(o' \mid o, l)$, by first synthesizing representations of anticipated future states, from which actions are subsequently derived; and (2) \textit{Joint WAM} directly modeling the joint distribution ($p(o', a \mid o, l)$)
, where state prediction and action generation are co-optimized within a shared representational space (see \autoref{fig:Treemap} for the temporal evolution of these architectures). The integration of world modeling brings stronger physical understanding, improved generalization across novel environments, and the ability to leverage large-scale human video data that lack action annotations—substantially expanding the data foundations available for embodied policy learning.

This survey provides the \textbf{first} systematic and critical analysis of the World Action Model landscape. Our aim is to offer both a conceptual framework for understanding the design space and a practical guide for researchers entering this rapidly evolving field. The organization of this survey (as illustrated in \autoref{fig:survey-sturcture}) is as follows:

\begin{itemize}
    \item \textbf{Definition (\autoref{sec:def}).} We provide a formal definition of World Action Models and disambiguate them from related concepts, including Video Policies, Action-Conditioned World Models, and standard VLA models, clarifying the terminological boundaries that currently fragment the literature.
    
    \item \textbf{Background (\autoref{sec:background}).} We trace the intertwined development of world modeling and action generation from classical model-based reinforcement learning through modern foundation model approaches, situating WAMs within a broader intellectual lineage.
    
    \item \textbf{Architecture (\autoref{sec:arch}).} We categorize existing WAM methods into Cascaded and Joint paradigms, with further subdivision by generation modality, conditioning mechanism, and action decoding strategy, providing a unified framework for comparing approaches across the design space.
    
    \item \textbf{Training Datasets (\autoref{sec:data}).} We analyze the four major data sources fueling WAM development—robot teleoperation, portable human demonstrations, simulation, and internet-scale egocentric video—examining how each source's characteristics shape the capabilities of trained models.
    
    \item \textbf{Evaluation (\autoref{sec:eval}).} We synthesize the emerging evaluation landscape, organized around visual fidelity, physical commonsense, and action plausibility, and identify gaps where current protocols fall short of assessing WAM-relevant capabilities.

    \item \textbf{Open Challenges (\autoref{sec:open}).} We conclude with a discussion of the critical hurdles and future directions for the field, highlighting the path toward more robust and generalizable World Action Models.
\end{itemize}

\definecolor{hidden-draw}{RGB}{0,0,0}
\newcommand{\papercite}[2]{#1~\cite{#2}}
\definecolor{coral}{RGB}{255, 127, 80}      
\definecolor{sage}{RGB}{128, 177, 133}      
\definecolor{azure}{RGB}{91, 155, 213}      
\definecolor{lilac}{RGB}{180, 160, 207}     
\definecolor{tealgray}{RGB}{102, 153, 153}   

\tikzstyle{leaf}=[draw=hiddendraw,
    rounded corners,minimum height=1.2em,
    fill=hidden-orange!40,text opacity=1,    align=center,
    fill opacity=.5,  text=black,align=left,font=\scriptsize,
inner xsep=3pt,
inner ysep=1pt,
]

\begin{figure}[thp]
\vspace{-3em}
\centering
\begin{forest}
    forked edges,
    for tree={
        grow=east,
        reversed=true,
        anchor=base west,
        parent anchor=east,
        child anchor=west,
        base=left,
        font=\footnotesize,
        rectangle,
        draw,
        rounded corners,align=left,
        minimum width=2.5em,
        minimum height=1.2em,
        edge+={darkgray, line width=0.6pt},
        s sep=6pt,
        l sep=10pt,
        inner xsep=3pt,
        inner ysep=1pt,
        ver/.style={rotate=90, child anchor=north, parent anchor=south, anchor=center},
    },
    where level=1{font=\scriptsize}{},
    where level=2{font=\scriptsize}{},
    where level=3{font=\scriptsize}{},
    where level=4{font=\scriptsize}{},
    where level=5{font=\scriptsize}{}, 
    [Roadmap to WAM, draw=gray, color=gray!100, fill=gray!15, thick, text=black, ver
        [Background, color=sage, fill=sage!15, thick, text=black
            [VLAs, color=sage, fill=sage!15
            ]
            [World Model, color=sage, fill=sage!15
                [Action-Conditioned, color=sage, fill=sage!15
                    [
                        \papercite{iVideoGPT}{iVideoGPT}{,}
                        \papercite{FlowDreamer}{FlowDreamer}{,}
                        \papercite{EnerVerse}{EnerVerse}{,}
                        \papercite{PlaNet}{PlaNet}{,} \\
                        \papercite{TransDreamer}{TransDreamer}{,}
                        \papercite{V-JEPA}{V-JEPA}{$\ldots$}
                    ]
                ]
                [Langugae-Conditoned, color=sage, fill=sage!15
                    [
                        \papercite{MoCoGAN}{MoCoGAN}{,}
                        \papercite{U-Net}{UNet}{,}
                        \papercite{Latte}{Latte}{,}
                        \papercite{Wan}{Wan}{,}
                        \papercite{Sora 2}{Sora_2}{$\ldots$}
                    ]
                ]
                [Embodied World Model, color=sage, fill=sage!15
                    [
                        \papercite{SWIM}{SWIM}{,}
                        \papercite{DreamDojo}{DreamDojo}{,}
                        \papercite{RoboDreamer}{RoboDreamer}{,}
                        \papercite{RoboScape}{RoboScape}{$\ldots$}
                    ]
                ]
            ]
            [WM for VLA, color=sage, fill=sage!15
                [Imitation Learning, color=sage, fill=sage!15
                    [
                        \papercite{Ctrl-World}{Ctrl-World}{,}
                        \papercite{RoboScape}{RoboScape}{,}
                        \papercite{DREMA}{DREMA}{}
                    ]
                ]
                [Reinforcement Learning, color=sage, fill=sage!15
                    [
                        \papercite{Dreamer to Control}{Dreamer2}{}
                        \papercite{DreamerV2}{Dreamer3}{,}
                        \papercite{Dreamer 4}{Dreamer4}{,}
                        \papercite{RISE}{RISE}{} \\
                        \papercite{DreamerV3}{DreamerV3}{,}
                        \papercite{DayDreamer}{DayDreamer}{,}
                        \papercite{World-Env}{World-Env}{,} 
                        \papercite{RoboScape-R}{RoboScape-R}{}\\
                        \papercite{WMPO}{WMPO}{,}
                        \papercite{WoVR}{WoVR}{,}
                        \papercite{VLA-RFT}{VLA-RFT}{,}
                        \papercite{RWML}{RWML}{,}
                        \papercite{MoDem-V2}{MoDem-V2}{} \\
                        \papercite{World-Gymnast}{World-Gymnast}{,}
                        \papercite{RWM-U}{Uncertainty-Aware}{,}
                        \papercite{World4RL}{World4RL}{,}
                        \papercite{VIPER}{VIPER}{}\\
                        \papercite{PhysWorld}{Robot_Learning_from_a_Physical_World_Model}{,}
                        \papercite{Diffusion Reward}{Diffusion_Reward}{,}
                        \papercite{GenReward}{GenReward}{} 
                    ]
                ]
                [Evaluation, color=sage, fill=sage!15
                    [
                        \papercite{Ctrl-World}{Ctrl-World}{,}
                        \papercite{Veo Robotics}{geminiroboticsteam}{,}
                        \papercite{Interactive World Simulator}{InteractiveWorldSimulator}{}\\
                        \papercite{WorldEval}{WorldEval}{,}
                        \papercite{WorldGym}{WorldGym}{,}
                        \papercite{dWorldEval}{li2026dworldevalscalableroboticpolicy}{}
                    ]
                ]
            ]
        ]
        [Architecture, color=azure, fill=azure!15, thick, text=black
            [Cascaded WAM, color=azure, fill=azure!15
                [Explicit, color=azure, fill=azure!15
                [
                        \papercite{UniPi}{unipi}{,}
                        \papercite{VLP}{du2023vlp}{,}
                        \papercite{RoboEnvision}{yang2025roboenvision}{,}
                        \papercite{ThisThat}{wang2024thisthat}{,}
                        \papercite{TesserAct}{TesserAct1}{,} 
                        \papercite{MVISTA-4D}{wang2026mvista4d}{}\\
                        \papercite{Say,Dream,and Act}{gu2026saydreamact}{,}
                        \papercite{Gen2Act}{bharadhwaj2024gen2act}{,}
                        \papercite{AVDC}{ko2024avdc}{,}
                        \papercite{Im2Flow2Act}{xu2024im2flow2act}{,}
                        \papercite{3DFlowAction}{zhi2025threedflowaction}{}\\
                        \papercite{NovaFlow}{li2025novaflow}{,}
                        \papercite{Dream2Flow}{dharmarajan2025dream2flow}{,}
                        \papercite{Dreamitate}{liang2024dreamitate}{,}
                        \papercite{4DGen}{liu2025fourdgen}{,}
                        \papercite{RIGVid}{patel2025rigvid}{,}
                        \papercite{LVP}{chen2025lvp}{}\\
                        \papercite{Vidar}{vidar}{,}
                        \papercite{Veo-Act}{zhang2026veoact}{,}
                        \papercite{pi0.7}{pi0.7}{,}
                        \papercite{VAG}{lang2026vag}{}
                    ]
                ]
                [Implicit, color=azure, fill=azure!15
                    [
                        \papercite{VPP}{hu2024vpp}{,}
                        \papercite{VILP}{xu2025vilp}{,}
                        \papercite{Video Policy}{liang2025videopolicy}{,}
                        \papercite{ARDuP}{huang2024ardup}{,}
                        \papercite{mimic-video}{pai2025mimicvideo}{,}\\
                        \papercite{LAPA}{LAPA}{,}
                        \papercite{villa-X}{chen2025villax}{,}
                        \papercite{S-VAM}{yan2026svam}{,}
                        \papercite{OmniVTA}{zheng2026omnivta}{,}
                        \papercite{MWM}{lou2026mwm}{}
                    ]         
                ]
            ]
            [Joint WAM, color=azure, fill=azure!15
                [Autoregression, color=azure, fill=azure!15
                    [
                        \papercite{GR1}{GR1}{,}
                        \papercite{grmg}{grmg}{,}
                        \papercite{GR2}{GR2}{,}
                        \papercite{CoTVLA}{CoTVLA}{,}
                        \papercite{WorldVLA}{WorldVLA}{,}
                        \papercite{rynnvla2}{rynnvla2}{}\\
                        \papercite{VLA-JEPA}{VLA-JEPA}{,}
                        \papercite{F1-VLA}{f1vla}
                    ]
                ]
                [Diffusion-based, color=azure, fill=azure!15
                    [
                        \papercite{PAD}{PAD}{,}
                        \papercite{VideoVLA}{VideoVLA}{,}
                        \papercite{UWM}{UWM}{,}
                        \papercite{DreamZero}{DreamZero}{,}
                        \papercite{CosmosPolicy}{CosmosPolicy}{,} 
                        \papercite{FLARE}{FLARE}{,}
                        \papercite{UVA}{UVA}{}\\
                        \papercite{FRAPPE}{FRAPPE}{,}
                        \papercite{CoVAR}{CoVAR}{,}
                        \papercite{LDA1B}{LDA1B}{,}
                        \papercite{WAV}{WAV}{,}
                        \papercite{DUST}{DUST}{,} 
                        \papercite{LingBotVA}{LingBotVA}{,}
                        \papercite{AIM}{AIM}{}\\
                        \papercite{DexWorldModel}{DexWorldModel}{,}
                        \papercite{FastWAM}{FastWAM}{,}
                        \papercite{MotuBrain}{MotuBrain}{} 
                        \papercite{AdaWorldPolicy}{AdaWorldPolicy}{,}
                        \papercite{DiT4DiT}{DiT4DiT}{,}\\
                        \papercite{Motus}{Motus}{,}
                        \papercite{Act2Goal}{Act2Goal}{,}                
                        \papercite{PhysGen}{PhysGen}{,}
                        \papercite{GigaWorld-Policy}{gigaworld-policy}{,}
                        \papercite{UD-VLA}{udvla1}{,}
                        \papercite{X-WAM}{guo2026xwam}{}
                    ]
                ]
            ]
        ]
        [Training data, color=lilac, fill=lilac!15, thick, text=black
            [Robot-centric\\ Teleoperation, color=lilac, fill=lilac!15
                [
                    \papercite{QT-Opt}{QT-Opt}{,}
                    \papercite{MIME}{MIME}{,}
                    \papercite{RoboNet}{RoboNet}{,}
                    \papercite{RoboTurk-Real}{RoboTurk}{,}
                    \papercite{BridgeData}{Bridge}{,}
                    \papercite{MT-Opt}{MT-Opt}{} \\
                    \papercite{BC-Z}{BC-Z}{,}
                    \papercite{RT-1}{RT-1}{,}
                    \papercite{Language-Table}{Language-Table}{,}
                    \papercite{BridgeData v2}{BridgeData_V2}{,}
                    \papercite{Jaco Play}{Jaco-Play}{} \\
                    \papercite{Cable Routing Dataset}{Cable-Routing-Dataset}{,}
                    \papercite{RH20T}{RH20T}{,}
                    \papercite{OXE}{OXE}{,}
                    \papercite{DROID}{DROID}{,}
                    \papercite{RH20T-P}{RH20T-P}{,}
                    \papercite{RoboMIND}{RoboMIND}{} \\
                    \papercite{ARIO}{ARIO}{,}
                    \papercite{RoboData}{RoboData}{,}
                    \papercite{DexCap}{DexCap}{,}
                    \papercite{FuSe}{FuSe}{,}
                    \papercite{AgiBot World}{AgiBot_World_Colosseo}{,}
                    \papercite{REASSEMBLE}{REASSEMBLE}{} \\
                    \papercite{OmniAction}{OmniAction}{,}
                    \papercite{UnifoLM-WBT}{UnifoLM-WBT}{}
                ]
            ]
            [UMI-style Human \\Demonstration, color=lilac, fill=lilac!15
                [
                    \papercite{UMI}{UMI}{,}
                    \papercite{FastUMI}{FastUMI}{,}
                    \papercite{FastUMI-100K}{FastUMI-100K}{,}
                    \papercite{RealOmin}{RealOmin}{,}
                    \papercite{Hoi!}{Hoi!}{,}
                    \papercite{RDT2}{RDT2}{} \\
                    \papercite{ActiveUMI}{ActiveUMI}{}{,}
                    \papercite{exUMI}{exUMI}{,}
                    \papercite{Tactile-Conditioned Diffusion Policy}{Tactile-Conditioned_Diffusion_Policy}{,}
                    \papercite{DexUMI}{DexUMI}{}\\
                    \papercite{UMI on Legs}{UMI_on_Legs}{,}
                    \papercite{HoMMI}{HoMMI}{,}
                    \papercite{MV-UMI}{MV-UMI}{}
                ]
            ]
            [Simulation Data, color=lilac, fill=lilac!15
                [
                    \papercite{MimicGen}{MimicGen}{,}
                    \papercite{ManiSkill2}{ManiSkill2}{,}
                    \papercite{RoboCasa}{RoboCasa}{,}
                    \papercite{RoboTwin}{RoboTwin}{,}
                    \papercite{DexMimicGen}{DexMimicGen}{} \\
                    \papercite{TesserAct}{TesserAct1}{,}
                    \papercite{RoboCerebra}{RoboCerebra}{,}
                    \papercite{SynGrasp-1B}{SynGrasp-1B}{,}
                    \papercite{RoboTwin 2.0}{RoboTWIN_2}{,}
                    \papercite{TLA Dataset}{TLA}{}\\
                    \papercite{InternData-M1}{InternVLA-M1}{,}
                    \papercite{InternData-A1}{InternData-A1}{,}
                    \papercite{QUARD-Auto}{QUARD-Auto}{}
                ]
            ]
            [Human Data, color=lilac, fill=lilac!15
                [
                    \papercite{SSv2}{SSv2}{,}
                    \papercite{EPIC-KITCHENS}{EPIC-KITCHENS}{,}
                    \papercite{HowTo100M}{HowTo100M}{,}
                    \papercite{Kinetics-700}{Kinetics-700}{,}
                    \papercite{EGTEA Gaze+}{EGTEA_Gaze+}{}\\
                    \papercite{Ego4D}{Ego4D}{,}
                    \papercite{HOI4D}{HOI4D}{,}
                    \papercite{EgoVid-5M}{EgoVid-5M}{,}
                    \papercite{COM Kitchens}{COM_Kitchens}{,}
                    \papercite{Egocentric-10k}{Egocentric-10k}{,}
                    \papercite{DreamDojo}{DreamDojo}{}\\
                    \papercite{Assembly101}{Assembly101}{,}
                    \papercite{H2O}{H2O}{,}
                    \papercite{EgoPAT3D}{EgoPAT3D}{,}
                    \papercite{Ego-Exo4D}{Ego-Exo4D}{,}
                    \papercite{ARCTIC}{ARCTIC}{,}
                    \papercite{HoloAssist}{HoloAssist}{}\\
                    \papercite{HOT3D}{HOT3D}{,}
                    \papercite{TACO}{TACO}{,}
                    \papercite{Kaiwu}{Kaiwu}{,}
                    \papercite{OAKINK2}{OAKINK2}{,}
                    \papercite{Nymeria}{Nymeria}{,}
                    \papercite{EgoMimic}{EgoMimic}{}\\
                    \papercite{PH$^2$D}{Human_Policy}{,}
                    \papercite{Humanoid Everyday}{Humanoid_Everyday}{,}
                    \papercite{IndEgo}{IndEgo}{,}
                    \papercite{PLAICraft}{PLAICraft}{,}
                    \papercite{HD-EPIC}{HD-EPIC}{,}
                    \papercite{UniHand}{UniHand}{}\\
                    \papercite{Ego-Centric Human Manipulation Dataset}{Ego-centric-Human-Manipulation-Dataset}{,}
                    \papercite{Aria Everyday Activities}{Aria-Everyday-Activities}{,}
                    \papercite{EgoDex}{EgoDex}{}
                ]
            ]
        ]
        [Evaluation, color=tealgray, fill=tealgray!15, thick, text=black
            [World Model, color=tealgray, fill=tealgray!15
                [Visual Fidelity, color=tealgray, fill=tealgray!15
                    [
                        \textnormal{PSNR}{,}
                        \papercite{SSIM}{ssim}{,}
                        \papercite{LPIPS}{lpips}{,}
                        \papercite{DreamSim}{fu2023dreamsim}{,}
                        \papercite{DINO}{oquab2023dinov2}{,}
                        \papercite{FVD}{unterthiner2018fvd}{}
                    ]
                ]
                [Physical Commonsense, color=tealgray, fill=tealgray!15
                    [
                        \papercite{VideoPhy}{bansal2024videophy}{,}
                        \papercite{PhyGenBench}{meng2024phygenbench}{,}
                        \papercite{VBench-2.0}{zheng2025vbench}{,}
                        \papercite{WorldModelBench}{li2025worldmodelbench}{}\\
                        \papercite{Physics-IQ}{motamed2026physicsiq}{,}
                        \papercite{WorldScore}{duan2025worldscoreunifiedevaluationbenchmark}{,}
                        \papercite{EWMBench}{yue2025ewmbench}{}
                    ]
                ]
                [Action Plausibility, color=tealgray, fill=tealgray!15
                    [
                        \papercite{WorldSimBench}{qin2024worldsimbenchvideogenerationmodels}{,}
                        \papercite{Wow, wo, val!}{fan2026wow}{}
                    ]
                ]
            ]
            [Action Policy, color=tealgray, fill=tealgray!15
                [General, color=tealgray, fill=tealgray!15
                    [
                        \papercite{MetaWorld}{yu2019metaworld}{,}
                        \papercite{RLBench}{RLBench}{,}
                        \papercite{Robomimic}{mandlekar2021robomimic}{,}
                        \papercite{Franka Kitchen}{gupta2019relay}{,}
                        \papercite{ManiSkill}{mu2021maniskill}{}\\
                        \papercite{ManiSkill2}{ManiSkill2}{,}
                        \papercite{ManiSkill3}{tao2024maniskill3}{,}
                        \papercite{RoboCasa}{RoboCasa}{,}
                        \papercite{CALVIN}{mees2022calvin}{,}
                        \papercite{VIMAbench}{VIMA-Bench}{}\\
                        \papercite{VLMbench}{zheng2022vlmbench}{,}
                        \papercite{LIBERO}{LIBERO}{,}
                        \papercite{Libero-plus}{fei2025liberoplus}{,}
                        \papercite{Libero-pro}{zhou2025liberopro}{,}
                        \papercite{Libero-X}{wang2026liberox}{}\\
                        \papercite{COLOSSEUM}{pumacay2024colosseum}{,}
                        \papercite{AGNOSTOS}{zhou2025agnostos}{,}
                        \papercite{RoboEval}{wang2025roboeval}{,}
                        \papercite{RoboVerse}{geng2025roboverse}{,}
                        \papercite{PolaRiS}{jain2025polaris}{}\\
                        \papercite{RoboMME}{dai2026robomme}{,}
                        \papercite{GenManip}{gao2025genmanip}{,}
                        \papercite{VLABench}{zhang2024vlabench}{,}
                        \papercite{RoboSuite}{robosuite}{,}
                        \papercite{RoboLab}{yang2026robolab}{}\\
                        \papercite{SimplerEnv}{li2024simplerenv}{,}
                        \papercite{ARNOLD}{gong2023arnold}{,}
                        \papercite{GemBench}{garcia2025gembench}{}
                    ]
                ]
                [Bimanual and Humanoid Form, color=tealgray, fill=tealgray!15
                    [
                        \papercite{RoboTwin}{RoboTwin}{,}
                        \papercite{BiGym}{chernyadev2024bigym}{,}
                        \papercite{HumanoidBench}{sferrazza2024humanoidbench}{} \\
                        \papercite{HumanoidGen}{li2025humanoidgen}{}
                    ]
                ]
                [Mobile Manipulation, color=tealgray, fill=tealgray!15
                    [
                        \papercite{ManipulaTHOR}{ehsani2021manipulathor}{,}
                        \papercite{HomeRobot}{yenamandra2023homerobot}{,}
                        \papercite{BEHAVIOR-1K}{li2023behavior1k}{}
                    ]
                ]
                [Contact and Deformation Manipulation, color=tealgray, fill=tealgray!15
                    [
                        \papercite{SoftGym}{lin2021softgym}{,}
                        \papercite{PlasticineLab}{huang2021plasticinelab}{,}
                        \papercite{DaXBench}{chen2023daxbench}{}\\
                        \papercite{TacSL}{akinola2024tacsl}{,}
                        \papercite{ManiFeel}{xu2025manifeel}{}
                    ]
                ]
                [Real-Device, color=tealgray, fill=tealgray!15
                    [
                        \papercite{RoboArena}{atreya2025roboarena}{,}
                        \papercite{RoboChallenge}{yakefu2025robochallenge}{,}
                        \papercite{Maniparena}{sun2026maniparena}{}
                    ]
                ]
            ]
        ]
    ]
 \end{forest}
\caption{The comprehensive roadmap and taxonomy of World Action Models (WAMs) reviewed in this survey. The literature is systematically categorized into four core dimensions: background (\autoref{sec:background}), architecture (\autoref{sec:arch}), training data (\autoref{sec:data}), and evaluation protocols (\autoref{sec:eval}).}
\label{fig:survey-sturcture}
\end{figure}
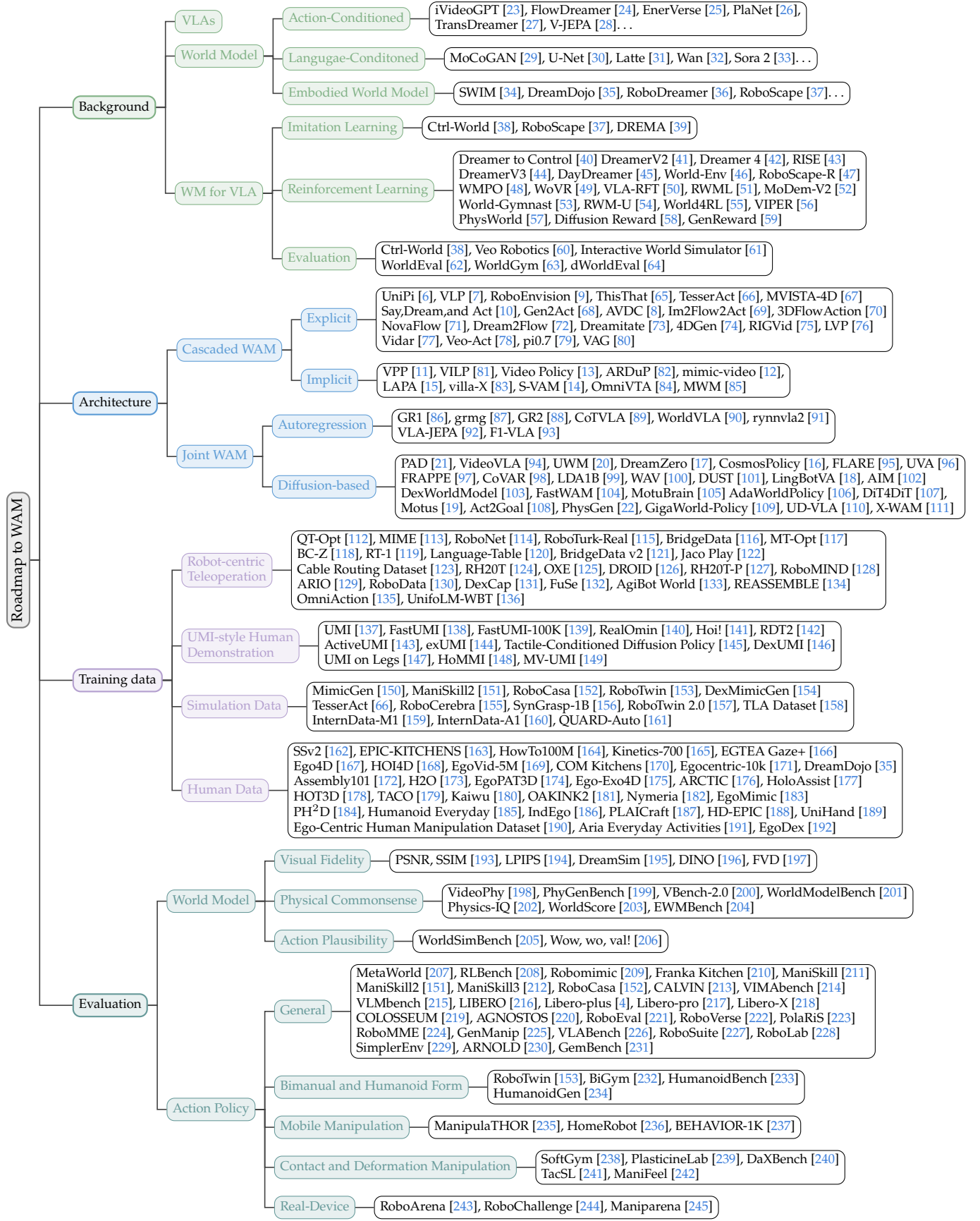

\section{Definitions and Formalism}
\label{sec:def}

To establish a rigorous foundation for World Action Models (WAMs), we define the embodied intelligence task through a probabilistic lens. We consider an embodied agent interacting with an environment. 
At each timestep, the agent receives an observation 
$o \in O$ (encompassing visual inputs, proprioceptive 
signals, and any other sensory modalities), a language instruction 
$l \in L$, and produces actions $a \in A$. 
We use $o'$ to denote the observation at the subsequent timestep.

\subsection{Foundational Paradigms}

\begin{figure}
    \centering
    \includegraphics[width=1\linewidth]{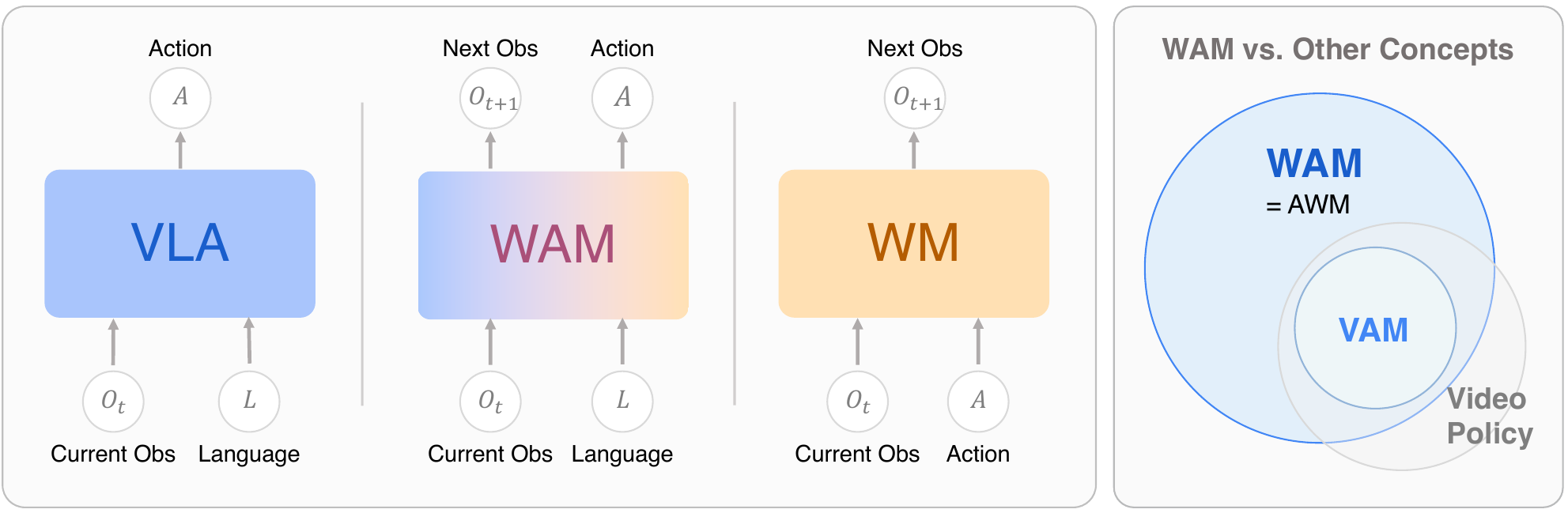}
    \caption{Conceptual definition and comparison of World Action Models (WAMs). The left panel contrasts the input-output formulations of Vision-Language-Action (VLA) models, WAMs, and standard World Models (WMs), highlighting WAM's capability to jointly predict actions and future observations. The right panel illustrates the conceptual scope of WAMs relative to other paradigms such as Video Action Models (VAMs) and Video Policies.}
    \label{fig:definition}
\end{figure}

\textbf{Vision-Language-Action (VLA)} models are a class of embodied foundation models that frame robot control as a multimodal sequence modeling task. In this paradigm, the agent processes current observation $o$ and linguistic instructions $l$ to generate a sequence of action tokens $a$. The VLA architecture typically leverages the pre-trained semantic latent spaces of Large Language Models (LLMs) or Vision-Language Models (VLMs) to map perceptual inputs directly to the action space. Formally, the VLA objective is defined by the conditional probability of actions given the multimodal context:

$$ \mathcal{L}_{\text{VLA}} = \mathbb{E}_{(o, l, a) \sim \mathcal{D}} \left[-\log p(a \mid o, l)\right] $$

\textbf{World Models (WM)} 
are defined as predictive transition functions that internalize the causal dynamics of the physical environment. The functional role of a WM is to model the forward-dynamics of the world—simulating the evolution of environmental observed states $o'$ conditioned on a preceding state $o$ and a sequence of hypothetical interventions $a$. This is typically expressed as:

$$\mathcal{L}_{\text{WM}} = \mathbb{E}_{(o, a, o') \sim \mathcal{D}} 
\left[-\log p(o' \mid o, a)\right]$$

Within this framework, the model acts as a probabilistic propagator of states, providing a representation of how the environment changes in response to specific actions.

\textbf{World Action Models (WAMs)} are a class of embodied foundation models that unify environmental dynamics modeling (world modeling) with motor control (action generation). Unlike standard Vision-Language-Action (VLA) models that learn direct observation-to-action mappings, WAMs predict the future evolution of the physical environment. Formally, a WAM must fulfill two core criteria: 

\begin{enumerate}
    \item \textbf{Forward Predictive Modeling:}
    The model must forecast the physical evolution of the environment by generating or utilizing a quantifiable representation of future states $o'$. This modeling can manifest as explicit visual predictions (e.g., pixel-level video frames, dense optical flow) or implicit physical representations (e.g., physics-grounded latent spaces).
    
    \item \textbf{Coupled Action Generation:} The model must deduce its motor commands $a$ by strictly aligning them with the anticipated future states $o'$. This coupling can manifest as a joint probabilistic output or as policy conditioning within a cascaded or unified latent architecture.
\end{enumerate}

Formally, a WAM seeks to characterize the joint or conditional distribution of future states and actions within a unified framework:

$$\mathcal{L}_{\text{WAM}} = \mathbb{E}_{(o, l, o', a) \sim \mathcal{D}} \left[-\log p(o', a \mid o, l)\right].$$

By moving beyond observation-to-action mapping towards joint state-action prediction, WAMs leverage rich spatiotemporal priors to achieve deeper physical understanding and stronger zero-shot generalization.

\subsection{Disambiguation: WAM vs. Related Concepts}

To ensure conceptual clarity, we distinguish World Action Models (WAMs) from several related concepts in the generative robotics literature.

\begin{enumerate}
    \item \textbf{Video Action Models (VAMs):} 
    VAMs often refer to the models that integrate video prediction with action generation, typically aligning action with synthesized visual futures.
    We define WAMs as a broader, modality-independent superset of predictive agents. While VAMs are specifically optimized to align actions with video frame synthesis, the WAM paradigm posits that video is merely one possible proxy for modeling the world. WAMs encompass models that utilize other predictive targets—such as single-image state transitions, dense point clouds, or multi-sensory modalities like tactile and force feedback. The term ``World'' emphasizes the model's internalization of underlying physical laws and causal dynamics, rather than a commitment to the pixel-level video format.

    \item \textbf{Video Policies:} Video policies often refer to models defined by their structural heritage—using generative video architectures (e.g., Diffusion Transformers) as a backbone to extract strong spatiotemporal representations. This distinction with WAM is rooted in two dimensions: structural heritage and predictive commitment. First, similar to the WAM/VAM distinction, \textit{Video Policies} are conceptually tied to video-generation backbones (e.g., Video Diffusion Transformers). In contrast, WAMs are backbone-agnostic and can be instantiated via any architecture capable of state synthesis across diverse modalities. Second, a model can be categorized as a video policy if it merely inherits the pre-trained spatiotemporal representations of a video model to map observations directly to actions ($p\left( a | o \right)$). A WAM, however, requires an active predictive commitment; it must be supervised by a world-modeling objective where the synthesis of the next state $o'$ is an explicit component of the model's reasoning and output, rather than just an implicit feature within the backbone.

    \item \textbf{Action World Models (AWM):} 
    Action World Model (AWM) is employed in early literature to describe models that integrate world modeling with action generation ($p(o', a \mid o, l)$). 
    While functionally similar, the choice of WAM over AWM reflects a strategic shift in the hierarchy of embodied intelligence. In the ``AWM'' phrasing, the noun is ``World Model'', which characterizes the system as an augmented simulator. World Action Model repositions the system as a primary category of Agent, where ``World'' (predictive physics) and ``Action'' (motor control) are co-equal components. This nomenclature establishes WAM as the direct conceptual successor to the Vision-Language-Action lineage, emphasizing its identity as a complete foundation model for robotics.
\end{enumerate}

\section{VLAs and World Models: Foundations and Early Integration}
\label{sec:background}

\subsection{Vision-Language-Action Models}

Traditional imitation learning often suffered from narrow task-specific designs, where models were trained for isolated skills, severely limiting their generalization in open-world environments. To overcome this, the field shifted toward language-conditioned policies that interpret multimodal task descriptions to compute control actions. Early VLA architectures primarily explored three distinct paradigms for fusing visual and linguistic inputs: (1) Feature Modulation, using FiLM-based layers to condition visual features on language embeddings \cite{BC-Z, RT-1}; (2) Cross-attention Mechanisms, enabling dynamic interaction between task prompts and visual tokens \cite{ VIMA-Bench, RT-2}; and (3) Simple Concatenation, where multimodal tokens are flattened into a unified sequence for joint processing \cite{Gato,  VIMA-Bench}. These works demonstrated the potential for open-vocabulary manipulation \cite{CLIPort} and introduced critical techniques like Action Chunking and Temporal Ensembling to improve motion smoothness and temporal consistency \cite{ACT, RoboSet}.

The success of Large Language Models (LLMs) catalyzed a second wave of VLA research, emphasizing the use of Knowledge Priors and Large-scale Scaling. By inheriting weights from pre-trained Large Vision-Language Models (LVLMs), these agents leverage internet-scale data to achieve sophisticated reasoning and semantic understanding \cite{RT-2, OpenVLA}. Methodologically, this era is characterized by two parallel action generation heads: Autoregressive Tokenization, which treats actions as discrete linguistic tokens generated sequentially \cite{RT-2, OpenVLA, FAST, RoboFlamingo}, and Diffusion-based Synthesis, which attaches a generative action expert to the VLM backbone to produce continuous, multi-modal action distributions \cite{pi0, RDT-1B, pi05, DiffusionVLA}. This duality allows models to balance high-level logical planning with low-level physical precision, scaling from single-arm tasks to complex bimanual manipulations.

Moving beyond the traditional image-to-action pipeline, the definition of VLA has recently expanded to incorporate richer embodied observations.
To enhance the model's perception and interaction capabilities within the physical world, researchers have begun incorporating diverse information sources \cite{OmniAction, FuSe}. These include the integration of 3D geometric information to strengthen spatial representations \cite{3D-VLA}, the utilization of depth perception to improve operational precision \cite{DepthVLA}, and the fusion of force or tactile feedback to enable fine-grained force-controlled assembly tasks \cite{TA-VLA, Tactile-VLA}. This fusion of multimodal inputs is driving VLA models to evolve from simple ``vision-language'' driven systems toward comprehensive ``multimodal physical interaction'' foundation models. However, as these models remain rooted in reactive mapping without capturing the underlying world dynamics, they often struggle with manipulation generalization.

\subsection{World Models}

The definition of a \textbf{world model} has long been the subject of considerable debate. Different works adopt varying definitions of world models \cite{World_Model_Def_1, Sora_simulation, Beyond_World_Models, A_Step_Toward_World_models, WorldModels, lecun2022path}. A consensus is that a world model is an internal representation that models environmental dynamics and the effects of actions. Based on this representation, it can predict the consequences of actions, thereby enabling simulation, decision-making, and planning. Based on the different modes of conditions, world models can be divided into \textbf{Action-conditioned World Models} (\autoref{Action-conditioned}) and \textbf{Language-conditioned World Models} (\autoref{Language-conditioned and Multimodal}). We also discuss world models specialized for embodied environments (\autoref{Embodied}).

\subsubsection{Action-conditioned World Models} 
\label{Action-conditioned}

Action-conditioned world models describe how an environment evolves in response to the agent's actions, where actions refer to executable control signals issued by the agent that directly intervene in the environment and drive state transitions over time. Given the current state and an action, the model predicts the resulting future state or observation, thereby capturing the causal effect of actions on environmental dynamics. Formally, this process can be written as 
$$P\left(o' \mid o, a\right), $$
where the next observed state $o'$ is predicted based on the current observed state $o$ and action $a$. Based on the space in which environment dynamics are modeled and predicted, world models can be divided into two lines: \textbf{Explicit World Models}, which directly predict future observations such as pixels or video frames, and \textbf{Implicit World Models}, which model environment dynamics in a latent representation space.

\paragraph{Explicit Pixel-level Prediction}

Early pixel-level prediction models directly operated in the pixel space by predicting future frames at the pixel level. \textbf{ACVP} \cite{ACVP} designs an Encoding-Transformation-Decoding network architectures, which is one of the first works to make and evaluate long-term predictions on pixel-level video conditioned by actions. \textbf{CDNA} \cite{PixelLevel2} proposes an action-conditioned video prediction model. By explicitly modeling pixel motion through a mechanism that predicts a distribution over transformations, the model can synthesize future frames by moving pixels from previous frames rather than reconstructing appearance from scratch. \textbf{Deep Visual Foresight} \cite{finn2017deepvisualforesightplanning} learns an action-conditioned video prediction model that represents visual dynamics through implicit stochastic pixel flow. The model predicts future frames by generating spatially varying flow operators and masks that advect pixels across time, yielding probabilistic pixel-level transitions without explicit supervision of optical flow, object identity, pose, or physical state. \textbf{SV2P} \cite{babaeizadeh2018stochasticvariationalvideoprediction} further extends this class of models by introducing stochastic variational latent variables into video prediction. By sampling latent variables, the model generates multiple plausible future frame sequences rather than a single deterministic prediction, which is particularly important in real-world and action-conditioned settings where ambiguous interactions may lead to different outcomes. \textbf{MCnet} \cite{PixelLevel3} decomposes motion and content through an asymmetric two-stream encoder. By separating these components, the model simplifies the prediction task to transforming existing content based on identified motion features. \textbf{ContextVP} \cite{PixelLevel4} eliminates blind spots in pixel prediction by using Parallel Multi-Dimensional LSTM (PMD) units that aggregate context from all directions at every processing layer, combined with a Weighted Context Blending block to learn the importance of each direction.

As generative modeling continues to advance, increasingly powerful pretrained video generation models have emerged \cite{VideoGPT, hong2022cogvideolargescalepretrainingtexttovideo, MAGVIT, CogVideoX, Video_Diffusion_Model, Stable_Video_Diffusion, wu2021nuwavisualsynthesispretraining}. Building on these advances, many recent works explore video-based world models by leveraging such pretrained generative models to predict future visual observations directly in pixel space. These approaches can be divided into two paradigms according to their generative formulation: autoregressive video world models and diffusion-based video world models.

\textbf{Autoregressive video world models} represent an important paradigm of generative world models. These approaches tokenize video frames into discrete visual tokens and train a dynamics model to autoregressively predict future tokens conditioned on past observations. Based on VideoGPT \cite{VideoGPT}, a pretrained video model with GPT-like architecture, \textbf{iVideoGPT} \cite{iVideoGPT} designs a scalable autoregressive transformer framework that integrates multimodal signals—visual observations, actions, and rewards—into a sequence of tokens, facilitating agent-environment interaction via next-token prediction. The design of iVideoGPT enables it to function as a fully operational world model. \textbf{Genie} \cite{Genie} further develops interactive world models by introducing a latent action model to infer action variables from unlabeled video clips, which are used to train a dynamics model that predicts future visual tokens. This design allows world models to be trained from large-scale internet videos without requiring action annotations, substantially improving the scalability of world model training. The autoregressive formulation enables world models to generate frame sequences of arbitrary length, making them suitable for interactive world modeling. However, these models suffer from serious error accumulation and may struggle to model highly multimodal distributions. 

\textbf{Diffusion-based video world models} extend generative world models by explicitly modeling the distribution over possible future observations. They generate videos by iteratively denoising a noisy spatiotemporal signal into a coherent sequence conditioned on data or prompts. Instead of predicting a single deterministic future, diffusion models learn the conditional distribution over future trajectories and therefore capture multimodal environment dynamics. These approaches have achieved significant progress in text-to-video generation \cite{Video_Diffusion_Model, Stable_Video_Diffusion}. Diffusion-based video models are increasingly used in world modeling. \textbf{Diffusion World Model} \cite{DiffusionWorldModel} formulates world modeling as conditional diffusion over future video frames. By generating future observations through an iterative denoising process, the model alleviates the error accumulation that often occurs in step-wise dynamics prediction. Subsequent works further develop this paradigm \cite{UWM, Inference-Time_Enhancement, FlowDreamer, EnerVerse, Cosmos-Transfer1, DIAMOND, GameNGen, PEVA}. Owing to the iterative denoising mechanism, diffusion-based video models generally achieve better performance on tasks requiring long-horizon consistency or high-quality generative outputs. However, diffusion-based world models are computationally intensive.

\paragraph{Implicit Latent-space Dynamics Models}

To overcome the inefficiencies of pixel-level modeling, a major line of work focuses on learning dynamics in a compact latent space. Latent dynamics models encode observations into latent states and learn transition functions in this space.

\textbf{Recurrent State-Space Model (RSSM).} 
Some approaches employ recurrent architectures to predict the evolution of latent representations over time. David Ha and Jürgen Schmidhuber \cite{WorldModels,Recurrent_World_Models} proposed a generative recurrent neural network (RNN) trained unsupervised to model standard RL environments. Motivated by this idea, \textbf{PlaNet} \cite{PlaNet} introduces the Recurrent State-Space Model (RSSM), which combines deterministic and stochastic components to capture both predictable dynamics and uncertainty. PlaNet demonstrates the effectiveness of learning dynamics in latent space. The \textbf{Plan2Explore} \cite{Plan2Explore} agent extends the RSSM framework to self-supervised exploration by leveraging planning in latent space to seek expected future novelty, using disagreement among an ensemble of latent dynamics predictors as an intrinsic reward. The \textbf{Dreamer} series and some subsequent works \cite{Dreamer2, Dreamer3, Dreamer4, DreamerV3, DayDreamer, DreamerPro, Dreaming, DreamingV2, SafeDreamer, HarmonyDream, Dream_to_Generalize} extends this framework by enabling planning entirely in latent space, achieving strong performance with improved sample efficiency. Meanwhile, \textbf{LEXA} \cite{LEXA} leverages RSSM to train explorer and achiever policies in latent imagination for unsupervised goal-conditioned reinforcement learning.

\textbf{Transformer State-Space Model (TSSM).} 
World models based on RSSM may inherit some limitations of RNNs. To enable world models to benefit from advances in transformers \cite{Transformer}, \textbf{TransDreamer} \cite{TransDreamer} introduces the Transformer State-Space Model (TSSM), which leverages a transformer for predicting dynamics. TransDreamer and subsequent works \cite{Transformer_1, Transformer_2, STORM, IRIS} show that world models with transformer-based latent sequence modeling outperform Dreamer in modeling long-term dependencies and reasoning.

\textbf{Predictive Representation Learning.}
Instead of reconstructing observations or modeling full dynamics, predictive representation learning focuses on modeling latent embeddings of missing or future observations. \textbf{JEPA} \cite{JEPA}, short for joint-embedding predictive architecture, provides a general paradigm for learning predictive representations in an abstract embedding space. Rather than reconstructing raw inputs, JEPA learns to predict the target embedding from contextual information, encouraging the model to capture high-level, predictable structure while abstracting away low-level details. A instantiation of this paradigm is \textbf{I-JEPA} \cite{I-JEPA}, which learns image representations by predicting the latent representations of multiple target blocks from a single context block within the same image. This line of work was further extended toward motion-aware visual representation learning by \textbf{MC-JEPA} \cite{MC-JEPA}, which jointly learns optical flow and content features within a shared encoder, showing that self-supervised motion estimation can complement content representation learning and improve localized representations for segmentation and video understanding. Building on this idea, \textbf{V-JEPA 2} \cite{V-JEPA-2} extends the JEPA paradigm to video-based world modeling. By predicting future latent embeddings instead of reconstructing pixels, V-JEPA 2 learns more abstract and generalizable environment representations with substantially lower reconstruction overhead. Taking this further into embodied control, \textbf{LeWorldModel} \cite{maes2026leworldmodelstableendtoendjointembedding} proposes an end-to-end JEPA trained stably from raw pixels for latent world modeling, replacing complex multi-term regularization objectives with a single Gaussian-distribution regularizer (SIGReg) to prevent representation collapse, and enabling efficient latent-space planning for continuous control tasks. Beyond visual inputs, \textbf{A-JEPA} \cite{A-JEPA} applies joint-embedding prediction to audio spectrograms, using curriculum time-frequency masking and latent-space prediction to learn scalable audio and speech representations. This predictive representation learning perspective has further inspired subsequent efforts to improve JEPA-style predictive objectives \cite{Rethinking_JEPA, V-JEPA-2.1} and to extend the paradigm to broader embodied and vision-language settings \cite{Drive-JEPA, JEPA-VLA, VLA-JEPA, VL-JEPA}.

\subsubsection{Language-conditioned World Models}
\label{Language-conditioned and Multimodal}

Unlike action-conditioned world models, which predict future states based on low-level control signals, language-conditioned world models use language as a higher-level and more abstract form of conditioning. Instead of specifying an exact action sequence, language provides semantic guidance about the desired scene, event, or evolution process, enabling the model to generate futures that are consistent with textual or multimodal instructions. Formally, this paradigm can be described as
$$P\left(o' \mid o, l\right),$$
where $l$ denotes linguistic conditions.

Within this paradigm, language-conditioned video foundation models have become one of the most prominent realizations of world models. By learning from large-scale video-text pairs, these models acquire rich priors about objects, scenes, physical interactions, camera motion, and temporal dynamics. As a result, they can generate plausible visual futures from high-level semantic descriptions, and thus provide a natural interface for modeling world evolution under linguistic instructions. Existing methods have evolved along multiple key dimensions, including generative objectives, backbone architectures, latent representations, and scaling strategies. We therefore organize the following discussion according to this technical evolution, from early GAN-based video generators to diffusion-based and Transformer-based video foundation models.

Early video foundation models were predominantly GAN-based, such as \textbf{MoCoGAN} \cite{MoCoGAN}, \textbf{TGAN} \cite{TGAN}, and \textbf{DVD-GAN} \cite{DVD-GAN}. Subsequently, with the introduction of diffusion models, diffusion-based video foundation models have become the dominant paradigm.  

Early diffusion-based video foundation models were built upon the \textbf{U-Net} \cite{UNet} architecture, which was originally designed for 2D image generation. Video diffusion models achieve temporal coherence by incorporating dedicated temporal layers into the standard 2D U-Net architecture. Early pioneering works such as \textbf{VDM} \cite{Video_Diffusion_Model} directly extend traditional 2D convolution kernels into 3D kernels to process video sequences as unified spatio-temporal blocks. In contrast, other approaches focus on attention mechanisms, utilizing cross-frame attention blocks to share information across frames, a strategy adopted in models like \textbf{Text2Video-Zero} \cite{Text2Video-Zero} and \textbf{AnimateDiff} \cite{AnimateDiff} to maintain consistent object appearance and motion.

Subsequent works have adopted the Vision Transformer (ViT) \cite{ViT} as a more flexible alternative for video diffusion backbones. Models such as \textbf{Sora} \cite{Sora} and \textbf{Latte} \cite{Latte} treat video data as sequences of spacetime patches, enabling the model to handle variable resolutions and durations more effectively than fixed-grid U-Net. This transition allows models to benefit from the scalability of Transformers, often employing causal attention or spatio-temporal self-attention to model complex dynamics and generate future frames.

Pixel-level diffusion denoising in RGB space is computationally expensive. To address this issue, LDM \cite{LDM} employs \textbf{variational autoencoders (VAEs)} to compress images from pixel space into a latent space, and perform denoising within this compact representation. Motivated by this idea, \textbf{VideoGPT} \cite{VideoGPT} utilizes a 3D-VQVAE to learn discrete latent representations for video generation. Subsequent works further develop this paradigm \cite{Align_Your_Latents, MAGVIT, Movie-Gen, Latte, CogVideoX, LTX-Video}.

With the maturation of video foundation model design, numerous high-performing video foundation models have emerged.
Some open-source video foundation models have demonstrated strong performance. \textbf{Wan} \cite{Wan}, built on the Diffusion Transformer (DiT) paradigm and combined with the Flow Matching framework, demonstrates strong performance across various applications, including video generation, video editing, real-time synthesis, and audio-visual synchronization. Subsequent extensions further specialize the capabilities of Wan toward character animation \cite{Wan-Animate} and motion control \cite{Wan-Move} tasks.
Many closed-source video foundation models, trained on internet-scale data, have demonstrated rich world knowledge and strong predictive capabilities, effectively serving as powerful multimodal world models. Representative models include Sora 2 \cite{Sora_2} by OpenAI, Kling 3 \cite{kling3} by Kuaishou, Veo 3 \cite{Veo_3} by Google, Gen-4 \cite{Gen_4} by Runway, and Pika 2.2 \cite{Pika-22} by Pika Labs.

With strong prior world knowledge and a strong capacity for semantic understanding, language-conditioned and multimodal world models exhibit significant potential in applications such as data synthesis and task planning.

\subsubsection{Embodied World Model}
\label{Embodied}

Embodied environments require models to capture physical world dynamics and anticipate how the world changes in response to agent interactions. This ability to predict future states is crucial for simulation, planning, and data synthesis in embodied scenarios. As world models predict environment evolution conditioned on observations and actions, they provide a natural framework for embodied environment modeling. Many works have focused on improving the predictive accuracy, physical awareness, and dynamics knowledge of world models in embodied settings, with the goal of generating more realistic robotic demonstrations and supporting embodied intelligence.

Internet-scale, unlabeled videos contain rich information about physical dynamics, and learning from such data is crucial for pretraining embodied world models. Several works enable world models to acquire physics dynamics from human operation videos or video-only datasets. \textbf{Genie} \cite{Genie}  introduces a latent action model that infers latent actions from video clips in an unsupervised manner, enabling world models to be trained using video-only internet data. \textbf{SWIM} \cite{SWIM}, built upon the \textbf{Dreamer} \cite{PlaNet, Dreamer2, Dreamer3} architecture, extracts grasp points and motion trajectories from human operation videos to mitigate the substantial morphology gap between humans and robots, enabling the learning of robotic manipulation dynamics from human motion videos. \textbf{DreamDojo} \cite{DreamDojo} introduces continuous latent actions by encoding inter-frame motion into compact vectors, enabling embodiment-agnostic action representations and cross-embodiment knowledge transfer from human demonstrations. \textbf{DexWM} \cite{DexWM} extracts 3D hand keypoints from egocentric videos to represent dexterous actions, employing a hand consistency loss to capture fine-grained finger-object interactions and enabling zero-shot transfer to real-world dexterous manipulation.

To enhance the quality of robotic demonstrations generated by embodied world models, many works have pursued several complementary directions, including improving zero-shot generation, video-instruction alignment, multi-view consistency, and physical awareness through architectural design and data-driven approaches.
Some works improve the robustness of zero-shot video generation. \textbf{RoboDreamer} \cite{RoboDreamer} leverages the inherent compositionality of natural language to decompose video generation into conditional control over multiple underlying primitives, forming a compositional world model with strong zero-shot generalization capabilities and the ability to flexibly combine multimodal instructions. 
Other works focus on improving the alignment of video and instructions. \textbf{IRASim} \cite{IRASim} introduces a frame-level action conditioning module, which injects each action vector into the spatial attention layers of the transformer blocks, enabling robot actions to exert more precise control over the generated video content. \textbf{MiLA} \cite{MiLA} adopts an innovative Coarse-to-(Re)fine divide-and-conquer strategy. By integrating joint denoising correction flows with a temporally progressive denoising schedule, the method maintains multi-view spatial alignment while effectively addressing error accumulation and dynamic distortions in long-horizon generation. 
Another line of work attempts to unify multi-view robot videos. \textbf{Ctrl-World} \cite{Ctrl-World} concatenates images from multiple cameras along the token dimension within a spatial transformer, enabling joint prediction across all viewpoints. Additionally, many works have carefully designed training paradigms and data preprocessing pipelines to accommodate multi-view generation \cite{EnerVerse, geminiroboticsteam, Scalable_Policy_Evaluation, Geometry-aware, vidar, Genie-Envisioner}. 
In addition, some works focus on enhancing the physical awareness of world models. \textbf{RoboScape} \cite{RoboScape} improves physics-aware modeling through adaptive keypoint dynamics learning, which combines dynamic keypoint sampling with trajectory tracking. It also introduces a dual-branch co-autoregressive transformer for coordinated generation of RGB and depth streams, further enhancing the physics-informed capabilities of the world model. \textbf{WoW} \cite{WoW} introduces SOPHIA, a self-optimizing framework in which a vision-language critic evaluates generated videos and guides prompt refinement to improve physical plausibility and causal consistency. WoW also incorporates a Flow-Mask Inverse Dynamics Model that maps predicted visual transitions into executable end-effector actions, thereby closing the imagination-to-action loop.

Another line of work attempts to broaden the capabilities of world models. \textbf{VT-WM (Visuo-Tactile World Models)} \cite{Visuo-Tactile} introduces tactile modality into embodied world modeling. It employs the Sparsh-X model to encode tactile information into tactile tokens, which are concatenated with visual tokens along the spatial dimension, thereby incorporating touch into embodied world models. \textbf{PointWorld} \cite{PointWorld} unifies environment states and robot actions under a 3D point-flow representation. By scaling training on large-scale real-world and simulated datasets, it constructs a general 3D world model that can generalize across diverse robotic manipulation tasks, supports real-time inference, and can be directly integrated with model predictive control. 

\subsection{World Model for VLA}

World models are capable of modeling how the environment evolves under actions, language instructions, or multimodal contexts, therefore providing a key mechanism for enhancing VLA systems beyond direct policy learning from static datasets. World models enable VLA agents to reason about future observations, generate imagined trajectories, estimate task outcomes, and evaluate policy behaviors before real-world execution. This capability is particularly important for robotic learning, where real-world data collection is expensive, physical interaction is potentially unsafe, and exhaustive policy evaluation is difficult to conduct at scale. We discuss how world models contribute to VLA from two complementary perspectives: learning and evaluation. For learning, world models can augment imitation learning, support model-based reinforcement learning, and provide reward signals from predicted futures. For evaluation, they act as data-driven simulators that enable scalable, reproducible, and safety-aware testing of VLA policies.

\begin{figure}
    \centering
    \includegraphics[width=1\linewidth]{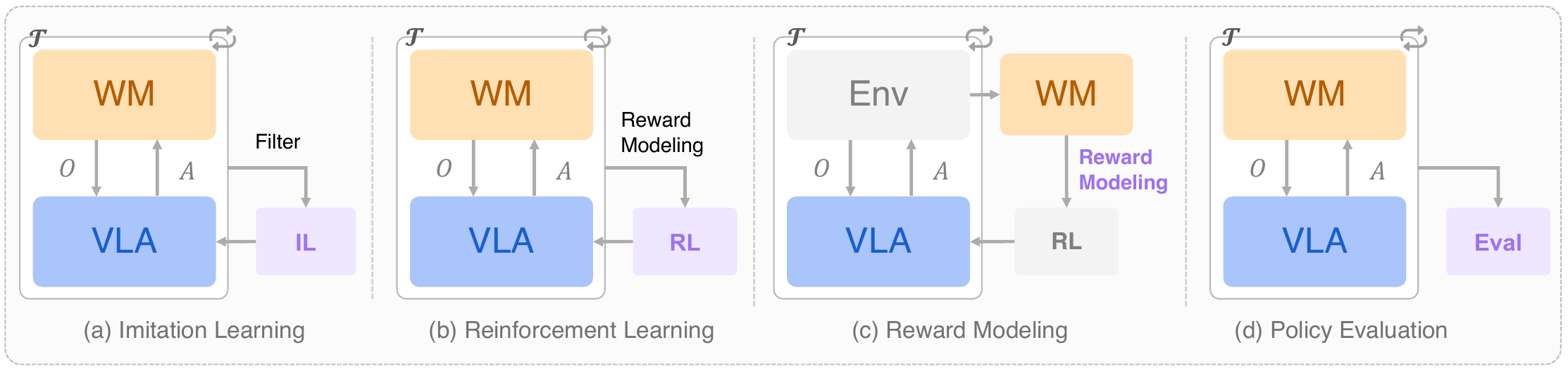}
    \caption{Schematic overview of world models for VLA learning and evaluation. World models can support \textbf{(a) imitation learning} by generating or filtering training trajectories, \textbf{(b) reinforcement learning} by enabling imagined interaction and reward-guided policy optimization, \textbf{(c) reward modeling} by producing reward signals from learned dynamics or future outcomes, and \textbf{(d) policy evaluation} by serving as data-driven simulators for virtual rollout and testing. Here, $\mathcal{T}$ denotes the rollout trajectories.}
    \label{fig:placeholder}
\end{figure}

\subsubsection{World Models for Learning}

\paragraph{Imitation learning (IL).} IL is a primary training paradigm for VLA, in which robotic policies are learned by mimicking expert demonstrations. However, its effectiveness is largely constrained by the limited amount and diversity of expert data. To address these limitations, numerous studies leverage embodied world models to generate diverse training data for imitation learning, thereby improving its performance. \textbf{DREMA} \cite{DREMA} constructs a compositional manipulation world model by combining object-centric Gaussian Splatting with a physics simulator, enabling robots to imagine novel object configurations via equivariant transformations and generate diverse demonstrations for imitation learning, achieving one-shot policy learning on a real robot. \textbf{Ctrl-World} \cite{Ctrl-World} significantly enhances VLA models by enabling effective learning in imagination. By synthesizing successful trajectories for complex tasks and using them for supervised fine-tuning, it improves the success rate of the $\pi_{0.5}$-DROID \cite{pi05} policy by 44.7\% on downstream tasks. \textbf{RoboScape} \cite{RoboScape} shows that VLA policies like $\pi_0$ \cite{pi0} and Diffusion Policy \cite{Diffusion_Policy} trained on its synthetic data achieve performance comparable to those trained on real-world demonstrations.

\paragraph{Reinforcement learning (RL).} RL is an important paradigm for policy optimization in VLA. However, its application is often hindered by the high cost and risk of real-world robot deployment, the difficulty of designing effective reward functions, as well as the challenge of achieving reliable sim-to-real transfer in simulation environments. World models model environment dynamics by predicting future states, while also providing structured representations of the environment with capabilities for world understanding and cross-modal transformation. These properties enable them to mitigate the aforementioned challenges by supporting safe and scalable policy learning without extensive real-world interaction, facilitating implicit reward estimation, and improving the robustness of sim-to-real transfer. Consequently, world models have been increasingly employed to support policy learning in model-based RL settings.

\textbf{World Models as Surrogate Environments.} 
Many studies employ world models as surrogate environments to facilitate policy learning in RL. In this paradigm, the agent performs imagined rollouts within a learned world model. The predicted trajectories are then used to compute rewards using task-specific reward functions or rollout-level consistency objectives. 
The \textbf{Dreamer} series \cite{Dreamer2, Dreamer3, Dreamer4, DreamerV3, DayDreamer} extends the RSSM introduced by PlaNet \cite{PlaNet} from planning to policy learning, and demonstrates the effectiveness of model-based RL in virtual environments. 
Based on pretrained video generation-based world models that possess pixel-level generation capabilities, several works have attempted to use world models as environments in reinforcement learning \cite{World-Env, world-vla-loop, WMPO, RISE, WoVR, MoDem-V2, World-Gymnast, Uncertainty-Aware, World4RL, DiWA}.
These approaches feed actions generated by the policy, together with current observations, into the world model, which then predicts future observations. Policies are then trained using these generated observations. 
Another line of work \cite{RWML, VLA-RFT, Robot_Learning_from_a_Physical_World_Model} leverages world-model-generated trajectories and measures their agreement with target future states, using this signal to guide policy learning.

\textbf{World Models for Reward Modeling.} 
In addition to serving as simulators for policy learning, several recent works explore world models that are more directly coupled with reward modeling, enabling reward signals to be derived from learned generative representations or jointly modeled environment dynamics rather than relying solely on externally specified reward functions. 
One line of work derives reward signals directly from pretrained generative world models. \textbf{VIPER} \cite{VIPER} leverages pretrained video prediction models to evaluate the compatibility between observed trajectories and the learned dynamics of successful behaviors, and uses this signal as an action-free reward for RL. \textbf{Diffusion Reward} \cite{Diffusion_Reward} derives reward signals from the conditional entropy of video generation distributions learned from expert videos via conditional video diffusion models. \textbf{GenReward} \cite{GenReward} leverages pretrained video diffusion models to provide goal-driven reward signals by measuring the latent alignment between agent observations and generated goal videos, thereby offering video-level guidance for policy learning. \textbf{SRPO} \cite{fei2025srposelfreferentialpolicyoptimization} exploits the latent space of V-JEPA 2 as a pretrained world representation for reward shaping. By encoding successful and failed trajectories into the latent world space and measuring their behavioral similarity, SRPO assigns progress-wise rewards to failed rollouts, providing denser supervision for VLA-RL and alleviating reward sparsity without additional expert demonstrations or manually designed reward functions.
Another line of work integrates reward prediction directly into the world model. For example, \textbf{RoboScape-R} \cite{RoboScape-R} jointly models future observations and rewards, allowing the learned world model to generate reward signals during imagined rollouts and thereby providing an endogenous reward mechanism for RL.

\subsubsection{World Models for Evaluation}

Evaluating robotic policies on physical systems can be time-consuming, resource-intensive, unsafe, and non-reproducible. A common alternative is to conduct evaluation in manually designed simulators such as IsaacGym \cite{IsaacGym} and MuJoCo \cite{Mujoco}. However, these simulators often struggle to capture the complexity and variability of real-world environments, leading to a significant sim-to-real gap. 

By learning environment dynamics directly from data, world models can construct data-driven simulation environments that more faithfully reflect real-world complexity and variability. This reduces reliance on costly and unsafe physical evaluations by enabling large-scale policy testing in a fully virtual setting. Moreover, their ability to perform closed-loop rollouts supports consistent and reproducible evaluation protocols, mitigating the non-reproducibility issues of real-world experiments. In addition, world models can better bridge the sim-to-real gap by grounding simulation in real data distributions, thereby providing more reliable estimates of policy performance prior to deployment. A common approach is to roll out policies and predict their consequences within imagined environments. \textbf{Ctrl-World} \cite{Ctrl-World} injects action signals into every frame via frame-level action conditioning, supporting closed-loop interaction between policies and simulators for policy evaluation. \textbf{Veo Robotics} \cite{geminiroboticsteam} leverages an action-conditioned video generation model as a world simulator to evaluate robotic policies without real-world execution. Veo Robotics also leverages generative image editing to synthesize out-of-distribution scenarios for generalization testing, and performs red teaming by simulating safety-critical situations, with all predictions validated against extensive real-world evaluations. Some works also explore enhancing interactivity and human-in-the-loop control within world model-based simulators. \textbf{Interactive World Simulator} \cite{InteractiveWorldSimulator} enables users to interact with the world simulator by building a real-time teleoperation interface, resulting in a more comprehensive simulation environment. Many similar works also improve the quality, efficiency, and flexibility of evaluation \cite{WorldEval, WorldGym, li2026dworldevalscalableroboticpolicy}.

\section{Architecture}
\label{sec:arch}

While the preceding section discussed world models as exogenous tools for policy training and evaluation—serving as simulation environments, reward models, or robust benchmarks—recent advances have shifted toward integrating world modeling directly into the policy architecture. This transition transforms the world model from an offline supervisor into an internal predictive core, enabling World Action Models (WAMs) to reason about world dynamics in real-time. 
We categorize these World Action Models architectures into two primary paradigms based on their structural flow and corresponding training regimes: (1) \textbf{Cascaded WAM} (\autoref{sec:cascaded_wam}) employs a sequential pipeline that first predicts the next state (e.g., in pixel, latent, or flow space) and subsequently derives the corresponding action. Due to this structural decoupling, they are characterized by a \textit{separated training} process where the world model and the action decoder are optimized as distinct modules; (2) \textbf{Joint WAM} (\autoref{sec:joint_wam}) unifies predictive state modeling and action generation within a single cohesive model, producing future states and actions simultaneously. In this paradigm, world modeling and action generation undergo \textit{joint training} under a unified objective, forcing the model to internalize the causal interdependencies between environmental dynamics and control signals.

\begin{figure}
    \centering
    \includegraphics[width=1\linewidth]{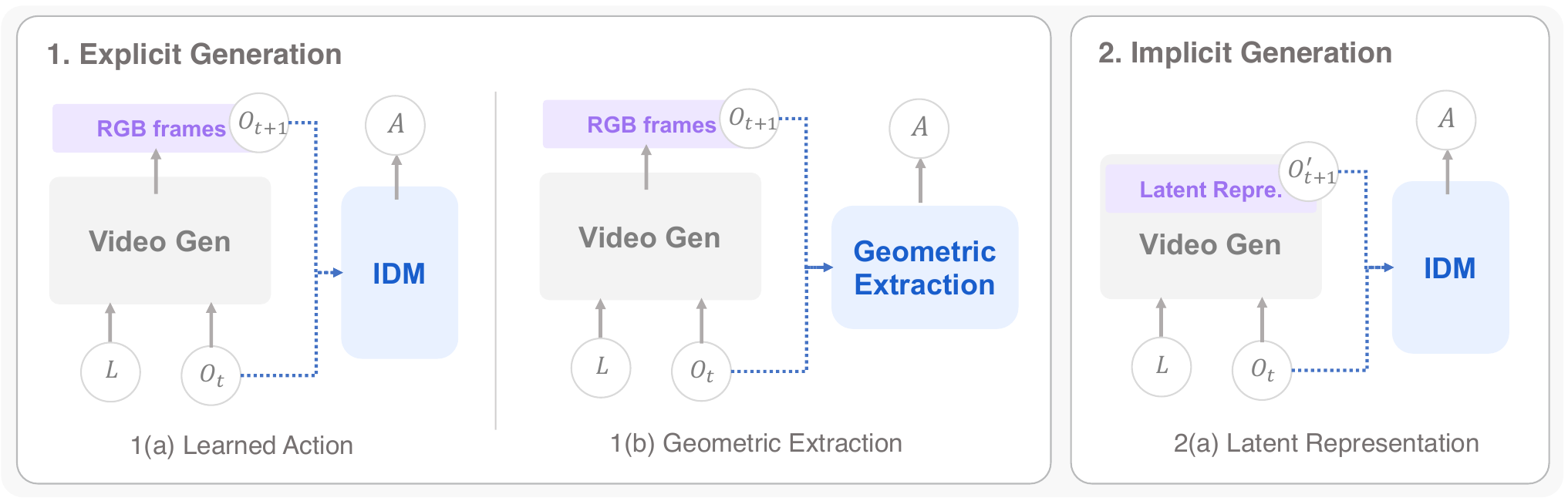}
    \caption{Schematic comparison of cascaded WAM structures. \textbf{1(a) Learned Action}: a world model generates an explicit pixel-space future plan, which is mapped to actions by a learned inverse-dynamics or action model. \textbf{1(b) Geometric Extraction}: the explicit visual plan is converted into actions or trajectories through geometric extraction. \textbf{2(a) Latent Representation}: the intermediate planning carrier is a latent future representation rather than future RGB frames, and the downstream action model decodes executable commands from it.}
    \label{fig:cascaded_wam_taxonomy}
\end{figure}

Cascaded World-Action-Models implement the world-action mapping through a sequential two-stage pipeline: a world model first synthesizes a visual plan representing the anticipated future, after which a separate action model decodes executable robot commands from that plan. This decomposition offers a natural inductive bias — the world model need not reason about robot kinematics, while the action model need not solve long-horizon scene prediction — but it also introduces a coupling between the two stages that shapes every design decision within this family of approaches. Based on the type of intermediate planning carrier, Cascaded WAMs can be broadly categorized into two major classes: (1) \textbf{Explicit Planning} (\autoref{sec:cacaded_wam_explicit}) via pixel-space representations and (2) \textbf{Implicit Planning} (\autoref{sec:cacaded_wam_implicit}) via latent representations. \autoref{fig:cascaded_wam_taxonomy}  provides a schematic overview of these cascaded patterns.

\subsection{Cascaded World-Action-Model}

\label{sec:cascaded_wam}

\subsubsection{Explicit Planning via Pixel-Space Representations}

\label{sec:cacaded_wam_explicit}

The most direct realization of cascaded WAMs uses raw pixel frames as the intermediate representation between the two stages. Pixel-space plans are immediately interpretable and leverage the full representational capacity of large video generative models pretrained on internet-scale data, making them a natural starting point. Work in this space divides broadly by how actions are subsequently extracted from the synthesized video: through learned inverse dynamics, or through closed-form geometric computation.

\paragraph{Learned Action Extraction.}

\textbf{UniPi}~\cite{unipi} established the foundational two-stage blueprint: a text-conditioned spatiotemporal U-Net diffusion model synthesizes a task execution video, from which a convolutional inverse dynamics model (IDM) regresses actions between consecutive frame pairs. While demonstrating the viability of this closed loop, single-pass long-horizon generation was found to suffer from semantic drift and compounding errors. Subsequent work attacked this problem from complementary angles. \textbf{VLP}~\cite{du2023vlp} introduced semantic-level intervention by equipping the pipeline with a vision-language model (VLM) for hierarchical sub-action generation and tree-search-based value scoring, bounding error accumulation within each individual segment. \textbf{RoboEnvision}~\cite{yang2025roboenvision} instead adopted a non-autoregressive strategy, using VLM-decomposed subtask instructions to condition the generation of keyframes representing subtask terminal states, with full video subsequently synthesized by interpolation between those anchors.

Controllability and generation efficiency received focused attention in parallel. \textbf{ThisThat}~\cite{wang2024thisthat} addressed the fundamental ambiguity of language-only grounding in scenes containing multiple instances of the same object class by conditioning video generation jointly on deictic referring expressions (``this''/``that'') and paired gesture coordinates, transmitting manipulation intent without linguistic ambiguity. \textbf{Say, Dream, and Act}~\cite{gu2026saydreamact} pursued efficiency from a different direction, employing adversarial distillation to reduce denoising steps while introducing a frame-rate-agnostic video prediction mechanism that decouples trajectory planning from any fixed execution frequency, enabling unified prediction over variable-length horizons.

The spatial expressiveness of the planning representation has also been extended beyond standard RGB video. \textbf{TesserAct}~\cite{TesserAct1} augments the video prediction target with depth and surface normal channels, introducing explicit geometric constraints into the planning carrier and providing richer cues for downstream action extraction. \textbf{MVISTA-4D}~\cite{wang2026mvista4d} replaces the conventional IDM entirely with a two-step mechanism consisting of trajectory-level latent optimization followed by residual IDM refinement, alleviating the ill-posed nature of per-step action extraction.

Extending the approach across embodiment types and data sources has driven a further line of work. \textbf{Vidar}~\cite{vidar} builds on a diffusion-based video foundation model to translate human manipulation videos into robot execution videos, simultaneously addressing unified bimanual robot modeling by encoding robot type, camera layout, task instructions, and scene context as global generation conditions with masked prediction to focus on interaction-critical regions. \textbf{Gen2Act}~\cite{bharadhwaj2024gen2act} pushes further toward embodiment-agnostic operation by calling a pretrained \textbf{VideoPoet}~\cite{VideoPoet} model zero-shot to generate human manipulation video without any fine-tuning, and replacing hard-coded geometric extraction rules with a closed-loop neural policy that uses point tracking as an auxiliary loss to implicitly encode motion information.
\textbf{Veo-Act}~\cite{zhang2026veoact} addresses the precision gap of foundation video models in contact-rich phases via a gating mechanism: a multi-head IDM extracts actions from Veo-3-generated video for coarse navigation, and an online interaction detector hands control over to a reactive VLA policy once imminent contact is detected.

Rather than extracting actions via a specific IDM, several works replace it with alternative decoders. \textbf{VAG}~\cite{lang2026vag} adopts a 1D U-Net as the action decoder: the video diffusion branch completes the full denoising process to pixel space, and intermediate features from the video U-Net condition a separate action U-Net. \textbf{$\pi_{0.7}$}~\cite{pi0.7} replaces the second stage IDM with a pretrained VLA. A BAGEL-based world model \cite{bagel} generates multi-view subgoal images of the desired near-future state, which are injected directly into the VLA's context window alongside language and episode metadata; the VLA's flow-matching action expert then produces action chunks conditioned on this imagined future. This shifts the burden of generalization from inverse dynamics learning to the VLA's pretrained in-context understanding, enabling strong out-of-the-box and zero-shot cross-embodiment performance without task-specific fine-tuning.

\paragraph{Geometric Action Extraction.}
A parallel line of work replaces the learned IDM with geometric computation over structured intermediate representations, converting the action extraction problem from an inverse dynamics learning problem into an analytically tractable geometric one. This family divides by whether motion is represented as optical flow or as tracked object poses.

Optical flow as an intermediate representation was introduced by \textbf{AVDC}~\cite{ko2024avdc}, which separates video synthesis from action extraction entirely: a text-conditioned diffusion model generates full pixel-level video, from which dense optical flow is computed and SE(3) transformations are analytically derived, requiring no action annotations during training.

\newcommand{\yes}{$\checkmark$}
\newcommand{\no}{$\times$}

\begin{table}[!t] 
\centering
\scriptsize
\setlength{\tabcolsep}{3.5pt}
\renewcommand{\arraystretch}{1.08}
\caption{Comparison of Cascaded World-Action-Model (WAM) methods. \textbf{Act.\ Label}: whether action annotations are required. \textbf{Zero-shot}: no task-specific training/fine-tuning needed. Sim = simulation benchmark; Real = physical robot.}
\label{tab:cascaded_wam}

\begin{adjustbox}{width=\linewidth}
\begin{threeparttable}
\begin{tabular}{
  l
  >{\RaggedRight\arraybackslash}p{2.20cm}
  >{\RaggedRight\arraybackslash}p{3.50cm}
  >{\RaggedRight\arraybackslash}p{3.60cm}
  l
  l
  >{\RaggedRight\arraybackslash}p{5.80cm}
}
\toprule
\textbf{Method} &
\textbf{Interm.\ Repr.} &
\textbf{Stage-1 Backbone} &
\textbf{Stage-2 Model} &
\makecell{\textbf{Act.}\\\textbf{Label}} &
\makecell{\textbf{Zero-}\\\textbf{shot}} &
\textbf{Evaluation} \\
\midrule

\rowcolor{gray!12}
\multicolumn{7}{l}{\textbf{Pixel-space Representations --- Learned Action Extraction}} \\
\midrule
UniPi \cite{unipi}
  & Pixel RGB
  & Video U-Net
  & Lightweight IDM (CNN\,+\,MLP)
  & \yes & \no
  & Sim: PDSketch block manip., CLIPort \newline Real: WidowX \\
VLP \cite{du2023vlp}
  & Pixel RGB
  & Video U-Net\,+\,PaLM-E 12B
  & LAVA
  & \yes & \no
  & Sim: Language Table \newline Real: Language Table robot, 7-DoF mobile arm, 14-DoF ALOHA bimanual \\
RoboEnvision \cite{yang2025roboenvision}
  & Pixel RGB
  & OpenSora (DiT)
  & OpenSora DiT
  & \yes & \no
  & Sim: LanguageTable, LHMM (custom) \newline Real: - \\
This\&That \cite{wang2024thisthat}
  & Pixel RGB
  & U-Net
  & Custom closed-loop neural policy
  & \yes & \no
  & Sim: IsaacGym \newline Real: - \\
Say, Dream \& Act \cite{gu2026saydreamact}
  & Pixel RGB
  & COSMOS-PREDICT2
  & Transformer (ACT)\,+\,Qwen2.5
  & \yes & \no
  & Sim: LIBERO \newline Real: Franka 7-DoF arm \\
TesserAct \cite{TesserAct1}
  & Pixel RGB-DN
  & CogVideoX (DiT)
  & PointNet++, MLP
  & \yes & \no
  & Sim: RLBench \newline Real: - \\
MVISTA-4D \cite{wang2026mvista4d}
  & Pixel RGB-D
  & WAN2.2 TI2V
  & PointNet
  & \yes & \no
  & Sim: RLBench, RoboTwin2 \newline Real: AgileX Piper \\
Vidar \cite{vidar}
  & Pixel RGB
  & Wan2.2 (sim)\,/\,Vidu~2.0 (real)
  & Masked IDM (MIDM)
  & \yes & \no
  & Sim: RoboTwin 2.0 \newline Real: Aloha \\
Gen2Act \cite{bharadhwaj2024gen2act}
  & Pixel RGB
  & VideoPoet
  & Custom closed-loop neural policy
  & \yes & \no
  & Sim: - \newline Real: Mobile manipulation robot \\
Veo-Act \cite{zhang2026veoact}
  & Pixel RGB
  & Veo-3
  & Multi-head IDM + VLA ($\pi_{0.5}$)
  & \yes & \no
  & Sim: IsaacLab \newline Real: 7-DoF arm\,+\,12-DoF dexterous hand \\
VAG \cite{lang2026vag}
  & Pixel RGB
  & Cosmos-Predict2
  & 1D U-Net
  & \yes & \no
  & Sim: LIBERO \newline Real: AgiBot G1, AgileX Cobot Magic \\
$\pi_{0.7}$ \cite{pi0.7}
  & Pixel RGB 
  & BAGEL 
  & VLA
  & \yes & \no
  & Sim: - \newline Real: Mobile manipulators, Static bimanual, UR5e \\
\midrule
\rowcolor{gray!12}
\multicolumn{7}{l}{\textbf{Pixel-space Representations --- Geometric Extraction}} \\
\midrule
AVDC \cite{ko2024avdc}
  & Pixel RGB $\to$ optical flow
  & U-Net
  & Off-the-shelf geometric pipeline (learning-free)
  & \no & \no
  & Sim: Meta-World, iTHOR \newline Real: Franka Emika Panda \\
Im2Flow2Act \cite{xu2024im2flow2act}
  & Optical flow
  & AnimateDiff\,+\,Stable Diffusion (latent)
  & Flow-conditioned IL policy
  & \yes & \no
  & Sim: Custom tasks \newline Real: UR5e \\
3DFlowAction \cite{zhi2025threedflowaction}
  & 3-D optical flow
  & AnimateDiff\,+\,Stable Diffusion
  & Flow-constrained optimization policy
  & \no & \no
  & Sim: Custom tasks \newline Real: Franka, XTrainer \\
NovaFlow \cite{li2025novaflow}
  & Optical flow
  & Wan~2.1\,+\,off-the-shelf modules
  & Flow-based executor
  & \no & \yes
  & Sim: - \newline Real: Franka Emika Panda, Boston Dynamics Spot \\
Dream2Flow \cite{dharmarajan2025dream2flow}
  & Optical flow
  & Off-the-shelf I2V model
  & CoTrackerV3\,+\,trajectory opt.\,/\,RL
  & \no & \yes
  & Sim: OmniGibson, RoboSuite \newline Real: Franka arm, Spot, GR1 humanoid \\
Dreamitate \cite{liang2024dreamitate}
  & Pixel RGB
  & U-Net
  & MegaPose
  & \no & \no
  & Sim: - \newline Real: UFACTORY xArm~7, UR5 \\
4DGen \cite{liu2025fourdgen}
  & Pixel RGB-D\,+\,4-D point maps
  & U-Net
  & FoundationPose\,+\,SAM2
  & \no & \no
  & Sim: LBM \newline Real: Dual Franka Panda \\
RIGVid \cite{patel2025rigvid}
  & Pixel RGB
  & Kling v1.6
  & GPT-4o, depth estimator,\allowbreak FoundationPose,\allowbreak Grounding DINO\,+\,SAM-2,\allowbreak AnyGrasp
  & \no & \yes
  & Sim: - \newline Real: XArm7, Aloha \\
LV-P \cite{chen2025lvp}
  & Pixel RGB
  & WAN~2.1 I2V 14B
  & HaMeR, MegaSAM,\allowbreak Dex-Retargeting, cuRobo
  & \no & \yes
  & Sim: - \newline Real: Franka Emika\,+\,parallel gripper, Unitree G1\,+\,Inspire dexterous hand \\
\midrule

\rowcolor{gray!12}
\multicolumn{7}{l}{\textbf{Implicit Planning via Latent Representations}} \\
\midrule
Video Policy \cite{liang2025videopolicy}
  & Latent features
  & U-Net
  & Action U-Net
  & \yes & \no
  & Sim: RoboCasa, Libero10 \newline Real: Unspecified \\
ARDuP \cite{huang2024ardup}
  & Latent video frames\,+\,active-region encoding
  & U-Net
  & Lightweight CNN
  & \yes & \no
  & Sim: CLIPort \newline Real: WidowX \\
mimic-video \cite{pai2025mimicvideo}
  & Latent video
  & Cosmos-Predict2
  & Action Decoder DiT
  & \yes & \no
  & Sim: SIMPLER-Bridge, LIBERO \newline Real: Dual Franka Emika Panda\,+\,16-DoF dexterous hands \\
VPP \cite{hu2024vpp}
  & Latent video
  & Stable Video Diffusion
  & VideoFormer\,+\,DiT Diffusion Policy
  & \yes & \no
  & Sim: CALVIN, MetaWorld \newline Real: Franka Panda, Xarm\,+\,12-DoF Xhand \\
VILP \cite{xu2025vilp}
  & Latent video
  & U-Net\,+\,3-D convolution
  & Two CNN encoders\,+\,MLP head
  & \yes & \no
  & Sim: Custom tasks \newline Real: Franka Panda \\
LAPA \cite{LAPA}
  & Discrete latent action tokens
  & C-ViViT tokenizer
  & LWM-Chat-1M
  & \yes & \no
  & Sim: Language Table, SIMPLER \newline Real: 7-DoF Franka Emika Panda \\
villa-X \cite{chen2025villax}
  & Discrete latent action tokens
  & Spatial-Temporal Transformer \,+\,ViT \,+\,proprio FDM
  & PaliGemma VLM
  & \yes & \no
  & Sim: SIMPLER, LIBERO \newline Real: gripper manip., dexterous hand manip. \\
S-VAM \cite{yan2026svam}
  & Latent features
  & Stable Video Diffusion
  & Uni-Perceiver
  & \yes & \no
  & Sim: CALVIN, MetaWorld \newline Real: AgileX Cobot dual-arm \\
OmniVTA \cite{zheng2026omnivta}
  & Latent tactile features
  & Two-stream DiT
  & Diffusion Policy\,+\,MLP controller
  & \yes & \no
  & Sim: - \newline Real: 7-DoF xArm\,+\,tactile sensors \\
MWM \cite{lou2026mwm}
 & Latent features
 & DiT
 & Diffusion Policy
 & \yes & \no
 & Sim: LIBERO, RLBench \newline Real: Franka  \\
\bottomrule
\end{tabular}
\end{threeparttable}
\end{adjustbox}
\end{table}
\textbf{Im2Flow2Act}~\cite{xu2024im2flow2act} relocates flow estimation to latent space, using an \textbf{AnimateDiff}~\cite{guo2024animatediff}-based flow generation network to bypass pixel-level video synthesis altogether; a separately trained flow-conditioned policy then maps the flow field directly to actions, trading pixel-level appearance information for computational efficiency. \textbf{3DFlowAction}~\cite{zhi2025threedflowaction} lifts this representation from two dimensions to three, using video diffusion to generate dense three-dimensional flow fields that capture rotational and depth-displacement motion components inaccessible to planar flow. \textbf{NovaFlow}~\cite{li2025novaflow} and \textbf{Dream2Flow}~\cite{dharmarajan2025dream2flow} push the paradigm toward zero-training operation, using pretrained video generation models directly and deriving three-dimensional object-level flow via depth estimation and point tracking before converting to robot commands without demonstration data or additional training.

Pose tracking constitutes the second geometric extraction route. \textbf{Dreamitate}~\cite{liang2024dreamitate} uses a tool as the linking proxy between human and robot action: a stereo video diffusion model fine-tuned on human tool-use demonstrations generates task execution video conditioned on stereo scene images, from which \textbf{MegaPose}~\cite{labbe2022megapose} tracks the tool's 6-DoF pose frame-by-frame; stereo geometry constrains depth estimation, and inverse kinematics maps the resulting trajectory to joint commands, making action extraction entirely independent of robot morphology. \textbf{4DGen}~\cite{liu2025fourdgen} extends this to multi-view consistent generation, jointly predicting RGB video and three-dimensional point map sequences from two-view RGB-D conditioning using an SVD backbone. \textbf{RIGVid}~\cite{patel2025rigvid} applies \textbf{FoundationPose}~\cite{wen2024foundationpose} to track the manipulated object rather than the tool, replaces inverse kinematics with trajectory retargeting, and substitutes human demonstration video with zero-shot diffusion-generated video filtered for quality by a VLM. \textbf{LVP}~\cite{chen2025lvp} reconstructs three-dimensional hand poses from generated human manipulation video and maps them to end-effector trajectories through kinematic mapping, while adopting Diffusion Forcing — partitioning video into low-noise historical and high-noise future segments with causal masking — to improve temporal coherence in generation.

\subsubsection{Implicit Planning via Latent Representations}

\label{sec:cacaded_wam_implicit}

The computational overhead of pixel-level video synthesis constitutes the primary bottleneck against real-time deployment. The implicit planning path is motivated by the observation that intermediate latent representations formed during diffusion already encode the dynamical information required for planning; decoding back to pixel space is therefore unnecessary. The planning carrier is replaced by latent feature sequences that remain in the compressed representation space throughout.

\textbf{VPP}~\cite{hu2024vpp} demonstrated this trade-off directly: a pretrained VAE encodes observation frames, a diffusion model performs single-step prediction of future latent sequences, and a lightweight policy network conditions on these latents to produce actions, achieving planning inference speeds compatible with real-time control for the first time in this framework. \textbf{VILP}~\cite{xu2025vilp} introduced multi-view latent planning within the same paradigm, generating complete latent video sequences from two simultaneous viewpoints and recovering actions with a separately trained state policy network.
\textbf{S-VAM}~\cite{yan2026svam} directly addresses the generation quality cost incurred by VPP's single-step inference, using self-distillation to bridge the gap between efficiency and fidelity: during training, a frozen multi-step SVD backbone provides structured teacher representations that supervise lightweight spatio-temporal decouplers, condensing iterative generation into a single forward pass. At inference, one-step features are aggregated via a QFormer-style perceiver and condition a DiT-based diffusion policy, recovering high-quality planning at real-time control frequencies.

\textbf{Video Policy}~\cite{liang2025videopolicy}, while using explicit pixel video in its first stage, pioneered a key implicit extraction technique: after fine-tuning SVD for video prediction, all video U-Net weights are frozen and a separate action U-Net is trained to generate action sequences conditioned on intermediate features from the video decoder, establishing feature-level conditioning as an alternative to pixel or flow inputs. \textbf{ARDuP}~\cite{huang2024ardup} derives pseudo-supervision from \textbf{Co-Tracker}~\cite{karaev2023cotracker} dense motion point tracking and \textbf{SAM}~\cite{kirillov2023sam}-generated interaction region masks during training, using the resulting active-region proposals as conditions injected into a video diffusion model to concentrate generation on task-relevant areas. \textbf{mimic-video}~\cite{pai2025mimicvideo} advances generation efficiency through flow matching in place of DDPM and a partial denoising strategy that extracts features at an intermediate ODE integration checkpoint, bypassing the full generation path. \textbf{MWM}~\cite{lou2026mwm} replaces RGB prediction with future semantic mask latents, caching the hidden states from its frozen mask-forecasting backbone to condition an action diffusion head. This geometric information bottleneck effectively filters out photometric nuisances, yielding highly robust control under severe visual shifts.

\textbf{OmniVTA}~\cite{zheng2026omnivta} extends implicit latent planning to the visuo-tactile domain. A two-stream diffusion transformer jointly generates future visual and tactile latents, and a downstream fusion policy consumes the predicted tactile latents via a differential encoder that captures discrepancies between predicted and current contact states, enabling robust generalization to unseen objects in contact-rich manipulation.

\textbf{LAPA}~\cite{LAPA} proposes an unsupervised latent action pretraining framework: a \textbf{VQ-VAE}~\cite{oord2017vqvae}-based latent action model learns ``state-latent action'' priors from unlabeled videos in a self-supervised manner, requiring only a small amount of real action annotations during downstream fine-tuning to map to actual joint actions, significantly reducing annotation demands. \textbf{villa-X}~\cite{chen2025villax} advances latent action modeling in two key aspects. First, a proprioceptive Forward Dynamics Model (proprio-FDM) is incorporated to ground latent actions in physical dynamics through joint optimization of visual reconstruction and proprioceptive prediction losses. Second, a joint diffusion framework comprising a latent action expert and a robot action expert is proposed, whereby latent actions explicitly condition low-level action generation for more structured knowledge transfer.

\subsection{Joint World-Action-Model}

\label{sec:joint_wam}
Joint World-Action Models denote a family of architectures in which future world states and actions are predicted within a single unified model, with both world modeling and action generation serving as joint supervision targets during training \cite{PAD,UWM,UVA,PhysGen,FLARE,FRAPPE,CosmosPolicy,Motus,Act2Goal,DUST,DiT4DiT,CoVAR}. 
The central design question for joint architectures is therefore how world state and action generation are coupled within the shared predictive system. Under this unified definition, existing joint world-action models can be further organized into two broad generation routes according to the substrate on which world--action prediction is realized：(1) \textbf{Autoregressive Generation} (\autoref{sec:joint_discerte}), in which future world variables and action variables are serialized into token space and modeled through autoregressive prediction; (2) \textbf{Diffusion-based Non-Autoregressive Generation} (\autoref{sec:joint_continuous}), in which future observations, latent world states, or action trajectories are generated through diffusion- or flow-based generative processes, enabling the joint refinement of both modalities within a continuous latent space or through parallel denoising streams.

\subsubsection{Joint Prediction via Autoregressive Generation}
\label{sec:joint_discerte}

\textbf{Autoregressive Generation} denotes the branch of joint world-action models that rely on causal, left-to-right sequential decoding to parameterize both future states and control signals \cite{GR1, grmg, GR2, CoTVLA, WorldVLA, rynnvla2, VLA-JEPA}. In these architectures, heterogeneous variables are serialized into a unified temporal sequence where the joint distribution of world and action is factorized sequentially. This fundamental generative mechanism ensures that earlier predictions causally condition subsequent steps within a given generation window.

However, while this shared reliance on \textbf{causal sequence modeling} unifies these architectures, casting control and video generation into a strict left-to-right prediction problem introduces profound architectural tensions. The primary challenge lies in mitigating catastrophic error propagation—where early visual hallucinations cascade into subsequent action failures—while balancing the inherent computational bottleneck of sequential decoding against the low-latency demands of real-time robotic execution.

Within this branch, we organize the literature by tracing the evolution of their underlying target representations and output interfaces. We identify three distinct representational paradigms: (1) \textbf{Explicit Decoupled Representation}, where modalities maintain heterogeneous formats and are decoded through structurally separate output heads; (2) \textbf{Unified Discrete Representations}, where modalities are fully quantized into a homogeneous token space and governed by a shared prediction head; and (3) \textbf{Predictive Latent Representations}, which abandon explicit token generation in favor of autoregressing over abstract, continuous latent spaces.

\begin{table}[t]
\centering
\scriptsize
\setlength{\tabcolsep}{3.5pt}
\renewcommand{\arraystretch}{1.08}
\caption{Taxonomy-oriented summary of \textbf{Autoregressive Generation} papers reviewed in this section. ``Params.'' reports the main backbone scale or explicitly stated module sizes. ``N/A'' indicates that a clean parameter figure is not explicitly stated in the paper.}
\label{tab:discrete_token_joint_prediction_summary}

\begin{adjustbox}{width=\linewidth}
\begin{threeparttable}
\begin{tabular}{
  >{\RaggedRight\arraybackslash}p{2.20cm}
  >{\RaggedRight\arraybackslash}p{1.60cm}
  >{\RaggedRight\arraybackslash}p{2.80cm}
  >{\RaggedRight\arraybackslash}p{4.20cm}
  >{\RaggedRight\arraybackslash}p{5.80cm}
}
\toprule
\textbf{Model} &
\textbf{Params.} &
\textbf{Backbone} &
\textbf{I/O Modality} &
\textbf{Evaluation} \\
\midrule

\rowcolor{gray!12}
\multicolumn{5}{l}{\textbf{Explicit Decoupled Representation}} \\
\midrule
GR-1 \cite{GR1}
  & 195M
  & GPT-style Causal Transformer
  & Obs, text, proprio $\rightarrow$ future patches, continuous actions
  & Sim: CALVIN \newline Real: Kinova Gen2 \\
GR-MG \cite{grmg}
  & N/A
  & GPT-style + InstructPix2Pix
  & Obs, text, history $\rightarrow$ goal image, task progress, actions
  & Sim: CALVIN \newline Real: Kinova Gen3 \\
GR-2 \cite{GR2}
  & 30-719M
  & GPT-style Causal Transformer
  & Obs, text, proprio $\rightarrow$ future VQ tokens, continuous actions
  & Sim: CALVIN \newline Real: Kinova Gen3 \\
\midrule

\rowcolor{gray!12}
\multicolumn{5}{l}{\textbf{Unified Discrete Representations}} \\
\midrule
CoT-VLA \cite{CoTVLA}
  & 7B
  & VILA-U
  & Obs, text $\rightarrow$ future visual tokens, discrete actions
  & Sim: LIBERO, Bridge-V2, Franka-Tabletop \newline Real: - \\
WorldVLA \cite{WorldVLA}
  & 7B
  & Chameleon-based MLLM
  & Obs, text $\rightarrow$ future VQ tokens, discrete actions
  & Sim: LIBERO \newline Real: - \\
RynnVLA-002 \cite{rynnvla2}
  & 5B
  & Chameleon + Action Head
  & Obs, text, state $\rightarrow$ future VQ tokens, continuous actions
  & Sim: LIBERO \newline Real: SO100 \\
$\mathcal{F}_{1}$ \cite{f1vla}
  & 4.2B
  & Mixture-of-Transformer (MoT)
  & Obs, text $\rightarrow$ future VQ tokens, continuous actions
  & Sim: LIBERO, SimplerEnv \newline Real: Genie-1, Franka, ARX LIFT II \\
\midrule

\rowcolor{gray!12}
\multicolumn{5}{l}{\textbf{Predictive Latent Representation}} \\
\midrule
VLA-JEPA \cite{VLA-JEPA}
  & 2B
  & Qwen3-VL-2B
  & Obs, text $\rightarrow$ future latents, continuous actions
  & Sim: LIBERO, SimplerEnv, LIBERO-Plus \newline Real: Franka \\

\bottomrule
\end{tabular}
\end{threeparttable}
\end{adjustbox}
\end{table}

\paragraph{\textbf{Explicit Decoupled Representation.}}

Early approaches to autoregressive joint prediction maintained strict modality separation at the representational level. Rather than forcing continuous physical dynamics and high-dimensional visual states into a single shared vocabulary, these architectures preserve their heterogeneous data formats. The central design principle here is representational decoupling: they rely on explicitly injected control tokens (e.g., [ACT], [OBS]) to route interleaved sequence features into structurally separate, task-specific output heads, typically pairing a continuous action decoder with a discrete visual patch-regression branch.

\textbf{GR-1} \cite{GR1} established this paradigm by demonstrating that a transformer pre-trained on video reconstruction could be fine-tuned to concurrently decode future visual patches and continuous actions via dual-branch heads. By explicitly enforcing the model to anticipate forthcoming visual events, the internalized video prediction acts as a potent regularizer for action generation. However, pure visual rollout can be brittle. To address this, \textbf{GR-MG} \cite{grmg} decoupled the world-rollout process into a macro/micro-step hierarchy. By introducing a [PROG] token and conditioning the policy on a diffusion-generated visual goal, GR-MG enables the language modality to guide execution even when intermediate pixel predictions fail. Scaling this foundational concept, \textbf{GR-2} \cite{GR2} transitioned to a fully discrete visual pipeline using VQGAN tokens. Crucially, GR-2 treats the visual rollout as an implicit planner, coupling discrete visual foresight with continuous action chunking parameterized by a CVAE. While these multi-head methods proved the feasibility of joint autoregressive prediction, their reliance on separate branches often exposed significant latency bottlenecks and limited the depth of cross-modal grounding.

\paragraph{\textbf{Unified Discrete Representations.}}

As general-purpose vision-language foundation models scaled, the representational focus shifted from decoupled routing toward deep integration within a homogeneous token space. In this paradigm, the heterogeneous nature of physical and visual modalities is entirely collapsed: continuous actions and high-dimensional images are fully quantized and mapped into a single, shared LLM vocabulary. Consequently, physical dynamics and visual states are represented as unified discrete symbols, generated by the exact same next-token prediction head. The primary challenge in this group shifts to designing attention and routing mechanisms that mitigate the severe compounding errors inherent in autoregressively sampling long strings of ungrounded action tokens.

Researchers have proposed distinct attention and routing mechanisms to solve this compounding error problem. For instance, employing hybrid attention routing, \textbf{CoT-VLA} \cite{CoTVLA} bifurcates the attention mechanism. It first uses causal attention to autoregressively hallucinate a discrete visual chain-of-thought (CoT), and then switches to a full-attention mechanism to synchronously predict a sequence of action tokens required to reach that generated visual state. Instead of splitting the generation phase, \textbf{WorldVLA} \cite{WorldVLA} relies on modality-specific causal masking. It operates strictly via interleaved prompt templates and modifies the standard causal mask during policy rollout, explicitly prohibiting current action tokens from attending to previously generated actions in the same chunk. This forces local predictions to ground entirely in the historical visual and linguistic context.

Alternatively, to mitigate the high variance of pure discrete control, \textbf{RynnVLA-002} \cite{rynnvla2} appends a lightweight continuous Action Transformer head to the discrete MLLM backbone, decoding continuous action chunks in parallel while preserving unified world modeling. Taking this hybridization further, $\mathcal{F}_{1}$ \cite{f1vla} employs a Mixture-of-Transformer (MoT) framework to decouple the predictive pathways. It introduces a Generation expert that autoregressively predicts discrete VQ tokens via next-scale prediction. Through progressive attention, the Action expert deduces control signals directly from the hallucinated visual foresight, effectively reformulating action generation as a foresight-guided inverse dynamics problem.

\paragraph{\textbf{Predictive Latent Representations.}} 

While autoregressing explicit visual tokens or decoupled patches provides high interpretability, it incurs immense computational overhead and is highly susceptible to ``pixel-matching shortcuts''—where the model expends capacity reconstructing nuisance background variations rather than learning actionable transition physics. To bypass this, \textbf{Predictive Latent Representations} shift the representational substrate entirely from explicit pixels to abstract, continuous latent embeddings. 

This paradigm is most prominently exemplified by the Joint-Embedding Predictive Architecture (JEPA) family. Specifically, \textbf{VLA-JEPA} \cite{VLA-JEPA} grounds joint sequence modeling strictly in this high-level latent space. It extracts continuous latent action tokens to conditionally guide an autoregressive world model, which predicts future representations encoded by a frozen target network. Because future frames are used solely as isolated supervision targets, the architecture remains structurally leakage-free. For physical execution, this abstract transition knowledge is bridged via an embodied action token that conditions a flow-matching head. By replacing explicit visual synthesis with latent transition alignment, this approach inherently prioritizes semantic abstraction and robustness over pixel-level reconstruction.

\subsubsection{Joint Prediction via Diffusion-based Generation}
\label{sec:joint_continuous}

\textbf{Diffusion-based Generation} constitute a significant technical route for joint world-action modeling, characterized by multi-step generative processes to capture the complex distributions of future states \cite{PAD,VideoVLA,DreamZero,UWM,FLARE,FRAPPE,CosmosPolicy,Motus,LingBotVA,LDA1B,CoVAR,AdaWorldPolicy,Act2Goal,DUST,DiT4DiT,UVA,PhysGen,udvla1,f1vla}. By leveraging generative frameworks such as diffusion processes or continuous flow-matching, these architectures generate future world states and action sequences concurrently across a multi-step horizon. This approach fundamentally overcomes the sequential bottleneck of autoregressive modeling, enabling the high-frequency execution necessary for closed-loop control.

\begin{figure}
    \centering
    \includegraphics[width=1\linewidth]{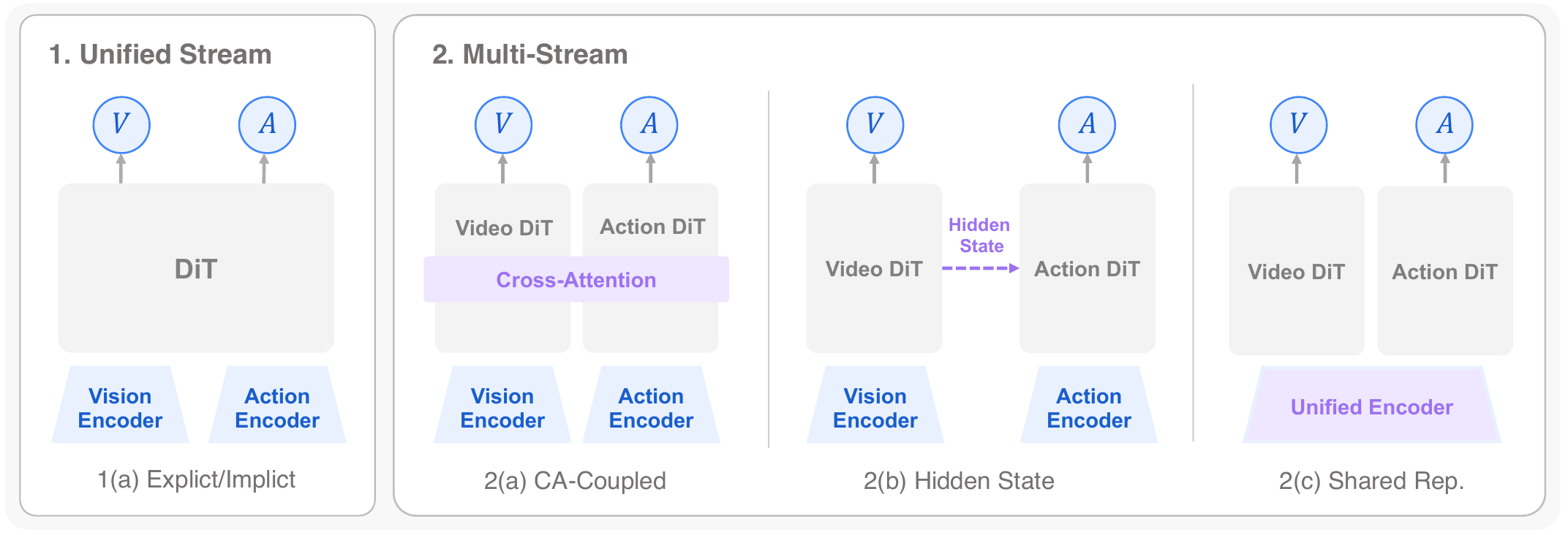}
    \caption{Taxonomy of the main architectural patterns in the diffusion-based joint WAMs. \textbf{1(a) Unified Stream}: World and action are integrated within one single DiT backbone, with world modeling realized either explicitly or implicitly. \textbf{2(a) Multi-Stream -- Cross-Attention Coupled}: separate video and action DiTs are coupled through explicit cross-attention. \textbf{2(b) Multi-Stream -- Hidden-State Coupling}: intermediate hidden states from the video DiT condition the action DiT. \textbf{2(c) Multi-Stream -- Shared Representation}: video and action are first fused through a unified encoder before being decoded into their respective outputs.}
    \label{fig:continu-diff-tax}
\end{figure}

To systematically characterize the diverse landscape of diffusion-based joint models, we organize our discussion based on the structural coupling of their predictive streams: (1) \textbf{Unified Stream} Architectures: These models integrate world and action variables into a single, homogeneous predictive trunk (e.g., a Single DiT). In this regime, world modeling and action generation are treated as a joint denoising task within a shared latent space, ensuring the tightest possible synchronization through unified attention mechanisms. (2) \textbf{Multi-Stream} Architectures: These designs distribute generation across coordinated branches or modality-specific experts. The interaction between the world and action branches is achieved through explicit coupling mechanisms, such as cross-attention, hidden-state conditioning, or shared encoders.
This backbone-level distinction provides the primary axis for understanding how world and action branches are structurally fused—whether through shared representations in a single stream or through modular interactions between multiple streams—to achieve coherent joint predictions. 
Fig.~\ref{fig:continu-diff-tax} provides a schematic overview of this category and its main sub-patterns.

\paragraph{\textbf{Unified Stream.}}
\label{sec:single-dit}
This family comprises structures in which the parallel generation of world and action variables is absorbed into a unified backbone (typically a single Diffusion Transformer) \cite{PAD,VideoVLA,DreamZero,UWM,FLARE,FRAPPE,CosmosPolicy}. Accordingly, world-related and action-related variables are processed within the same primary trunk, and method-specific variation is usually realized through the form of the prediction targets, the organization of auxiliary tokens or prefixes, or the way additional conditioning information is injected into the backbone, rather than through an explicit multi-branch decomposition. Within this family, we further distinguish between two recurrent routes, namely \textbf{Explicit Future Prediction} and \textbf{Implicit Future Prediction}.

\textcolor{openmossblue}{\textit{Explicit Future Prediction.}}
This pattern introduces future world supervision by making future observations or their explicit latent surrogates direct prediction targets of the model \cite{PAD,VideoVLA,DreamZero,UWM,CosmosPolicy}. In this family, world modeling is realized through the explicit generation of future images or targets from other modalities, so future-state prediction remains part of the main denoising objective throughout training. The differences across methods then lie primarily in the form of these future targets and in how they are organized within training and rollout.

The most straightforward instantiation jointly denoises concatenated sequences of future image latents and action tokens. 
\textbf{PAD} \cite{PAD} exemplifies this design, encoding multi-modal inputs—RGB images, robot pose, and depth—into a unified latent sequence and jointly predicting future frames and action chunks under a shared denoising objective. Training from scratch, PAD incorporates action-free internet video data during pretraining by extending the token sequence with action padding and applying attention masking accordingly. Ablations confirm that removing either future-image prediction or video co-training degrades control performance—demonstrating that explicit world modeling supervision meaningfully contributes to both action training quality and inference-time performance.
\textbf{VideoVLA} \cite{VideoVLA} shares the same joint denoising structure but takes a different starting point: rather than training from scratch, it repurposes the pretrained CogVideoX-5B \cite{CogVideoX} video diffusion backbone, treating future video latents and 7-DoF action chunks as a jointly denoised target sequence conditioned on the current frame and language instruction. Grounding in a pretrained video model obviates the need for action-free co-training, as rich visual dynamics are already encoded in the backbone.

A more flexible design is introduced by \textbf{UWM} \cite{UWM}, which decouples the diffusion schedules of world and action variables by assigning them separately controlled noise levels within one shared Transformer. This seemingly simple change has a significant functional consequence: by independently controlling each side's noise level at test time, one model can be steered into qualitatively different operating modes—policy inference, forward dynamics prediction, inverse dynamics, or pure video generation—without any architectural switching. It also provides an elegant solution for action-free video data, where the missing action side is simply fully noised and the standard denoising objective is applied unchanged.

Where UWM generalizes the coupling through independent noise schedules, \textbf{Cosmos Policy} \cite{CosmosPolicy} instead broadens \textit{what} is jointly predicted. Built on the Cosmos-Predict2 \cite{cosmos-predict2} video backbone, it introduces latent frame injection: proprioception, action chunks, future states, and predicted value functions are encoded as latent frames and interleaved with image latents in the same denoising sequence. This formulation allows a single checkpoint to serve simultaneously as a policy, a world model, and a value function, with planning via best-of-N sampling enabled simply by switching to autoregressive decoding at inference time.

\textbf{DreamZero} \cite{DreamZero} represents a representative instantiation of this design, building directly on the pretrained Wan2.1 \cite{Wan} image-to-video backbone with only lightweight additions—state/action encoders and an action decoder—while jointly denoising video latents and action latents under a shared objective. To maintain closed-loop conditioning without accumulating generative visual drift, it replaces imagined future frames with ground-truth observations after each executed chunk via KV-cache-based observation replacement. To bring inference latency within the range of real-time control, it further introduces a suite of system-level optimizations including asynchronous execution, DiT caching, quantization, and CUDA-graph compilation. 
\textbf{GigaWorld-Policy}~\cite{gigaworld-policy} adopts the same architectural design but adjusts the action generation attention to attend only to historical and current observations rather than imagined future frames, thereby eliminating the need to generate future video at inference time and achieving further acceleration. 
\textbf{X-WAM}~\cite{guo2026xwam} instead extends the prediction target by replicating the final DiT blocks as an interleaved depth branch, enabling explicit RGB-D joint modeling without disrupting pretrained visual priors, and employs Asynchronous Noise Sampling to align training and test-time noise schedules for efficient action dispatching.

Other works also explore discrete diffusion as an alternative to continuous denoising for joint future generation: \textbf{UD-VLA} \cite{udvla1} jointly denoises future image tokens and action tokens through iterative mask-and-predict steps over a unified discrete token space.

\textcolor{openmossblue}{\textit{Implicit Future Prediction.}}
This pattern incorporates future world supervision through implicit representations of future states rather than explicitly generating future observations \cite{FLARE, FRAPPE}. In this family, the policy still operates with a single action-generation backbone, but future information is introduced through auxiliary future tokens or future prefixes, whose intermediate representations are aligned to embeddings of future observations provided by teacher encoders or pretrained visual models. The supervision target is therefore not a reconstructed future frame or video sequence, but a compact future representation learned at the latent level. Compared with the preceding Explicit Future Prediction group, these methods place world modeling inside the policy network as an internal predictive constraint rather than as an additional generative output, while preserving a direct action-generation interface at inference time.

\textbf{FLARE} \cite{FLARE} establishes the core design: learnable future tokens are appended to the action-token sequence and propagated through the same DiT. At an internal layer, the activations of these future tokens are projected by an MLP and supervised to match the visual embeddings of the actual future observation encoded by a frozen teacher encoder—effectively compelling the policy's internal representation to anticipate what the world will look like after action execution. This alignment loss extends naturally to action-free video data, where it is applied alone without the action objective, allowing the policy to learn latent world dynamics from video-only demonstrations.
\textbf{FRAPPE} \cite{FRAPPE} develops this approach into a post-training recipe by introducing multiple alignment experts over a frozen RDT backbone. A Mixture-of-Prefix-and-LoRA formulation assigns each expert a different teacher representation, with a lightweight router aggregating their outputs. To stabilize multi-expert training, FRAPPE conducts a mid-training stage of full-parameter alignment to a single distilled teacher before expanding into the parallel parameter-efficient setting—demonstrating that implicit alignment can be modularly composed through staged training rather than requiring joint training from scratch.

\paragraph{\textbf{Multi-Stream.}} 
\label{sec:multi-dit}
The second family places world--action coupling at the level of persistent architectural decomposition.  In these methods, world modeling and action generation are no longer absorbed into one homogeneous predictive trunk; instead, the joint predictive computation is distributed across multiple coordinated branches, streams, experts, or shared-encoder--separate-decoder components whose interaction must itself be architecturally specified \cite{Motus,LingBotVA,LDA1B,CoVAR,AdaWorldPolicy,Act2Goal,DUST,DiT4DiT,UVA,PhysGen,AIM,MotuBrain,FastWAM,WAV,DexWorldModel}. The central architectural question in this family shifts from how to format data for a single engine, to how distinct computational pathways should exchange information while still functioning synchronously as one unified predictive system. We categorize these multi-engine architectures by their dominant coupling interface: (1) \textbf{Cross-Attention Coupling}, (2) \textbf{Hidden-State Coupling}, and (3) \textbf{Shared Representation}.

\textcolor{openmossblue}{\textit{Cross-Attention Coupling.}} In this family, world modeling and action generation are assigned to two structurally independent branches—a Video DiT and an Action DiT—with coupling realized through explicit cross-attention between them. Unlike Single DiT methods where world and action share the same DiT block, the interaction here is an explicit architectural design object: each branch maintains its own generative pathway while continuously exchanging information with the other through dedicated attention mechanisms.

Several works instantiate this dual-branch pattern while exploring different design choices in the coupling mechanism and world representation. 
\textbf{CoVAR} \cite{CoVAR} addresses this by introducing a dedicated Bridge Attention module that concatenates video and action features, performs joint attention, and splits the result back into separate streams. In this way, the video and action branches remain structurally separate while exchanging information repeatedly during generation.
\textbf{LDA-1B} \cite{LDA1B} takes a different route: rather than a dedicated coupling module, it absorbs both modalities into a shared MM-DiT attention layer with modality-specific projections, so the two streams remain distinct in representation but interact inside shared self-attention. For world representation, it predicts future states in a structured DINO latent space instead of video VAE latents.
Supervision is stratified by data quality: high-quality trajectories for policy and dynamics, lower-quality for dynamics and forecasting, and action-free videos for forecasting alone. To support these different modes within one architecture, LDA-1B introduces learnable task embeddings and modality-specific register tokens, allowing the same backbone to switch between policy, dynamics, or forecasting tasks without changing the network topology.
\textbf{DUST} \cite{DUST} adopts a similar MM-DiT structure but identifies a further degree of freedom: the two modalities need not share the same generative dynamics. It perturbs world and action with independent noise timesteps, optimizes them with separate flow-matching losses within a joint objective, and samples them under an asynchronous schedule at inference, decoupling their denoising trajectories while preserving cross-modal coupling within each block.

Rather than custom coupling modules, a subsequent line of work converges on Mixture-of-Transformers as a unified coupling substrate, where world and action latents are processed through modality-specific DiT pathways that interact via shared attention. Within this shared architectural foundation, the design question shifts from how to couple the two streams to what additional capability the coupled system should provide.
\textbf{LingBot-VA} \cite{LingBotVA} focuses on temporal grounding across rollout steps: it interleaves video latents and action tokens into an autoregressive sequence and pairs the MoT with a KV-cache that accumulates the full interleaved history across chunks, preserving causal consistency over sustained closed-loop execution. The deployment pipeline further introduces asynchronous prediction and execution, together with a feedback-grounded forward-dynamics update, so that newly received real observations can be incorporated before forecasting the next chunk. \textbf{DexWorldModel} \cite{DexWorldModel} follows the same autoregressive MoT structure but addresses its memory cost, replacing the growing KV-cache with a dual-state TTT memory that separates long-term observations from short-term predicted latents; it also shifts the world representation from RGB to DINOv3 semantic features, reducing sensitivity to task-irrelevant visual variation. \textbf{AIM} \cite{AIM} turns to the interface between the two branches: an intent-causal attention mask prevents action tokens from attending directly to future RGB tokens, so future information reaches the action head only through predicted value maps that localize task-relevant contact and placement regions. \textbf{Being-H0.7} \cite{Being-H0.7} addresses the training–inference asymmetry inherent in joint prediction: during training, the world branch explicitly generates future representations and distills them into latent queries via an alignment loss; at inference, the world branch is discarded and only the latent queries are retained, achieving competitive performance with substantially reduced inference cost.

The dual-stream MoT formulation admits a natural extension: rather than coupling only world and action, additional modality experts can be introduced into the shared attention, making the MoT a more general multi-expert coordination framework. \textbf{Motus}\cite{Motus} instantiates this by adding a semantic understanding expert derived from a pretrained VLM, so that world, action, and semantic understanding branches interact through Tri-modal Joint Attention; a UniDiffuser-style scheduler assigns independent noise levels per modality, enabling the unified architecture to operate across multiple functional modes including VLA, world model, and inverse dynamics within one backbone. \textbf{MotuBrain} \cite{MotuBrain} strengthens the language branch further by giving text tokens explicit participation in multimodal attention rather than treating language as an external conditioning signal; to balance coupling strength against efficiency, full joint attention is applied only in middle layers while outer layers remain partially decoupled in an H-bridge structure, and multiview inputs are supported by assigning view-dependent 3D RoPE offsets within a shared spatial coordinate space. \textbf{AdaWorldPolicy} \cite{AdaWorldPolicy} instead introduces a force predictor as the third expert, shifting the additional branch from semantic grounding to physical interaction modeling; all three experts are implemented as flow-matching DiTs built around a pretrained Cosmos-Predict2 backbone and connected through Multi-modal Self-Attention, with training alternating between action-generation and future-imagination modes, and at deployment, discrepancies between imagined and observed states and force readings serve as self-supervised signals for LoRA-based online adaptation.

\begin{table*}[!ht]
\centering
\scriptsize
\setlength{\tabcolsep}{3.5pt}
\renewcommand{\arraystretch}{1.08}
\caption{Taxonomy-oriented summary of \textbf{Diffusion-based Generation} papers reviewed in this section. ``Params.'' reports the main backbone scale or explicitly stated module sizes when such information is available in the paper or the local reading reports. ``N/A'' indicates that a clean parameter figure is not explicitly stated in the current evidence base.}
\label{tab:continuous_diffusion_joint_prediction_summary}

\begin{adjustbox}{width=\linewidth}
\begin{threeparttable}
\begin{tabular}{
  l
  >{\RaggedRight\arraybackslash}p{1.20cm}
  >{\RaggedRight\arraybackslash}p{3.00cm}
  >{\RaggedRight\arraybackslash}p{5.00cm}
  >{\RaggedRight\arraybackslash}p{5.0cm}
}
\toprule
\textbf{Model} &
\textbf{Params.} &
\textbf{Backbone} &
\textbf{I/O Modality} &
\textbf{Evaluation} \\
\midrule

\rowcolor{gray!12}
\multicolumn{5}{l}{\textbf{Unified Stream: Explicit Future Prediction}} \\
\midrule
PAD \cite{PAD}
  & N/A
  & ImageNet-initialized DiT
  & Obs, pose, optional depth, text $\rightarrow$ future image, depth, actions
  & Sim: MetaWorld \newline Real: Panda \\
VideoVLA \cite{VideoVLA}
  & 5B
  & CogVideoX-5B
  & Obs, text $\rightarrow$ future video latents, actions
  & Sim: SIMPLER \newline Real: Realman \\
UWM \cite{UWM}
  & N/A
  & Single DiT
  & Obs $\rightarrow$ future image latents, actions
  & Sim: LIBERO \newline Real: Franka \\
DreamZero \cite{DreamZero}
  & 14B
  & Wan2.1-I2V-14B-480P
  & Obs, text, proprio $\rightarrow$ future video frames, actions
  & Sim: PolaRiS, GenieSim3.0 \newline Real: AgiBot-G1, Franka \\
Cosmos Policy \cite{CosmosPolicy}
  & 2B
  & Cosmos-Predict2-2B
  & Obs, text, proprio $\rightarrow$ future video latents, actions, values
  & Sim: LIBERO, RoboCase \newline Real: ALOHA \\
GigaWorld-Policy \cite{gigaworld-policy}
  & 5B
  & Wan 2.2-5B
  & Obs, text, proprio $\rightarrow$ future video latents, continuous actions
  & Sim: RoboTwin 2.0 \newline Real: AgileX PiPER 6-DoF arm \\
X-WAM \cite{guo2026xwam}
  & 5B
  & Wan2.2-TI2V-5B
  & Obs, text, proprio $\rightarrow$ future RGB-D video, states, actions
  & Sim: RoboCasa, RoboTwin 2.0 \newline Real: AC One \\
\midrule

\rowcolor{gray!12}
\multicolumn{5}{l}{\textbf{Unified Stream: Implicit Future Prediction}} \\
\midrule
FLARE \cite{FLARE}
  & N/A
  & GR00T-style policy DiT
  & Obs, text, proprio $\rightarrow$ future latents, actions
  & Sim: RoboCasa, GR-1 tabletop tasks \newline Real: GR1 humanoid \\
FRAPPE \cite{FRAPPE}
  & 1B
  & RDT-1B
  & Obs, text, proprio $\rightarrow$ future representations, actions
  & Sim: RoboTwin 2.0 \newline Real: AgileX \\
\midrule

\rowcolor{gray!12}
\multicolumn{5}{l}{\textbf{Multi-Stream: Cross-Attention Coupled}} \\
\midrule
CoVAR \cite{CoVAR}
  & N/A
  & Open-Sora 1.2
  & Obs, text, proprio $\rightarrow$ future video frames, actions
  & Sim: CALVIN, Libero90 \newline Real: UR5 \\
LDA-1B \cite{LDA1B}
  & 1.6B
  & Qwen3-VL-4B-Instruct + DINOv3-ViT-s
  & Obs, text $\rightarrow$ future observation, actions
  & Sim: RoboCasa-GR1 \newline Real: Galbot G1, Unitree G1 \\
DUST \cite{DUST}
  & N/A
  & Eagle-2
  & Obs, text, proprio $\rightarrow$ future observation, actions
  & Sim: RoboCasa, GR-1 \newline Real: Franka Research 3 \\
LingBot-VA \cite{LingBotVA}
  & 5.3B
  & Wan2.2-5B
  & Obs, text, action history $\rightarrow$ future video latents, actions
  & Sim: Robotwin 2.0, LIBERO \newline Real: Franka \\
DexWorldModel \cite{DexWM}
  & N/A
  & DINOv3 + Wan2.2-5B
  & Obs, text, action history $\rightarrow$ future latent features, actions
  & Sim: RoboTwin \newline Real: Agilex CobotMagic \\
AIM \cite{AIM}
  & N/A
  & Wan2.2-TI2V-5B
  & Obs, text, action history $\rightarrow$ future RGB frames, spatial value maps, actions
  & Sim: RoboTwin 2.0 \newline Real: - \\
Motus \cite{Motus}
  & 8B
  & Wan2.2-2B + Qwen3-VL-2B
  & Obs, text $\rightarrow$ future video latents, actions
  & Sim: Robotwin 2.0, LIBERO-Long \newline Real: AC-One, Aloha-2 \\
MotuBrain \cite{MotuBrain}
  & N/A
  & Vidu
  & Obs, text $\rightarrow$ future video latents, actions
  & Sim: RoboTwin 2.0, WorldArena \newline Real: Humanoid platforms \\
AdaWorldPolicy \cite{AdaWorldPolicy}
  & 2.8B
  & Cosmos-Predict2
  & Obs, text, proprio $\rightarrow$ future video frames, actions, forces
  & Sim: LIBERO-10, Variant PushT, CALVIN \newline Real: INOVO \\
UD-VLA \cite{udvla1}
  & 8B
  & Emu3
  & Obs, text $\rightarrow$ future visual tokens, discrete actions
  & Sim: CALVIN, LIBERO, SimplerEnv \newline Real: UR5e \\
\midrule

\rowcolor{gray!12}
\multicolumn{5}{l}{\textbf{Multi-Stream: Hidden-State Coupled}} \\
\midrule
DiT4DiT \cite{DiT4DiT}
  & N/A
  & Cosmos-Predict2.5-2B
  & Obs, text, proprio $\rightarrow$ future video latents, actions
  & Sim: LIBERO, RoboCasa-GR1 \newline Real: Unitree G1 \\
Fast-WAM \cite{FastWAM}
  & 6B
  & Wan2.2-5B
  & Obs, text $\rightarrow$ latent world features, actions
  & Sim: RoboTwin 2.0, LIBERO \newline Real: Galaxea R1 Lite \\
WAV \cite{WAV}
  & 2.2B
  & Genie Envisioner
  & Obs, text $\rightarrow$ future visual trajectories, trajectory values, actions
  & Sim: LIBERO \newline Real: Piper \\
Act2Goal \cite{Act2Goal}
  & 1.76B
  & Genie Envisioner
  & Obs, goal image, proprio $\rightarrow$ future video latents, actions
  & Sim: Robotwin 2.0 \newline Real: AgiBot Genie-01 \\
\midrule

\rowcolor{gray!12}
\multicolumn{5}{l}{\textbf{Multi-Stream: Shared Representation}} \\
\midrule
UVA \cite{UVA}
  & 0.5B
  & Shared Transformer
  & Obs, optional text, action history $\rightarrow$ future image latents, actions
  & Sim: PushT/PushT-M, ToolHang, LIBERO-10 \newline Real: ARX X5 \\
PhysGen \cite{PhysGen}
  & 0.73B
  & NOVA
  & Obs, text, action history $\rightarrow$ future video frames, actions
  & Sim: LIBERO, ManiSkill \newline Real: Franka Panda \\
\bottomrule
\end{tabular}
\end{threeparttable}
\end{adjustbox}
\end{table*}

\textcolor{openmossblue}{\textit{Hidden-State Coupling.}}
The second \textbf{Multi DiT} pattern couples world modeling and action generation through intermediate representations passed from one branch to the other. In this family, the two components are still assigned to separate backbones, but the central interaction is no longer sustained branch-level co-generation through repeated cross-branch exchange. Instead, the world branch produces temporally informative internal representations—such as imagined latent trajectories, multi-scale transition features, or denoising hidden states—which are then used by the action branch as conditioning signals for control prediction \cite{Act2Goal,DiT4DiT}. Compared with \textbf{Cross-Attention Coupled Branches}, the coupling here is organized around the transfer of internal states from the world model to the action model, rather than continuous mutual interaction between parallel branches throughout generation. 

\textbf{DiT4DiT }\cite{DiT4DiT} instantiates this through a paired Video DiT–Action DiT architecture, where the action branch is conditioned on intermediate hidden states extracted from the video branch during future-frame denoising. A hook operator intercepts hidden activations at a selected feature-extraction timestep, either from a specific transformer block or aggregated across layers, and these are passed to the Action DiT via cross-attention with proprioceptive state embeddings and noisy action tokens. To coordinate the two modules, a tri-timestep design samples the video prediction timestep and action timestep independently while fixing the hidden-state extraction timestep to stabilize the conditioning signal; at inference, action prediction likewise relies on a single deterministic extraction step rather than the full video sampling loop. \textbf{Fast-WAM} \cite{FastWAM} pursues a related design but with inference efficiency as the primary concern: during training it retains the video branch as a future-video flow-matching objective within a Video DiT–Action DiT MoT, using a structured mask to prevent action tokens from attending to future video latents; at inference, the future-video branch is removed entirely, and the Video DiT encodes only the current visual context with a single forward pass whose latent world features condition action denoising. The hidden-state interface is thus preserved as a training-time coupling mechanism while its inference-time cost is eliminated.

\textbf{WAV} \cite{WAV} extends this one-way world-to-action interface by introducing an explicit trajectory-value branch between latent future prediction and action decoding. Its video module generates latent future visual rollouts, a value module evaluates these rollouts through cross-attention, and the action decoder conditions on both visual rollout features and value embeddings. At inference, this interface is turned into implicit planning by iteratively updating the video and value noise distributions with top-scoring samples, so actions are decoded from value-shaped latent futures without an external planner.

\textbf{Act2Goal} \cite{Act2Goal} organizes hidden-state coupling around a different objective: rather than improving the fidelity or efficiency of the transferred representation, it uses the coupling to support long-horizon goal-directed behavior. The world branch takes the current observation and a target goal image and generates a sequence of intermediate latent frames representing visual progress toward the goal; the action branch then predicts actions conditioned on multi-scale temporal features from these imagined transitions through cross-attention. A central component is Multi-Scale Temporal Hashing, which organizes both predicted visual states and actions into a proximal–distal temporal structure: proximal frames and actions are kept dense for short-horizon closed-loop control, while distal frames and actions are sparsified with logarithmically increasing spacing to preserve long-horizon goal consistency. Beyond offline training under a joint visual-and-action flow-matching objective, Act2Goal further supports reward-free online improvement through HER-style goal relabeling and LoRA-based finetuning.

\textcolor{openmossblue}{\textit{Shared Representation / Unified Encoder.}} Another \textbf{Multi DiT} pattern places world--action integration in a shared representation space or unified encoder, rather than in persistent branch-level interaction or in the transfer of intermediate hidden states from one branch to the other. In this family, visual observations and actions are first fused into a common latent substrate and jointly processed there, while modality-specific decoders or output heads subsequently recover the world-side and action-side predictions, so the key architectural distinction lies in performing joint modeling at the level of a unified representation and postponing modality specialization to later decoding stages \cite{UVA,PhysGen}.

\textbf{UVA} \cite{UVA} encodes historical observations and action chunks together with masked future observation tokens into a shared Transformer backbone, organizing future video and action information into a unified latent over the prediction horizon. These shared latent tokens are then decoded by two lightweight diffusion heads: a video diffusion head reconstructs future observations token by token, while an action diffusion head aggregates the latent tokens of each future step and predicts the corresponding action chunk. The two decoding pathways are trained jointly but remain decoupled at inference, so policy execution can bypass video generation and decode actions directly from the shared latent. UVA further adopts a masked training scheme with flexible input–output configurations, where unused components are replaced by learned mask tokens and losses are applied according to the selected objective, allowing the same backbone to operate as a policy model, video model, forward or inverse dynamics model, or combined policy–planner by masking different components rather than switching through task embeddings or modality-wise scheduling. \textbf{PhysGen} \cite{PhysGen} adopts the same shared-representation logic within a continuous autoregressive framework built on a pretrained autoregressive video backbone (NOVA \cite{NOVA}). It tokenizes observations and action chunks into frame tokens and action tokens, concatenates them into shared physical tokens, and models joint future dynamics through a unified Causal Transformer. The resulting tokens are decoded through separate pathways: frame tokens are reconstructed by the inherited video de-tokenizer, while action tokens are decoded by a lightweight Action-DiT conditioned on the autoregressive context through cross-attention. A causal masking scheme further allows action tokens to attend unidirectionally to frame tokens so that predicted future visual states inform action generation, and Lookahead Multi-Token Prediction generates multiple future action tokens in parallel at each autoregressive step while executing only the first at inference.

\section{Training data}
\label{sec:data}

\begin{wrapfigure}{r}{0.5\textwidth}
    \centering
    \vspace{-0.5cm}
    \includegraphics[width=\linewidth]{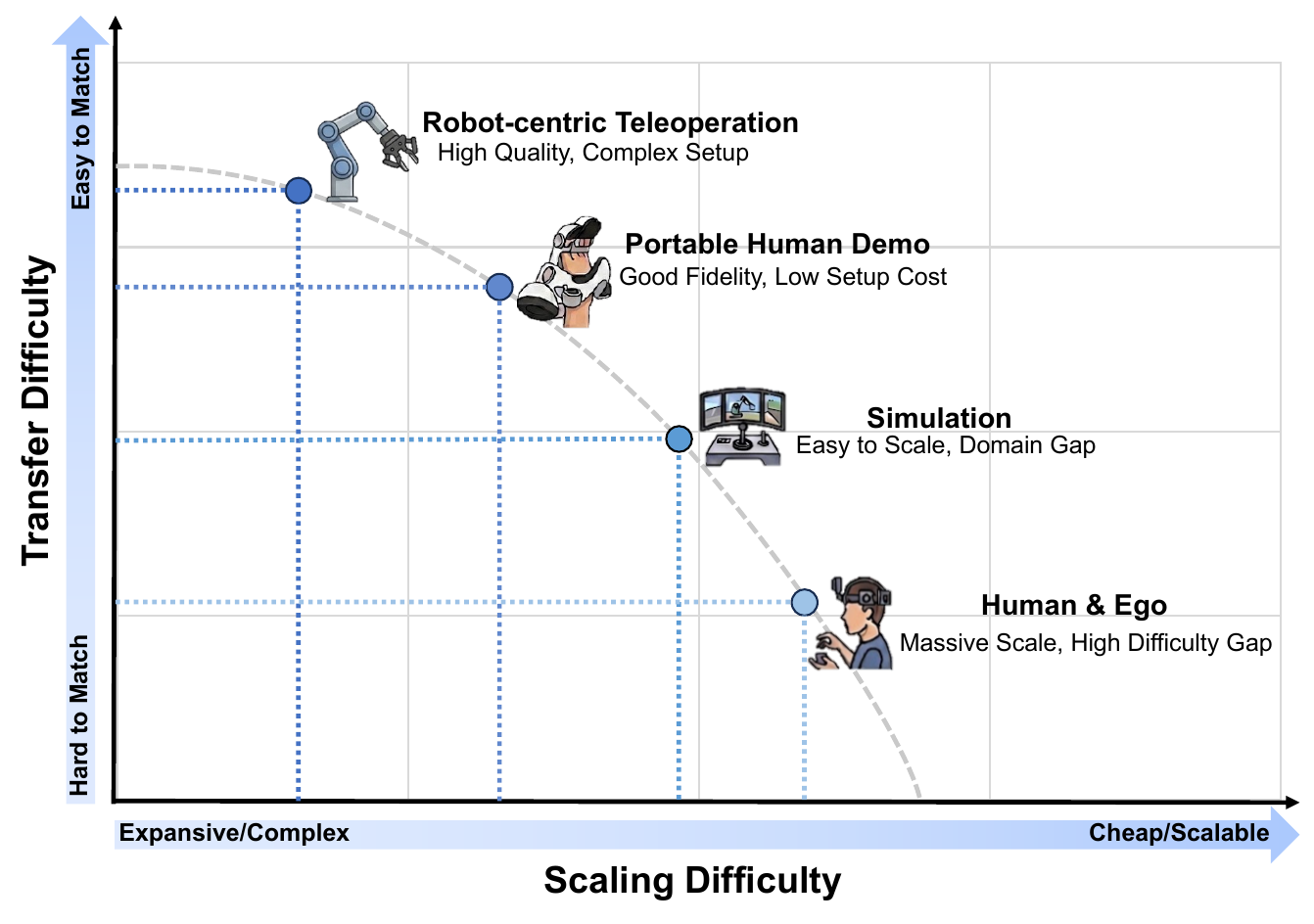}
    \vspace{-0.5cm}
    \caption{An overview of the embodied data landscape for training World Action Models, mapped across Transfer Difficulty (Y-axis) and Scaling Difficulty (X-axis).}
    \label{fig:data_overview}
    \vspace{-1cm}
\end{wrapfigure}

Training robust and generalizable World Action Models (WAMs) is fundamentally bottlenecked by the availability and quality of embodied data. Unlike large language models that thrive on passively scraped internet text, WAMs require strict physical grounding to capture complex state transitions, action conditioning, and intuitive physics. More importantly, the data requirements for WAMs diverge significantly from traditional robotic paradigms. Standard Vision-Language-Action (VLA) models strictly require paired $(o_t, a_t)$ trajectories, which severely restricts scalability given the high cost and scarcity of teleoperated demonstrations. Conversely, pure World Models thrive on action-free $(o_t, o_{t+1})$ sequences from internet videos but lack grounding in physical control.

The unique advantage of WAMs lies in their \textbf{unified data digestion}. They uniquely benefit from the intersection of these domains, leveraging high-quality $(o_t, a_t, o_{t+1})$ triplets to tightly couple their internal representations, while simultaneously possessing the architectural flexibility to ingest massive \textit{unpaired} data (e.g., action-free videos for visual physics) through joint training strategies. Consequently, constructing the data landscape for WAMs is not merely about scaling robot-centric data; it requires strategically mixing strictly coupled demonstrations with unconstrained observations.

This section provides a comprehensive overview of the four dominant data paradigms driving WAM training: (1) high-fidelity \textbf{Robot-Centric Teleoperation} (\autoref{sec:data_robot}), which provides exact kinematic grounding; (2) agile \textbf{Portable Human Demonstrations} (\autoref{sec:data_umi}), which bridge human dexterity with real-world interaction (such as UMI data); (3) highly scalable \textbf{Simulation Data} (\autoref{sec:data_sim}), which offers infinite procedural variations and privileged spatial supervision; and (4) broad-coverage \textbf{Human and Ego-Centric Data} (\autoref{sec:data_human}), which supplies near-unbounded priors for passive world dynamics. By navigating the inherent trade-offs mapped in Figure \ref{fig:data_overview}---ranging from the complex, rigid setups of robot-centric teleoperation to the cheap, massively scalable nature of internet videos---researchers are increasingly leveraging a complementary mixture of these datasets to bridge the gap between precise, low-level robotic control and broad, open-world generalization.

\begin{table}[!ht]
    \centering
    \footnotesize
    \scriptsize
    \caption{Summary of \textbf{robot-centric manipulation datasets} in this section. ``Modality'' denotes observation modalities: RGB (visual observations), P (proprioception), D (depth), A (audio), PC (point clouds), and T (tactile).}
    \label{tab: robot_data}
    \vspace{-6pt}
    
    \renewcommand{\arraystretch}{1.4}
    \renewcommand\tabcolsep{4pt}
    
    \begin{tabular}{l p{0.7cm} p{1.8cm} p{1.7cm} p{1.2cm} p{1.2cm} p{1.5cm} p{1.7cm} p{1.5cm}}
    \toprule
    \textbf{Name} & \textbf{Year} & \textbf{Scale} & \textbf{Skill} & \textbf{Task} & \textbf{Env} & \textbf{Embodiment} & \textbf{Collection} & \textbf{Modality} \\
    \midrule

    \textit{QT-Opt} \cite{QT-Opt} & 2018 & 580k traj. 
    & 1 & - & - & KUKA LBR iiwa & script & RGB \\

    \textit{MIME}\cite{MIME} & 2018 & $8,260$ traj. 
    & - & 20 & - & Baxter & visual demonstration & RGB, D, P \\
    
    \textit{RoboNet} \cite{RoboNet}& 2019 & $\sim$162k traj. 
    & - & - & - & 7 robots & script & RGB, P \\

    \textit{RoboTurk-Real} \cite{RoboTurk} & 2019 & $2,144$ traj. 
    & - & 3 & - & Sawyer & teleop. & RGB, D, P \\
    
    \textit{BridgeData} \cite{Bridge} & 2021 & 7.2k traj. 
    & 4 & 71 & 10 & WidowX250s & teleop. & RGB, text \\
    
    \textit{MT-Opt} \cite{MT-Opt} & 2021 & 800k traj. 
    & 2 & 12 & 3 & 7 KUKA IIWA robots  & script & RGB, text \\

    \textit{BC-Z} \cite{BC-Z}& 2021 & 25.9k traj. 
    & 9 & 129 & - & Everyday Robots & teleop. & RGB, text \\

    \textit{RT-1} \cite{RT-1} & 2022 & 130k+ traj. 
    & - & 700+ & - & Everyday Robots & teleop. & RGB, text \\

    \textit{TACO-RL Real-World} \cite{TACO-RL-Real-World} & 2022 & $\sim15$ h 
    & - & 25 & - & Franka & teleop. & RGB, P \\

    \textit{Language-Table} \cite{Language-Table}& 2022 & 594k traj. 
    & 1 & 5 & - & xArm6 & teleop. & RGB, text \\
    
    \textit{BridgeData v2} \cite{BridgeData_V2}& 2023 & 60k+ traj. 
    & 13 & - & 24 & WidowX 250 & teleop., script & RGB, D, text \\

    \textit{Jaco Play} \cite{Jaco-Play}& 2023 & $1,085$ traj. 
    & pick-and-place & - & - & Jaco 2 & teleop. & RGB, P, text \\

    \textit{VIMA-BENCH} \cite{VIMA-Bench}& 2023 & 650k+ traj. 
    & pick-and-place, push/wipe & 17 & - & UR5 & script & RGB, BB \\

    \textit{Cable Routing Dataset} \cite{Cable-Routing-Dataset}& 2023 & $1,699$ traj. 
    & - & - & 1 & Franka & teleop. & RGB, P \\

    \textit{RH20T} \cite{RH20T}& 2023 & 110k traj. 
    & 42 & 147 & - & 4 robots & teleop. & RGB, D, text \\
    
    \textit{OXE} \cite{OXE}& 2023 & 1M +traj. 
    & 527 & 160k+ & 60 & 22 robots & teleop., script, manual & RGB, D, text, PC \\

    \textit{Berkeley UR5} \cite{BerkeleyUR5Website}& 2023 & 896 traj. 
    & pick-and-place, sweeping, stacking & 4 & - & UR5 & demonstration & RGB, D, P, A, text \\

    \textit{TOTO} \cite{TOTO}& 2023 & $2,898$ traj. 
    & scooping, pouring & 2 & - & Franka & teleop. & RGB, P \\

    \textit{Grasp-Anything} \cite{Grasp-Anything}& 2023 & 1M+ traj. 
    & grasp detection & grasp detection & - & KUKA robot + Robotiq 2F-85 gripper & generative & RGB, text \\
    
    \textit{RoboSet} \cite{RoboSet}& 2023 & $7,500$ traj. 
    & 12 & 38 & 10 & Franka & teleop. & RGB, D, P \\
    
    \textit{Robo360} \cite{Robo360}& 2023 & 2k traj. 
    & - & - & - & xArm6 & teleop. & RGB, D, P \\
    
    \textit{DROID} \cite{DROID}& 2024 & 76k traj. 
    & 86 & 86 & 564 & Franka & teleop. & RGB, D, text \\
    
    \textit{RH20T-P} \cite{RH20T-P}& 2024 & 38k traj. 
    & 10+ & 67 & - & 4 robots & teleop. & RGB, D, text \\
    
    \textit{RoboMIND} \cite{RoboMIND}& 2024 & 107k traj. 
    & 38 & 479 & - & 4 robots & teleop. & RGB, P, text \\
    
    \textit{ARIO} \cite{ARIO}& 2024 & 3M+ traj. 
    & 345 & 321k &378 & 35 robots & aggregation, teleop. & RGB, D, A, T, text \\

    \textit{RoboData} \cite{RoboData}& 2024 & 70k traj. 
    & - & - & - & Franka, Google Robot & teleop., script & RGB, D, P, text \\
    
    \textit{DexCap} \cite{DexCap}& 2024 & 787 traj. 
    & - & - & 10+ & Franka & motion capture & RGB, D, PC, P \\
    
    \textit{FuSe} \cite{FuSe}& 2025 & 27k traj. 
    & 2 & 3 & - & WidowX 250 & teleop. & RGB, T, A, P, text \\

    \textit{Interleave-VLA} \cite{Interleave-VLA}& 2025 & 210k+ traj. 
    & - & - & - & None & automatic-annotation & RGB, P, text \\

    \textit{UniVoxGen Dataset} \cite{UniVoxGen-Dataset}& 2025 & 50k traj. 
    & - & - & - & Flexiv Rizon 4S & synthetic generation & RGB, D, 3D voxel \\

    \textit{AgiBot World} \cite{AgiBot_World_Colosseo}& 2025 & 1M+ traj. 
    & 87 & 217 & 100+ & AgiBot G1 & teleop. & RGB, D, T, P, text \\

    \textit{REASSEMBLE} \cite{REASSEMBLE}& 2025 & $4,551$ traj. 
    & 9 & 4 & - & Franka & teleop. & RGB, P \\
    
    \textit{OmniAction} \cite{OmniAction}& 2025 & 141k traj. 
    & 112 & 748 & 640 & WidowX 250S & sim., aggregation, generative & RGB, A, text \\

    \textit{UnifoLM-WBT} \cite{UnifoLM-WBT}& 2026 & $1,892,118$ traj. 
    & - & - & - & Unitree G1 humanoid & LeRobot framework & RGB, P \\
    
    \bottomrule
    \end{tabular}
    
    \vspace{-6pt}
\end{table}

\subsection{Robot-Centric Teleoperation Data}
\label{sec:data_robot}
Teleoperation remains one of the most reliable and prevalent paradigms for acquiring high-quality expert trajectory data in embodied AI. In this setup, a human operator manipulates a robotic agent via a remote control interface, recording continuous sensory observations, proprioceptive states, and corresponding executable actions. For the training of World Action Models, robot-centric datasets play an irreplaceable role: they provide the strictly aligned, high-frequency action-state pairs required to learn precise, action-conditioned physical dynamics with virtually zero sim-to-real gap. As robotic hardware and learning algorithms have co-evolved, the construction of these datasets has advanced along two highly strategic trajectories: scaling toward open-world diversity, and deepening the physical grounding of sensory perception.

\subsubsection{Scaling Up: Embodiment, Diversity, and Automated Augmentation.} 
The first major trajectory aims to shatter the closed-world assumption by drastically increasing dataset scale, environmental variations, and embodiment coverage. Early milestones, such as \textbf{QT-Opt} \cite{QT-Opt} and \textbf{MT-Opt} \cite{MT-Opt}, successfully demonstrated the viability of collecting massive, self-supervised trajectory data. Concurrently, pioneering works like \textbf{RoboNet} \cite{RoboNet} and \textbf{MIME} \cite{MIME} took the crucial first steps toward cross-robot and cross-domain generalization. However, these early collections were largely confined to isolated laboratory environments and short-horizon, primitive skills.

To bridge the gap to true open-world deployment, subsequent efforts deliberately targeted semantic diversity and sequential reasoning capabilities. Datasets like \textbf{BridgeData} \cite{Bridge}, \textbf{BC-Z} \cite{BC-Z}, \textbf{RT-1} \cite{RT-1}, and \textbf{Language-Table} \cite{Language-Table} shifted the focus toward language-conditioned tasks in varied household settings. The inclusion of language instructions is vital for WAMs, as it provides semantic anchors that help the model ground visual state transitions into generalized human concepts. Meanwhile, datasets such as \textbf{TACO-RL} \cite{TACO-RL-Real-World} and \textbf{RH20T-P} \cite{RH20T-P} emphasized long-horizon reasoning and hierarchical task decomposition, forcing models to predict extended sequences of future states. 

This scaling trend eventually culminated in massive, multi-platform aggregations. The \textbf{Open-X Embodiment (OXE)} \cite{OXE} harmonized over 1 million trajectories across 22 robots, a unified strategy pushed even further by \textbf{ARIO} \cite{ARIO} and \textbf{RoboMIND} \cite{RoboMIND}. Alongside explicit aggregation, in-the-wild massive collections like \textbf{DROID} \cite{DROID} significantly expanded scene diversity, while the recent \textbf{UnifoLM-WBT} \cite{UnifoLM-WBT} extends this paradigm to high-DoF humanoid embodiments. For WAMs, this extreme morphological diversity is transformative: it forces the world model to decouple universal physical laws from specific robot kinematics, enabling robust, embodiment-agnostic generalization.

Crucially, to bypass the inherent physical labor and annotation bottlenecks of manual, goal-directed teleoperation, this scaling trajectory has increasingly embraced alternative data acquisition mechanisms. One approach is the utilization of unstructured, play-based data sources. \textbf{Jaco Play} \cite{Jaco-Play} demonstrates how continuous, task-agnostic interaction priors can be collected without rigid task definitions, providing WAMs with broad exploratory knowledge of how objects react to arbitrary forces. 

In parallel, generative and automated augmentation have emerged as powerful scaling engines. Works like \textbf{Grasp-Anything} \cite{Grasp-Anything} leverage vision foundation models to generatively scale grasp data to the million-trajectory level, while \textbf{Interleave-VLA} \cite{Interleave-VLA} radically expands the supervision space through automated image-text interleaving. In the 3D domain, the \textbf{UniVoxGen Dataset} \cite{UniVoxGen-Dataset} scales spatial reasoning through massive synthetic voxel-based generation. By combining multi-robot aggregation with unscripted exploration and automated generation, the field has established a sustainable pathway to saturate WAMs with near-infinite training data, bypassing the strict limits of human collection.

\subsubsection{Deepening Perception: Multimodal and Contact-Rich Grounding} 
Parallel to the expansion in scale, a second critical trajectory has emerged to address the partial-observability bottleneck of purely visual (RGB) perception. In complex physical environments, many state transitions are visually ambiguous or occluded. To capture subtle physical interactions, recent datasets have systematically integrated denser multimodal signals. \textbf{Berkeley UR5} \cite{BerkeleyUR5Website} and \textbf{OmniAction} \cite{OmniAction} introduced audio modalities, capturing acoustic signatures essential for material identification and impact verification. Furthermore, modeling the dynamics of complex materials remains a notorious challenge; datasets like \textbf{TOTO} \cite{TOTO} specifically address this by focusing on visually ambiguous tasks such as pouring liquids and interacting with transparent objects, while the \textbf{Cable Routing Dataset} \cite{Cable-Routing-Dataset} and \textbf{REASSEMBLE} \cite{REASSEMBLE} target complex deformable manipulations and tight-tolerance physical assembly.

To further combat occlusion and improve spatial understanding, an increasing emphasis has been placed on 3D geometry. Datasets like \textbf{RH20T} \cite{RH20T}, \textbf{Robo360} \cite{Robo360}, and \textbf{RoboData} \cite{RoboData} emphasized dense multi-view capture and calibrated 3D point cloud alignment, ensuring that policies can learn robust physical representations independent of a single camera angle.
More recently, the frontier has pushed heavily toward high-fidelity tactile feedback and dexterous supervision. \textbf{RoboSet} \cite{RoboSet} emphasized forcefully constrained manipulations with articulated objects, while \textbf{DexCap} \cite{DexCap} captures fine-grained dexterous human hand motions spatially aligned with robot observations. Concurrently, \textbf{FuSe} \cite{FuSe} and \textbf{AgiBot World} \cite{AgiBot_World_Colosseo} integrate explicit visuotactile sensors alongside dynamic bimanual coordination. These multimodal datasets are absolutely foundational for WAMs. They allow models to internalize intuitive, lower-level physics—such as contact wrenches, sliding friction, mass distribution, and local deformations—that cannot be mathematically derived from 2D visual appearance alone, bridging the critical gap between high-level reasoning and physical execution.

\subsection{Portable Human Demonstration Data (UMI-style)}
\label{sec:data_umi}
Despite the high quality of robot-centric teleoperation datasets, they face persistent bottlenecks: high data collection costs, confinement to restricted laboratory settings, and limited embodiment diversity. These constraints restrict the exposure of models to the broad, unconstrained world dynamics necessary for WAMs. To overcome this, researchers have rethought the data collection interface, moving toward portable, low-cost human demonstration paradigms that bridge the gap between structured robot teleoperation and diverse internet videos.

The \textbf{Universal Manipulation Interface (UMI)} \cite{UMI} represents a milestone in this direction. By using a lightweight, handheld 3D-printed gripper combined with wearable cameras, UMI enables non-expert users to collect manipulation trajectories directly in everyday, in-the-wild environments. Through vision-based tracking and retargeting, human demonstrations are aligned into robot-executable actions.

Beyond foundational hardware developments, the UMI ecosystem has rapidly evolved to expand the sensory modalities and task complexity of collected data. Works like \textbf{FastUMI} \cite{FastUMI} optimized the pipeline for scalable, massive dataset construction. Subsequent extensions have systematically augmented the collected data by incorporating active perception \cite{ActiveUMI}, critical tactile feedback \cite{exUMI, Tactile-Conditioned_Diffusion_Policy}, and multi-view visual observations \cite{MV-UMI}. 
Furthermore, the paradigm has scaled beyond simple parallel-jaw grippers to capture high-DoF dexterous manipulation \cite{DexUMI} and whole-body mobile coordination \cite{UMI_on_Legs, HoMMI}. For WAM training, this evolution ensures that portable demonstration data can provide the multimodal, contact-rich, and complex action trajectories required to model real-world physics.

Consequently, UMI-style datasets have quickly evolved from small proof-of-concept collections to massive, wild-collected corpora, as summarized in Table~\ref{tab:UMI_table}. Early releases primarily validated the hardware paradigm, but recent efforts provide unparalleled resources for generalist policy learning. \textbf{FastUMI-100K} \cite{FastUMI-100K} pushes the trajectory count over 100K with enriched multimodal text annotations. \textbf{RealOmin} \cite{RealOmin} represents a qualitative leap, introducing a million-scale dataset collected across over 3,000 diverse household environments, capturing rich proprioceptive, IMU, and tactile signals. Similarly, \textbf{Hoi!} \cite{Hoi!} explicitly targets cross-view and force-aware manipulation, which is vital for WAMs to learn intuitive physics and contact dynamics. 
In particular, datasets such as the one proposed alongside \textbf{RDT2} \cite{RDT2} offer roughly 10,000 hours of demonstrations in hundreds of real-world scenes. 

For the advancement of World Action Models, these large-scale portable datasets act as a critical bridge. They possess the immense environmental and contextual diversity typical of egocentric human videos, yet they are strictly paired with high-frequency, centimeter-level action constraints. This unique combination allows WAMs to learn highly robust, action-conditioned state transition dynamics across unconstrained physical environments—a capability that is nearly impossible to acquire from traditional robot-centric datasets alone.

\begin{table}[t]
    \centering
    \scriptsize
    \setlength{\tabcolsep}{3.5pt}
    \renewcommand{\arraystretch}{1.08}
    \caption{Summary of \textbf{UMI datasets}. ``Modality'' denotes observation modalities: RGB, P (proprioception), D (depth), IMU, T (tactile), text.}
    \label{tab:UMI_table}
    \vspace{-6pt}
    
    \begin{adjustbox}{width=\columnwidth}
    \begin{threeparttable}
    \begin{tabular}{
    >{\RaggedRight\arraybackslash}p{2.65cm}
    c
    >{\RaggedRight\arraybackslash}p{1.85cm}
    >{\RaggedRight\arraybackslash}p{1.10cm}
    >{\RaggedRight\arraybackslash}p{1.25cm}
    >{\RaggedRight\arraybackslash}p{2.30cm}
    >{\RaggedRight\arraybackslash}p{3.00cm}
    >{\RaggedRight\arraybackslash}p{3.10cm}
    }
    \toprule
    \textbf{Name} & \textbf{Year} & \textbf{Scale} & \textbf{Task} & \textbf{Env} & \textbf{Embodiment} & \textbf{Collection} & \textbf{Modality} \\
    \midrule
    
    FastUMI-Data \cite{FastUMI} & 2024 & 10K+ traj. 
    & 22 & - & xArm 6 & human demonstrators & RGB, P, text \\
    
    FastUMI-100K \cite{FastUMI-100K} & 2025 & 100K+ traj. 
    & 54 & 5 & xArm6, Flexiv Rizon4 & human demonstration & RGB, P, text \\
    
    RealOmin \cite{RealOmin} & 2025 & 1M traj. 
    & 30 & 3000+ & - & human demonstration & RGB, P, IMU, T, text \\
    
    Hoi! \cite{Hoi!} & 2025 & $3,048$ traj. 
    & - & 38 & - & human demonstration, teleop. & RGB, D, P, T \\

    RDT2 \cite{RDT2} & 2026 & $10,000$ hours 
    & 52+ & 100+ & - & human demonstration & RGB, D, P \\
    
    \bottomrule
    \end{tabular}
    \end{threeparttable}
    \end{adjustbox}
    
    \vspace{-6pt}
\end{table}

\subsection{Simulation Data}
\label{sec:data_sim}
Simulation provides a scalable, fully controllable, and deterministic alternative to real-world data collection. For the training of World Action Models (WAMs), the fundamental value of simulation extends far beyond cost reduction. A physics engine is, essentially, an exact computational world model. Unlike real-world datasets that suffer from severe partial observability and sensory noise, simulation environments provide \textit{privileged information}—perfect depth, exact 6D object poses, precise collision boundaries, and unoccluded multi-view states. By training on simulation data, WAMs are not merely copying trajectories; they are actively distilling the fundamental laws of intuitive physics directly from the rendering and physics engines (e.g., MuJoCo \cite{Mujoco}, Isaac Sim \cite{IssacSim}, SAPIEN \cite{SAPIEN}). The evolution of these datasets reflects a transition toward massive procedural scaling and increasingly high-fidelity spatiotemporal and contact-rich supervision.

\subsubsection{Scaling Up: Procedural Generation and Environmental Complexity} 
The most immediate advantage of simulation is the ability to bypass the physical labor bottleneck of teleoperation, enabling the simultaneous scaling of data quantity, environmental diversity, and task complexity. Early foundational frameworks like \textbf{ManiSkill2} \cite{ManiSkill2} provided unified, high-performance benchmarking for manipulation. Building on this, pioneering works such as \textbf{MimicGen} \cite{MimicGen} and \textbf{DexMimicGen} \cite{DexMimicGen} demonstrated that a minimal set of human demonstrations could be procedurally expanded into tens of thousands of diverse trajectories across various tasks and bimanual robot setups. 

As simulation engines matured, recent datasets have pushed this scaling to the extreme. \textbf{InternData-A1} \cite{InternData-A1} scales automated trajectory synthesis to over 630k trajectories across hundreds of environments, while \textbf{SynGrasp-1B} \cite{SynGrasp-1B} leverages Isaac Sim and MuJoCo to generate an unprecedented 10 million grasping trajectories, reflecting the immense data appetite required for robust pretraining. Concurrently, scaling has expanded beyond pure quantity into sequential and semantic complexity. \textbf{RoboCasa} \cite{RoboCasa} introduces massive environmental diversity with over 100k trajectories situated in 120 realistic kitchen scenes. Pushing task complexity further, \textbf{RoboCerebra} \cite{RoboCerebra} emphasizes long-horizon manipulation with dense subtask instruction annotations, and \textbf{QUARD-Auto} \cite{QUARD-Auto} broadens embodiment scope to multi-task learning for quadrupeds. For WAMs, this automated expansion of long-horizon tasks and infinite scene variations is critical for learning stable, non-drifting state transitions over extended time steps.

\subsubsection{Spatiotemporal Dynamics: 3D and 4D Embodied Modeling} 
A core limitation of real-world video data is the loss of 3D spatial information due to 2D camera projections and dynamic occlusions. Simulation circumvents this by providing exact spatial ground truth. This capability has fueled a specialized trajectory of datasets designed explicitly for spatial and temporal world modeling. \textbf{TesserAct} \cite{TesserAct1} represents a major milestone in this direction, providing 285k aligned RGB, depth, and surface normal video clips to explicitly target 4D (3D space \text{+} time) embodied world modeling.

Similarly, \textbf{InternData-M1} \cite{InternVLA-M1} complements large-scale trajectory data with perfectly aligned, dense frame-level 2D/3D bounding boxes, segmentation masks, and explicit grasp points. Furthermore, frameworks like \textbf{RoboTwin 2.0} \cite{RoboTWIN_2} leverage high-fidelity 3D assets (e.g., from Objaverse) coupled with photorealistic rendering to generate digital twins of real-world objects. For WAMs, these datasets are invaluable; they provide the spatially dense supervision necessary to understand how object geometries and 3D scenes transform under specific robot actions, enabling a leap from 2D pixel prediction to true 3D spatial dynamics.

\subsubsection{Contact-Rich Physics: High-Fidelity Tactile Supervision} 
While simulating rigid body kinematics is well-established, simulating contact dynamics—such as friction, soft-body deformation, and tactile feedback—has historically been a challenge. However, as WAMs seek to master contact-rich manipulation, explicit force and tactile priors are required. Recent advancements in physics engines (e.g., NVIDIA FleX \cite{FleX}) have begun to unlock this capability. 

The \textbf{TLA Dataset} \cite{TLA} represents a pioneering effort in this domain, extending simulated supervision beyond vision by providing aligned Tactile-Language-Action pairs for fingertip-based assembly tasks. By generating synthetic tactile readings, simulation allows WAMs to model the invisible, sub-millimeter force interactions that occur upon contact. 

\vspace{1mm}
While the visual discrepancy between synthetic rendering and real-world observations—the sim-to-real gap—remains an ongoing challenge, the underlying physical principles (e.g., gravity, momentum, non-penetration) are strictly consistent. By employing extensive domain randomization over textures, lighting, and camera parameters (as seen in \textbf{SynGrasp-1B} and \textbf{RoboTwin 2.0}), these simulation datasets act as an expansive physics gym. They imbue WAMs with robust physical priors and spatial reasoning capabilities, dramatically reducing the amount of real-world data required for downstream finetuning.

\begin{table}[t]
    \centering
    \scriptsize
    \setlength{\tabcolsep}{3.5pt}
    \renewcommand{\arraystretch}{1.08}
    \caption{Summary of \textbf{simulation datasets}. ``Modality'' denotes observation modalities: RGB, P (proprioception), D (depth), PC (point clouds), T (tactile), and text.}
    \label{tab:sim_data}
    \vspace{-6pt}
    
    \begin{adjustbox}{width=\columnwidth}
    \begin{threeparttable}
    \begin{tabular}{
    >{\RaggedRight\arraybackslash}p{2.25cm}
    l l l l l
    >{\RaggedRight\arraybackslash}p{2.20cm}
    >{\RaggedRight\arraybackslash}p{2.05cm}
    >{\RaggedRight\arraybackslash}p{1.75cm}
    >{\RaggedRight\arraybackslash}p{1.80cm}
    }
    \toprule
    \textbf{Name} & \textbf{Year} & \textbf{Scale} & \textbf{Skill} & \textbf{Task} & \textbf{Env} & \textbf{Embodiment} & \textbf{Engine} & \textbf{Collection} & \textbf{Modality} \\
    \midrule
    
    MimicGen \cite{MimicGen} & 2023 & 50k traj. 
    & - & 18 & 1 & 4 robots & MuJoCo, Isaac Gym & teleop., augmentation & RGB \\

    ManiSkill2 \cite{ManiSkill2} & 2023 & 4M+ frames 
    & 4 & 20 & - & Franka & SAPIEN & simulation & RGB, D, PC, P \\

    RoboCasa \cite{RoboCasa} & 2024 & 100k+ traj. 
    & 8 & 100 & 120 & Franka & MuJoCo & simulation & RGB \\

    RoboTwin \cite{RoboTwin} & 2024 & - 
    & - & 17 & - & COBOT Magic & - & teleop. & RGB, D, P \\

    DexMimicGen \cite{DexMimicGen} & 2024 & $21,000$ traj. 
    & - & 9 & - & GR1, Panda & robosuite(MuJoCo) & real2sim2real & RGB, P \\

    QUARD-Auto \cite{QUARD-Auto} & 2024 & $258,418$ traj. 
    & 5 & 5 & - & Quadruped robot & Isaac Gym & auto-collected & RGB, text \\
        
    TesserAct \cite{TesserAct1} & 2025 & 285k video clips 
    & - & - & - & Google Robot, Franka, WidowX 250 & RLBench (CoppeliaSim) & aggregation & RGB, D \\

    RoboCerebra \cite{RoboCerebra} & 2025 & $1,000$ traj. 
    & 12 & 100 & - & - & Libero (MuJoCo) & simulation & RGB, D, text \\

    SynGrasp-1B \cite{SynGrasp-1B} & 2025 & 10M traj. 
    & 2 & 1 & - & Franka & Isaac Sim, MuJoCo & auto-collected & RGB, D, P \\
        
    RoboTwin 2.0 \cite{RoboTWIN_2} & 2025 & 100k+ traj. 
    & - & 50 & - & 5 robots & Isaac Sim & simulation & RGB, D, P \\

    TLA Dataset \cite{TLA} & 2025 & 24k tactile-action pairs 
    & - & - & - & gripper with GelStereo 2.0 & Isaac Gym + FleX & simulation & T, text \\
        
    InternData-M1 \cite{InternVLA-M1} & 2025 & 244k traj. 
    & 1 & 200 & - & Franka & Isaac Gym & auto-collected & RGB, D, P \\

    InternData-A1 \cite{InternData-A1} & 2025 & 630k traj. 
    & 18 & 70 & 227 & 4 robots & Isaac Sim & simulation & RGB, P \\
        
    \bottomrule
    \end{tabular}
    \end{threeparttable}
    \end{adjustbox}
    
    \vspace{-6pt}
\end{table}

\subsection{Human and Ego-Centric Data}
\label{sec:data_human}

A fundamental advantage of World Action Models (WAMs), relative to traditional VLA models, lies in their ability to internalize generalizable world dynamics. While conventional robot-centric datasets provide precise low-level action execution, they are inherently bottlenecked by the morphological constraints and high collection costs of physical robots. Consequently, they offer limited exposure to the long tail of unconstrained real-world physics. In contrast, internet-scale human and egocentric datasets (summarized in Table~\ref{tab:ego_data}) encapsulate near-infinite diversity in tasks, environments, and physical interactions. By acting as a massive repository of real-world priors, these datasets are increasingly recognized as the cornerstone for training generalist WAMs \cite{egoscale}. The evolution of these resources can be categorized into two pivotal trajectories: learning passive world dynamics from raw visual observations, and extracting active, action-conditioned dynamics through human pose and proprioceptive grounding.

\subsubsection{Passive World Modeling: Action Semantics and Visual Dynamics} 

A wide range of large-scale video datasets completely lack explicit proprioceptive signals but play a critical role in teaching WAMs the passive laws of intuitive physics. Early foundational works focused on breaking down physical common sense through short-horizon interactions. For instance, \textbf{SSv2} \cite{SSv2} provides over 100,000 clips capturing fundamental physical events—such as pushing, dropping, and tearing—forcing models to learn temporal reasoning and basic state transitions. Moving toward unscripted first-person perception, \textbf{EPIC-KITCHENS} \cite{EPIC-KITCHENS} introduced 55 hours of long-horizon daily activities, while \textbf{EGTEA Gaze+} \cite{EGTEA_Gaze+} incorporated explicit gaze tracking, providing WAMs with human-like visual attention priors during manipulation.
\begin{table}[t]
    \centering
    \scriptsize
    \setlength{\tabcolsep}{3.5pt}
    \renewcommand{\arraystretch}{1.08}
    \caption{Summary of \textbf{human / egocentric datasets}. "Modality" denotes observation modalities: RGB, P (proprioception), D (depth), A (audio), PC (point clouds), T (tactile), and text.}
    \label{tab:ego_data}
    \vspace{-6pt}
    
    \begin{adjustbox}{width=\columnwidth}
    \begin{threeparttable}
    \begin{tabular}{
    >{\RaggedRight\arraybackslash}p{2.65cm}
    c
    >{\RaggedRight\arraybackslash}p{2.35cm}
    >{\RaggedRight\arraybackslash}p{1.00cm}
    >{\RaggedRight\arraybackslash}p{1.35cm}
    >{\RaggedRight\arraybackslash}p{0.95cm}
    >{\RaggedRight\arraybackslash}p{1.75cm}
    >{\RaggedRight\arraybackslash}p{2.70cm}
    }
    \toprule
    \textbf{Name} & \textbf{Year} & \textbf{Scale} & \textbf{Skill} & \textbf{Task} & \textbf{Env} & \textbf{Collection} & \textbf{Modality} \\
    \midrule
    
    SSv2 \cite{SSv2} & 2018 & $108,499$ clips 
    & 174 & 1 & - & crowdsourced & RGB \\
    
    EPIC-KITCHENS \cite{EPIC-KITCHENS} & 2018 & 55 h 
    & - & 3 & - & manual & RGB, A \\
    
    HowTo100M \cite{HowTo100M} & 2019 & 136M clips 
    & - & 23K+ & - & script, ASR & RGB \\
    
    Kinetics-700 \cite{Kinetics-700} & 2019 & 650K clips 
    & - & - & - & web mining & RGB \\
    
    EGTEA Gaze+ \cite{EGTEA_Gaze+} & 2020 & 28.5 h 
    & - & 7 & - & manual & RGB, A \\
    
    Ego4D \cite{Ego4D} & 2021 & $3,670$ h 
    & - & 5 & - & manual & RGB, A \\
    
    HOI4D \cite{HOI4D} & 2022 & 2.4M frames 
    & - & 54 & - & manual & RGB, D \\
    
    EgoVid-5M \cite{EgoVid-5M} & 2024 & 5M clips 
    & - & - & - & derived & RGB, text \\
    
    COM Kitchens \cite{COM_Kitchens} & 2024 & 145 clips 
    & 131 & - & - & manual & RGB \\
    
    Egocentric-10K \cite{Egocentric-10k} & 2025 & 10K h 
    & - & - & - & manual & RGB \\
    
    DreamDojo-HV \cite{DreamDojo} & 2026 & $43,827$ h 
    & $6,015$ & $6,015$ & - & crowdsourced & RGB, text \\

    Assembly101 \cite{Assembly101} & 2021 & $4,321$ videos 
    & - & 4 & - & manual & RGB, P \\

    H2O \cite{H2O} & 2021 & $571,645$ frames 
    & - & 36 & 3 & manual & RGB, D, P, PC, text \\

    EgoPAT3D \cite{EgoPAT3D} & 2022 & 1M frames 
    & - & - & 15 & manual & RGB, D, PC, P \\
        
    Ego-Exo4D \cite{Ego-Exo4D} & 2023 & 1,286 h 
    & 8 & 4 & 123 & manual & RGB, A, PC, P \\
        
    ARCTIC \cite{ARCTIC} & 2023 & 2.1M frames 
    & - & 2 & - & manual & RGB, D, P \\
        
    HoloAssist \cite{HoloAssist} & 2023 & 166 h 
    & - & 20 & - & manual & RGB, D, P, A, text \\
        
    HOT3D \cite{HOT3D} & 2024 & 3.7M+ frames 
    & - & - & 4 & manual & RGB, PC, P \\
        
    TACO \cite{TACO} & 2024 & 5.2M frames 
    & 15 & 3 & - & manual & RGB, D, text, P \\

    Aria Everyday Activities \cite{Aria-Everyday-Activities} & 2024 & 7.3 h 
    & - & - & 5 & manual & RGB, A, P, PC, IMU \\
        
    OAKINK2 \cite{OAKINK2} & 2024 & 4.01M frames 
    & 60 & 150 & 4 & manual & RGB, P, text \\
        
    Nymeria \cite{Nymeria} & 2024 & 300 h 
    & - & 20 & 50 & manual & RGB, PC, P, A \\
        
    EgoMimic \cite{EgoMimic} & 2024 & 16 h
    & - & - & - & manual, teleop. & RGB, P \\
        
    PH$^2$D \cite{Human_Policy} & 2025 & 3M+ frames 
    & - & 4 & - & manual & RGB, P \\

    Human2Robot \cite{Human2Robot} & 2025 & $2,600$ episodes 
    & - & 14 & - & teleop. & RGB, P \\
        
    Humanoid Everyday \cite{Humanoid_Everyday} & 2025 & 10.3K traj., 3M frames 
    & 221 & 260 & - & teleop. & RGB, D, T, P, text \\
        
    IndEgo \cite{IndEgo} & 2025 & 294 h 
    & 5 & 25 & - & manual & RGB, P, A, PC \\

    PLAICraft \cite{PLAICraft} & 2025 & 10K+ h 
    & 10 & 8 & 1 & crowdsourced & RGB, A \\

    HD-EPIC \cite{HD-EPIC} & 2025 & 41.3 h 
    & - & - & 9 & manual & RGB, A, P \\

    UniHand \cite{UniHand} & 2025 & $1,155$ h 
    & - & 150+ & - & manual, annotation & RGB, P, text \\

    OpenEgo \cite{OpenEgo} & 2025 & $1,107$ h 
    & - & 290 & 600+ & aggregation, annotation & RGB, D, P, text \\

    Ego-Centric Human Manipulation \cite{Ego-centric-Human-Manipulation-Dataset} & 2025 & $500,000$ image-action pairs 
    & - & 12 & 25 & manual & RGB, P, text \\
        
    Kaiwu \cite{Kaiwu} & 2025 & $11,664$ actions 
    & - & - & - & manual & RGB, D, T, A, P \\
        
    EgoDex \cite{EgoDex} & 2026 & 829 h 
    & - & 194 & - & manual & RGB, P, text \\

    EgoScale \cite{egoscale} & 2026 & $20,854$ h 
    & 4 & $6,015$ & $9,869$ & manual & RGB, P, text \\

    EgoVerse \cite{EgoVerse} & 2026 & $1,362$ h 
    & 6 & $1,965$ & 240 & manual & RGB, D, P, text \\

    HumanNet \cite{HumanNet} & 2026 & 1M h 
    & 8 &  720K+ & - & aggregation, crawling, annotation  & RGB, P, text \\
        
    \bottomrule
    \end{tabular}
    \end{threeparttable}
    \end{adjustbox}
    
    \vspace{-6pt}
\end{table}

As deep learning scaled, so did the data volume and task complexity. \textbf{Ego4D} \cite{Ego4D} represented a monumental leap, offering over 3,600 hours of unscripted egocentric video. Crucially for WAM development, Ego4D introduced specific benchmarks for episodic memory and future forecasting, directly training models to predict future states ($s_{t\text{+}1}$) based on current visual contexts. To expose models to unbounded open-world semantics, web-scale datasets like \textbf{HowTo100M} \cite{HowTo100M} (with 136 million clips) and \textbf{Kinetics-700} \cite{Kinetics-700} provided massive video-text alignment. This enables WAMs to ground open-vocabulary semantic instructions into diverse visual state transitions across millions of instances. More recent efforts have further refined this passive supervision: \textbf{COM Kitchens} \cite{COM_Kitchens} introduces visual action graphs to chart the compositional rules between actions and state changes; \textbf{Egocentric-10K} \cite{Egocentric-10k} expands to high-precision industrial workflows; and \textbf{DreamDojo} \cite{DreamDojo} pushes the absolute limits of scale with 44,000 hours of crowdsourced activities. Pretraining on this massive corpus establishes a deeply ingrained understanding of object permanence and material properties long before any robot-specific actions are introduced.

\subsubsection{Bridging the Action Gap: Pose Estimation and Proprioceptive Grounding} 
While raw video provides visual common sense, learning action-conditioned dynamics ($s_{t\text{+}1} = f\left(s_t, a_t\right)$) strictly requires action inputs. To bypass the scarcity of robot data, researchers have increasingly treated the human hand as a universal end-effector. By annotating egocentric videos with 3D pose and motion tracking, these datasets mathematically bridge the gap between human videos and robotic policy learning.

Early datasets established the foundation for spatial supervision. \textbf{Assembly101} \cite{Assembly101} provided thousands of structured assembly videos, while \textbf{H2O} \cite{H2O} introduced explicit multi-view 3D hand and 6D object pose annotations, converting visual interactions into precise geometric trajectories. \textbf{EgoPAT3D} \cite{EgoPAT3D} pushed this further by focusing on future spatial anticipation. This precision was subsequently scaled to capture complex contact physics and view-invariant understanding. The milestone \textbf{Ego-Exo4D} \cite{Ego-Exo4D} dataset captures 1,286 hours of synchronized egocentric and exocentric perspectives, forcing models to understand 3D spatial transformations independently of camera angles. For dexterous manipulation, datasets like \textbf{ARCTIC} \cite{ARCTIC}, \textbf{HOT3D} \cite{HOT3D}, and \textbf{TACO} \cite{TACO} offer millimeter-accurate 3D hand-object meshes. These resources allow WAMs to model intricate force closures and bimanual geometric relationships. Furthermore, datasets have expanded toward global embodiment awareness: \textbf{Aria Everyday Activities} \cite{Aria-Everyday-Activities} integrates SLAM point clouds and eye gaze, \textbf{OAKINK2} \cite{OAKINK2} links hierarchical task intent with fine poses, and \textbf{Nymeria} \cite{Nymeria} provides 300 hours of full-body motion capture.

\subsubsection{Toward Generalist Pretraining Mixtures} 

The most recent trend for WAM development is the curation of massive data mixtures explicitly designed for generalist policy pretraining. Rather than serving as isolated academic benchmarks, diverse datasets are now being aggregated into monolithic learning engines. The \textbf{Ego-Centric Human Manipulation Dataset} \cite{Ego-centric-Human-Manipulation-Dataset} and \textbf{UniHand} \cite{UniHand} pool over 130 million frames and thousands of hours of video, aligning disparate RGB observations with unified kinematic hand poses. \textbf{EgoDex} \cite{EgoDex} represents the latest major leap in this progression, dramatically scaling to 829 hours and 194 tasks while maintaining high-fidelity 3D finger tracking for dexterous control. Concurrently, datasets like \textbf{Humanoid Everyday} \cite{Humanoid_Everyday} and \textbf{PH$^2$D} \cite{Human_Policy} explicitly align human egocentric demonstrations with the kinematics of humanoid robots. For the future of WAMs, these proprioception-grounded mixtures act as the ultimate catalyst, allowing models to absorb massive volumes of human behavioral data and project it directly into robotic action spaces.

\section{Evaluation}
\label{sec:eval}

Comprehensively evaluating WAMs requires assessing both the fidelity of predicted future states and the effectiveness of generated actions—and ideally, the causal alignment between the two. In practice, however, no established protocol jointly evaluates these interdependent components. Existing work instead adopts a decoupled paradigm, assessing world modeling capability and action policy capability through separate, module-specific metrics. Following this prevailing convention, we structure our review along these two complementary axes: (1) \textbf{World Modeling Capability} (\autoref{sec:eval_world}), which examines the physical integrity and structural consistency of synthesized transitions, and (2) \textbf{Action Policy Capability} (\autoref{sec:eval_action}), which measures the effectiveness of generated actions in fulfilling embodied tasks.

\subsection{How to Evaluate World Modeling Capability?}

\label{sec:eval_world}

Evaluating the world modeling component of a WAM differs fundamentally from assessing conventional video generation. Beyond surface-level visual realism, a world action model is expected to faithfully capture the underlying dynamics of the environment and preserve actionable information. Accordingly, we categorize the current evaluation of world modeling capability into three parallel aspects: \textbf{Visual Fidelity} (\autoref{sec:eval_world_visual}), which assesses the quality and temporal consistency of the visual interface; \textbf{Physical Commonsense} (\autoref{sec:eval_world_physical}), which examines the adherence to fundamental material and mechanical laws; and \textbf{Action Plausibility} (\autoref{sec:eval_world_action}), which measures whether the synthesized transitions contain sufficient information to be translated back into executable control signals.

\subsubsection{Visual Fidelity}
\label{sec:eval_world_visual}
Visual Fidelity remains the most basic layer in evaluating world action models, since physically plausible reasoning or action extraction becomes unreliable when generated videos suffer from severe artifacts, temporal inconsistency, or poor condition following. In practice, recent works typically combine low-level reconstruction metrics, perceptual similarity metrics, semantic alignment signals, and distribution-level realism metrics to assess video quality from multiple perspectives.

At the pixel level, \textbf{PSNR} (Eq.~\ref{eq:psnr}) measures reconstruction fidelity through the logarithmic ratio between the maximum signal value and the mean squared error, while \textbf{SSIM}, as shown in Eq.~\ref{eq:ssim}, evaluates structural similarity by comparing luminance, contrast, and structural information~\citep{ssim}. 

\begin{equation}
\label{eq:psnr}
\mathrm{PSNR}(x,y)=10\log\left(\frac{\mathrm{MAX}^2}{\mathrm{MSE}(x,y)}\right).
\end{equation}

\begin{equation}
\label{eq:ssim}
\mathrm{SSIM}(x,y)=\frac{(2\mu_x\mu_y+C_1)(2\sigma_{xy}+C_2)}{(\mu_x^2+\mu_y^2+C_1)(\sigma_x^2+\sigma_y^2+C_2)}.
\end{equation}

To go beyond low-level statistics, many recent works further introduce feature-based perceptual or semantic similarity metrics. \textbf{LPIPS} (Eq.~\ref{eq:lpips}) measures perceptual similarity in a deep feature space by extracting multi-layer features, normalizing them channel-wise, applying channel-wise weighting, averaging distances over spatial locations, and summing across layers; it is widely adopted in video generation and world-model evaluation~\citep{lpips}. \textbf{DreamSim} (Eq.~\ref{eq:dreamsim}) provides a human-aligned perceptual similarity signal trained on human judgments over image triplets, where each triplet contains a reference image and two candidate images and annotators select which candidate is more perceptually similar to the reference; this makes it useful for evaluating whether generated content remains perceptually similar to the reference in terms of object layout, identity, and scene semantics~\citep{fu2023dreamsim}. Some works also use \textbf{DINO}-based feature similarity, as shown in Eq.~\ref{eq:dino}, as a semantic or instance-level alignment signal, since self-supervised visual representations tend to preserve object identity and scene semantics more robustly than pixel-level distances~\citep{oquab2023dinov2}.

\begin{equation}
\label{eq:lpips}
\mathrm{LPIPS}(x,y)
=
\sum_{l}
\frac{1}{H_l W_l}
\sum_{h,w}
\left\|
w_l \odot
\bigl(
\hat f_l(x)_{hw} - \hat f_l(y)_{hw}
\bigr)
\right\|_2^2.
\end{equation}
where $\hat f_l(\cdot)_{hw}$ denotes the channel-normalized deep feature at spatial location $(h,w)$ from layer $l$, and $w_l$ is a learned channel-wise weight vector.

\begin{equation}
\label{eq:dreamsim}
\mathrm{DreamSim}(x,y) = 1 - \| E(x) - E(y) \|_2.
\end{equation}
where $E(\cdot)$ is the fused embedding.

\begin{equation}
\label{eq:dino}
\mathrm{DINO}(g_t, r_t) = \frac{\langle f(g_t), f(r_t) \rangle}{\|f(g_t)\|_2 \, \|f(r_t)\|_2}.
\end{equation}
where $f(\cdot)$ denotes the DINOv2 encoder.

At the distribution level, \textbf{FVD} (Eq.~\ref{eq:fvd}) is one of the most widely used metrics for video generation. Instead of comparing individual samples frame by frame, FVD computes the Fr\'echet distance between the distributions of real and generated videos in a pretrained video feature space, thereby reflecting overall realism and temporal dynamics at the dataset level~\citep{unterthiner2018fvd}. 

\begin{equation}
\label{eq:fvd}
\mathrm{FVD}
= \|\mu_r - \mu_g\|_2^2
+ \mathrm{Tr}\!\left(\Sigma_r + \Sigma_g 
- 2\left(\Sigma_r\Sigma_g\right)^{1/2}\right).
\end{equation}

Overall, video quality evaluation in world action models usually relies on a combination of \textbf{PSNR/SSIM} for low-level fidelity, \textbf{LPIPS/DreamSim/DINO similarity} for perceptual and semantic consistency, and \textbf{FVD} for distribution-level realism and temporal quality.
\begin{table*}[!h]
\centering
\scriptsize
\setlength{\tabcolsep}{4pt}
\renewcommand{\arraystretch}{1.15}
\caption{Summary of world modeling evaluation metrics and benchmarks.}
\label{tab:wam_eval_benchmarks}

\begin{adjustbox}{width=\textwidth}
\begin{threeparttable}
\begin{tabular}{
>{\RaggedRight\arraybackslash}p{2.6cm}
c
>{\RaggedRight\arraybackslash}p{5.2cm}
>{\RaggedRight\arraybackslash}p{4.4cm}
}
\toprule
\textbf{Name} & \textbf{Year} & \textbf{Evaluation Focus} & \textbf{Metric / Implementation} \\
\midrule

\rowcolor{gray!12}
\multicolumn{4}{l}{\textbf{Visual Fidelity}} \\
PSNR & -- & Pixel-level reconstruction fidelity & Log ratio between maximum signal value and MSE \\

SSIM \cite{ssim} & 2004 & Structural similarity in luminance, contrast, and structure & Structural similarity index \\

LPIPS \cite{lpips} & 2018 & Deep perceptual similarity & Weighted distance in deep feature space \\

DreamSim \cite{fu2023dreamsim} & 2023 & Human-aligned perceptual similarity & Similarity in fused embedding space trained on human triplet judgments \\

DINO \cite{oquab2023dinov2} & 2023 & Semantic / instance-level alignment & Cosine similarity in DINOv2 feature space \\

FVD \cite{unterthiner2018fvd} & 2018 & Distribution-level realism and temporal quality & Fr\'echet distance in pretrained video feature space \\

\midrule

\rowcolor{gray!12}
\multicolumn{4}{l}{\textbf{Object Dynamics}} \\
VideoPhy \cite{bansal2024videophy} & 2024 & Physical interaction scenarios including solid-solid, solid-fluid, and fluid-fluid interactions; object continuity and physical plausibility & Binary human annotations on semantic adherence and physical commonsense \\

PhyGenBench \cite{meng2024phygenbench} & 2024 & Physical commonsense alignment in generated videos & PhyGenEval with key physical phenomena detection, physics order verification, and overall naturalness \\

VBench-2.0 \cite{zheng2025vbench} & 2025 & Physics and commonsense violations, including mechanical, thermal, and material state changes, and abnormal entity behavior & Video-based multi-question answering for Physics; clip-level abnormal-entity detector for Commonsense \\

WorldModelBench \cite{li2025worldmodelbench} & 2025 & Physics adherence in long-horizon generated world dynamics & Five binary physical-law checks with human annotations and a fine-tuned VLM-based automatic judger \\

Physics-IQ \cite{motamed2026physicsiq} & 2026 & Future evolution of real-world physical events from conditioning frames & Spatial IoU, Spatiotemporal IoU, Weighted Spatial IoU, MSE \\

\midrule

\rowcolor{gray!12}
\multicolumn{4}{l}{\textbf{Motion and Trajectory Plausibility}} \\
WorldScore \cite{duan2025worldscoreunifiedevaluationbenchmark} & 2025 & Controllability, quality, and dynamics; dynamics includes motion accuracy, motion magnitude, and motion smoothness & Optical-flow-based motion magnitude; smoothness via comparison to interpolated references \\

EWMBench \cite{yue2025ewmbench} & 2025 & Motion correctness and trajectory consistency in embodied world models & EEF trajectory metrics: HSD, nDTW, DYN \\

\midrule

\rowcolor{gray!12}
\multicolumn{4}{l}{\textbf{Action Plausibility}} \\
WorldSimBench \cite{qin2024worldsimbenchvideogenerationmodels} & 2024 & Whether generated situation-aware videos preserve control-relevant information for downstream manipulation & Implicit Manipulative Evaluation \\

Wow, wo, val! \cite{fan2026wow} & 2026 & Whether generated videos can be translated back into executable actions & IDM Turing Test with downstream real-world execution success \\

\bottomrule
\end{tabular}
\end{threeparttable}
\end{adjustbox}
\end{table*}


\subsubsection{Physical Commonsense}
\label{sec:eval_world_physical}
While video quality measures whether a generated video is visually convincing, \textbf{physical commonsense} addresses a deeper question: whether the generated world behaves in a physically plausible way. Existing evaluations in this direction can be organized into two representative categories: \textbf{object dynamics} and \textbf{trajectory plausibility}.

\paragraph{Object Dynamics}
Object dynamics concerns whether generated videos preserve object continuity over time while also modeling physically grounded interactions and temporally coherent event evolution. This includes not only whether entities remain stable across frames, but also whether contacts, collisions, state changes, and their causal ordering unfold in plausible ways. Such capabilities are central to world models, which must represent how objects persist and evolve under physical constraints rather than merely produce visually realistic frames.

A representative early benchmark is \textbf{VideoPhy}~\citep{bansal2024videophy}, which evaluates text-to-video generation under physical interaction scenarios including solid-solid, solid-fluid, and fluid-fluid interactions. It relies on binary human annotations on semantic adherence and physical commonsense, making it effective for identifying obvious failures in physical plausibility. 

More targeted physical evaluation is provided by \textbf{PhyGenBench}~\citep{meng2024phygenbench}, which introduces physical commonsense alignment and measures it primarily through \textbf{PhyGenEval}, an automated evaluation framework built on VLMs and LLMs rather than human scoring alone. Specifically, PhyGenEval decomposes evaluation into key physical phenomena detection, physics order verification, and overall naturalness, while human annotators are used mainly to validate the metric by measuring its correlation with human judgments. 

\textbf{VBench-2.0}~\citep{zheng2025vbench} includes the physics dimension that evaluates whether generated videos obey real-world physical principles, including mechanical, thermal, and material state changes, mainly through video-based multi-question answering. Commonsense is evaluated with a specialized clip-level abnormal-entity detector rather than standard VQA. The detector checks whether a clip contains anomalies such as sudden merging, splitting, appearing, or disappearing, and assigns a binary score at the video level.

Another relevant benchmark is \textbf{WorldModelBench}~\citep{li2025worldmodelbench}. Its physics adherence score is defined with respect to five binary physical-law checks, including Newton’s first law, conservation of mass and solid mechanics, fluid mechanics, impenetrability, and gravitation. The benchmark first collects human annotations over these criteria, and then uses the resulting labels to fine-tune a VLM-based automatic judger.

Another important benchmark is \textbf{Physics-IQ}~\citep{motamed2026physicsiq}, which tests whether video generative models can predict the future evolution of real-world physical events from conditioning video frames. Unlike text-based physical benchmarks, it uses real videos and evaluates physical understanding with motion- and fidelity-based metrics, including Spatial IoU, Spatiotemporal IoU, Weighted Spatial IoU, and MSE.

\paragraph{Motion and Trajectory Plausibility}
Motion and trajectory plausibility evaluates whether the motion of objects or agents in generated videos is coherent, smooth, condition-aligned, and physically reasonable over time. Compared with object-centric physical dynamics, this aspect emphasizes long-horizon motion quality and whether generated trajectories evolve in a controllable and temporally stable manner.

A representative benchmark is \textbf{WorldScore}~\citep{duan2025worldscoreunifiedevaluationbenchmark}, which evaluates world generation from the perspectives of controllability, quality, and dynamics. Its dynamics dimension is further decomposed into motion accuracy, motion magnitude, and motion smoothness. In particular, motion magnitude is measured by estimating optical flow between consecutive frames, while motion smoothness is designed to capture temporal jittering through comparison against smooth interpolated references. 

A more trajectory-specific benchmark is \textbf{EWMBench}~\citep{yue2025ewmbench}, which evaluates embodied world models along visual scene consistency, motion correctness, and semantic alignment. To assess motion correctness, it explicitly uses end-effector (EEF) trajectories as the evaluation target and introduces trajectory consistency metrics including HSD, nDTW, and DYN, which respectively measure spatial deviation, spatiotemporal alignment, and motion dynamics such as velocity and acceleration.

Overall, this category evaluates not only whether objects move, but whether they move in ways that are temporally stable, physically grounded, and consistent with control or task intent.

Taken together, physical commonsense evaluation in world action models goes beyond whether videos are visually realistic. It tests whether generated worlds preserve stable object identity, whether interactions unfold according to causal and material constraints, and whether object motion remains plausible over time.

\subsubsection{Action Plausibility}
\label{sec:eval_world_action}
The final and most distinctive aspect in evaluating world action models is \textbf{action plausibility}. If video quality asks whether the generated world looks realistic, and physical commonsense asks whether it behaves plausibly, action plausibility asks whether the generated video preserves sufficient action-relevant information to support control inference and downstream execution.

An early benchmark in this direction is \textbf{WorldSimBench}~\citep{qin2024worldsimbenchvideogenerationmodels}, which introduces Implicit Manipulative Evaluation in addition to explicit perceptual assessment. Rather than judging generated videos only by visual quality, it evaluates whether a generated situation-aware video can be accurately translated into the correct control signals in dynamic embodied environments.

This perspective is further strengthened in \textbf{Wow, wo, val!}~\citep{fan2026wow}, which introduces the \textbf{Inverse Dynamics Modeling (IDM) Turing Test}. In this setting, an IDM is applied to generated videos to infer the underlying action sequence, and the resulting actions are evaluated by real-world execution success. Their results show that many visually convincing models collapse to nearly zero success under this test. This finding highlights a key gap between visually plausible video generation and executable robot behavior, and establishes action plausibility as a distinct evaluation axis beyond appearance and physical realism.

In summary, the evaluation of world action models for video generation can be structured into three complementary dimensions: \textbf{video quality}, \textbf{physical commonsense}, and \textbf{action plausibility}. Video quality ensures that generated videos are visually faithful and perceptually consistent; physical commonsense evaluates whether the generated world respects object continuity, interaction dynamics, and plausible motion; and action plausibility examines whether generated videos preserve sufficient action information to support downstream control and execution.


\subsection{How to evaluate Action Policy?}
\label{sec:eval_action}

While world modeling evaluation focuses on the fidelity of synthesized transitions, \textbf{action policy evaluation} targets the \textit{policy capability}—specifically the model's ability to generate precise, robust, and generalizable control signals across diverse scenarios. As WAMs transition from passive video generation to active robotic control, the systematic assessment of these policies has emerged as a core research priority. 
As summarized in \autoref{tab:benchmark_compare}, we conduct a systematic review of over 40 mainstream benchmarks proposed from 2019 to 2026, covering key design dimensions including simulation environment setup, sensor modality configuration, demonstration dataset scale, and multi-dimensional evaluation metrics. Broadly, existing benchmarks can be categorized into five groups based on robot morphology and manipulation scenarios: (1) \textbf{General Manipulation} (\autoref{sec:bench_general}), (2) \textbf{Bimanual and Humanoid Manipulation} (\autoref{sec:bench_bimanual}), (3) \textbf{Mobile Manipulation} (\autoref{sec:bench_mobile}), (4) \textbf{Contact-rich and Deformable Object Manipulation} (\autoref{sec:bench_contact}), and (5) \textbf{Real-Robot Evaluation} (\autoref{sec:bench_real}).

\subsubsection{General Manipulation Benchmarks}
\label{sec:bench_general}
General manipulation is currently the most widely covered direction among benchmarks and
constitutes the main body of the action policy evaluation ecosystem.

Early benchmarks established the foundational framework for multi-task manipulation evaluation. \textbf{MetaWorld}~\cite{yu2019metaworld} and \textbf{RLBench}~\cite{RLBench} provided 50 and 100 manipulation tasks respectively, laying the groundwork for standardized multi-task policy comparison. Subsequent work shifted toward offline imitation learning: \textbf{Robomimic}~\cite{mandlekar2021robomimic} systematically investigated what design decisions matter when learning from human demonstrations, while \textbf{Franka Kitchen}~\cite{gupta2019relay} offered an early testbed for sequential task execution. Building on these foundations, \textbf{LIBERO}~\cite{LIBERO} introduced a more comprehensive evaluation suite spanning four dimensions—generalization, long-horizon tasks, lifelong learning, and language understanding—across 130 tasks, representing one of the earliest attempts to assess manipulation policies along multiple axes simultaneously.

As data-driven paradigms matured, benchmark scale expanded substantially in terms of object diversity, task count, and trajectory volume. The \textbf{ManiSkill} series exemplifies this trend: from 100+ objects and 30K+ trajectories in ManiSkill~\cite{mu2021maniskill}, to 2,144 objects across 20 tasks in \textbf{ManiSkill2}~\cite{ManiSkill2}, to 10K+ objects and 62 tasks with data scalability as an explicit evaluation dimension in \textbf{ManiSkill3}~\cite{tao2024maniskill3}. \textbf{RoboCasa}~\cite{RoboCasa} further pushed this frontier with 2,509 objects, 100 tasks, and 100K+ trajectories across 4+ robot platforms. Complementing these simulation-centric efforts, \textbf{RoboVerse}~\cite{geng2025roboverse} introduced a unified cross-simulator framework via MetaSim, aggregating 276 tasks and 500K+ trajectories while explicitly studying the relationship between data scale and policy performance across multiple simulators. Together, these works reframed the central benchmark question from \textit{can a policy complete a task} to \textit{how does data scale affect policy capability}.

Unlike earlier benchmarks that evaluated generalization along limited axes, \textbf{COLOSSEUM}~\cite{pumacay2024colosseum} and \textbf{LIBERO-plus}~\cite{fei2025liberoplus} broadened the scope by systematically introducing multi-dimensional visual and environmental perturbations to assess policy robustness~\cite{zhou2025liberopro, wang2026liberox}. Going further, task-level generalization has emerged as a demanding evaluation target: \textbf{AGNOSTOS}~\cite{zhou2025agnostos} introduced a two-level cross-task zero-shot generalization protocol, with Level-1 covering tasks sharing partial similarity with seen tasks and Level-2 targeting entirely novel scenarios requiring stronger generalization. \textbf{GemBench}~\cite{garcia2025gembench} introduced four progressively challenging generalization levels—from novel placements and rigid objects to articulated objects and long-horizon task combinations—providing the most structured assessment of language-conditioned generalization to date.

Beyond these directions, other researchers have explored additional evaluation dimensions. On language-conditioned manipulation, benchmarks such as~\cite{VIMA-Bench, zheng2022vlmbench} explored multimodal prompt generalization, while \textbf{CALVIN}~\cite{mees2022calvin} pioneered natural language guidance in long-horizon evaluation using approximately 24 hours of teleoperated play data. On long-horizon manipualtion, subsequent benchmarks~\cite{gao2025genmanip, zhang2024vlabench} pushed evaluation toward multi-step task chains requiring semantic, spatial, and commonsense reasoning, exposing limitations that single-step benchmarks cannot reveal. On Sim-to-Real transferability, \textbf{SimplerEnv}~\cite{li2024simplerenv} reproduced real-world manipulation scenarios from the Bridgev2 dataset within SAPIEN simulation, offering a quantifiable reference for the reality gap; \textbf{PolaRiS}~\cite{jain2025polaris} leveraged Gaussian Splatting to convert short video scans of real scenes into interactive simulation environments, demonstrating strong correlation with real-world policy performance and marking a shift in benchmark design from constructing synthetic tasks to reconstructing real scenes.  On memory, \textbf{RoboMME}~\cite{dai2026robomme} constructed 16 manipulation tasks spanning four cognitive memory dimensions—temporal, spatial, object, and procedural—targeting policy performance in history-dependent scenarios where information from earlier steps is necessary for successful execution.

\subsubsection{Bimanual and Humanoid Form Benchmarks}
\label{sec:bench_bimanual}

This category of benchmarks targets two types of embodiment: bimanual robots centered on
coordinated dual-arm operation, and humanoid robots with full-body motion capability. Both
present significantly higher action space dimensionality, motion constraints, and task
complexity compared to single-arm manipulation, placing higher demands on the coordinated
planning ability of action policies.

\textbf{RoboTwin}~\cite{RoboTwin} uses the Aloha-AgileX dual-arm platform as its
carrier and builds an evaluation environment on SAPIEN/ManiSkill3, focusing on long-horizon
task chain execution under bimanual coordination.
\textbf{BiGym}~\cite{chernyadev2024bigym} is based on the Unitree H1 platform and provides
40 mobile bimanual manipulation tasks in household scenarios in MuJoCo, covering diverse
manipulation scenes from simple target reaching to complex kitchen cleaning, and provides
human teleoperation demonstration data for each task to support evaluation of imitation and
reinforcement learning algorithms. \textbf{HumanoidBench}~\cite{sferrazza2024humanoidbench}
is also based on Unitree H1, equipped with bilateral Shadow-Hand dexterous hands and
full-body tactile sensing (448 tactile sensing points in total), and provides 27 tasks in
MuJoCo, covering 15 whole-body manipulation tasks (e.g., moving cargo, using tools, playing
basketball) and 12 locomotion tasks (e.g., walking, running, obstacle crossing), making it
currently one of the richest humanoid robot benchmarks in terms of sensor modality
configuration. \textbf{HumanoidGen}~\cite{li2025humanoidgen} is similarly based on the
Unitree platform and supports joint evaluation of generalization ability, long-horizon
manipulation, and dexterity in SAPIEN with 200K+ trajectories at large scale, while
systematically exploring the impact of data scale on humanoid robot policy learning.


\definecolor{tagG}{HTML}{C8DFF0}       
\definecolor{tagMT}{HTML}{C5E0D4}      
\definecolor{tagLH}{HTML}{F5DEB3}      
\definecolor{tagLang}{HTML}{D8C9E8}    
\definecolor{tagRob}{HTML}{F2C4C4}     
\definecolor{tagDS}{HTML}{C9DFC9}      
\definecolor{tagS}{HTML}{FFE4B5}     
\definecolor{tagDex}{HTML}{B5D5E8}     
\definecolor{tagMem}{HTML}{E8D5B5}     
\definecolor{tagLL}{HTML}{D4C5B5}      

\newtcbox{\focustag}[1][]{
  on line,
  arc=2.5pt,
  colback=#1,
  colframe=#1,        
  boxsep=0pt,
  left=2.5pt, right=2.5pt,
  top=1pt,   bottom=1pt,
  boxrule=0pt,
  fontupper=\scriptsize\sffamily,
  nobeforeafter,
}

\def\tagG    {\focustag[tagG]{G}}
\def\tagMT   {\focustag[tagMT]{MT}}
\def\tagLH   {\focustag[tagLH]{LH}}
\def\tagLang {\focustag[tagLang]{Lang}}
\def\tagRob  {\focustag[tagRob]{Rob}}
\def\tagDS   {\focustag[tagDS]{DS}}
\def\tagS  {\focustag[tagS]{S2R}}
\def\tagDex  {\focustag[tagDex]{Dex}}
\def\tagMem  {\focustag[tagMem]{Mem}}
\def\tagLL   {\focustag[tagLL]{LL}}

\newcommand{\tspace}{\hspace{1.5pt}}

\begin{table}[!t]
\centering
\scriptsize
\setlength{\tabcolsep}{3.5pt}
\renewcommand{\arraystretch}{1.08}
\caption{Summary of evaluation benchmarks for action policy.
\textbf{Obs.} denotes observation modalities: RGB, Pose, D (depth),
S (segmentation), PC (point clouds), T (tactile), N (Normal Vector).
\textbf{Eval.\ Focus tags:}
\tagG~Generalization,
\tagMT~Multi-task,
\tagLH~Long-horizon,
\tagLang~Language,
\tagDS~Data Scaling,
\tagS~Sim-to-Real,
\tagDex~Dexterity,
\tagMem~Memory,
\tagLL~Lifelong.}
\label{tab:benchmark_compare}

\begin{adjustbox}{width=\columnwidth}
\begin{threeparttable}
\begin{tabular}{
l l l l l
>{\RaggedRight\arraybackslash}p{1.25cm}
>{\RaggedRight\arraybackslash}p{1.95cm}
l
l
}
\toprule
\textbf{Name} & \textbf{Year} & \textbf{Obj.} & \textbf{Tasks} & \textbf{traj.}
& \textbf{Obs.} & \textbf{Robots} & \textbf{Simulator}
& \textbf{Eval. Focus} \\
\midrule

\rowcolor{gray!12}
\multicolumn{9}{l}{\textbf{General Manipulation}} \\

MetaWorld~\cite{yu2019metaworld}
  & 2019 & 80   & 50   & -     & Pose      & Sawyer          & MuJoCo
  & \tagMT \\

RLBench~\cite{RLBench}
  & 2020 & 28   & 100  & -     & RGB, D, S & Franka Panda    & CoppeliaSim
  & \tagG\tspace\tagMT \\

Robomimic~\cite{mandlekar2021robomimic}
  & 2021 & 15   & 8    & 6K    & RGB, D    & Franka Panda    & MuJoCo
  & – \\

Franka Kitchen~\cite{gupta2019relay}
  & 2020 & 10   & 7    & 513   & Pose      & Franka Panda    & MuJoCo
  & \tagLH \\

CALVIN~\cite{mees2022calvin}
  & 2021 & 28   & 34   & 20K+  & RGB, D    & Franka Panda    & PyBullet
  & \tagLH \\

ManiSkill~\cite{mu2021maniskill}
  & 2021 & 100+ & 4    & 30K+  & RGB, D, S & Franka Panda    & SAPIEN
  & \tagG \\

ManiSkill2~\cite{ManiSkill2}
  & 2023 & 2144 & 20   & 30K+  & RGB, D, S & Franka Panda    & SAPIEN
  & \tagMT \\

ManiSkill3~\cite{tao2024maniskill3}
  & 2025 & 10K+ & 62   & -     & RGB, D, S & Franka Panda    & SAPIEN
  & \tagDS\tspace\tagMT \\

VIMA-Bench~\cite{VIMA-Bench}
  & 2023 & 100+ & 17   & 600K+ & RGB, D, S & UE5             & PyBullet/Ravens
  & \tagLang \\

VLMbench~\cite{zheng2022vlmbench}
  & 2024 & 22   & 8    & 6K+   & RGB, D, S & Franka Panda    & CoppeliaSim/RLBench
  & \tagLang \\

ARNOLD~\cite{gong2023arnold}
  & 2023 & 40   & 8    & 10K+  & RGB, D, S & Franka Panda    & IsaacSim
  & \tagG \\

LIBERO~\cite{LIBERO}
  & 2023 & 67   & 130  & 6.5K  & RGB, D    & Franka Panda    & MuJoCo/robosuite
  & \tagLL \\

Libero-plus~\cite{fei2025liberoplus}
  & 2025 & -    & 10030 & 20K+ & RGB, D    & Franka Panda    & MuJoCo
  & \tagG \\

Libero-pro~\cite{zhou2025liberopro}
  & 2025 & -    & -    & -     & RGB, D    & Franka Panda    & MuJoCo
  & \tagG \\

Libero-x~\cite{wang2026liberox}
  & 2026 & 60+  & 600  & 2520  & RGB, D    & Franka Panda    & MuJoCo
  & \tagG \\

COLOSSEUM~\cite{pumacay2024colosseum}
  & 2024 & -    & 20   & 2K    & RGB, D    & Franka Panda    & CoppeliaSim/RLBench
  & \tagG \\

GemBench~\cite{garcia2025gembench}
  & 2025 & 20+  & 60   & -     & RGB, D    & Franka Panda    & CoppeliaSim
  & \tagG \\

AGNOSTOS~\cite{zhou2025agnostos}
  & 2025 & -    & 41   & 3.6K  & RGB, D, S & Franka Panda    & CoppeliaSim/RLBench
  & \tagG \\

SimplerEnv~\cite{li2024simplerenv}
  & 2024 & -    & -    & - & RGB, D & Google Robot, WidowX & SAPIEN
  & \tagS \\

RoboCasa~\cite{RoboCasa}
  & 2024 & 2509 & 100  & 100K+ & RGB, D    & 4+              & MuJoCo/robosuite
  & \tagLH \\

GenManip~\cite{gao2025genmanip}
  & 2025 & 10   & 200  & -     & RGB, D, S & Franka Panda    & IsaacSim
  & \tagG \\

VLABench~\cite{zhang2024vlabench}
  & 2024 & 2000+ & 100 & 5K    & RGB, D    & Franka Panda    & MuJoCo
  & \tagLang \\

RoboVerse~\cite{geng2025roboverse}
  & 2025 & 5.5K & 276  & 500K  & RGB, D    & 5               & MetaSim
  & \tagS\tspace\tagDS\tspace\tagMT \\

RoboSuite~\cite{robosuite}
  & 2020 & 20   & 9    & 12K   & RGB, D    & 10              & MuJoCo
  & \tagMT \\

RoboEval~\cite{wang2025roboeval}
  & 2025 & -    & 8    & 3K+   & RGB, D    & Franka Panda    & –
  & \tagG\tspace\tagDex \\

RoboMME~\cite{dai2026robomme}
  & 2026 & -    & 16   & 1600  & RGB, D, S & Franka Panda    & ManiSkill3
  & \tagMem \\

RoboLab~\cite{yang2026robolab}
  & 2026 & -    & 120  & -     & RGB, D    & Franka Panda    & IsaacSim
  & – \\

PolaRiS~\cite{jain2025polaris}
  & 2025 & -    & -    & \textasciitilde350 & RGB, D & Franka Panda & IsaacSim
  & \tagS \\

\midrule

\rowcolor{gray!12}
\multicolumn{9}{l}{\textbf{Arms and Humanoid Form}} \\

RoboTwin~\cite{RoboTwin}
  & 2025 & 10+ & -  & 320   & RGB, D, S & Aloha-AgileX             & SAPIEN/ManiSkill3
  & \tagMT\tspace\tagDex \\

BiGym~\cite{chernyadev2024bigym}
  & 2024 & 10+ & 40 & 2K    & RGB, D    & Unitree H1               & MuJoCo
  & \tagMT\tspace\tagDex \\

HumanoidBench~\cite{sferrazza2024humanoidbench}
  & 2024 & 15  & 27 & 45K+  & RGB, D, T & Unitree H1 + Shadow-Hand & MuJoCo
  & \tagMT\tspace\tagDex \\

HumanoidGen~\cite{li2025humanoidgen}
  & 2025 & -   & 20 & 200K+ & RGB, D    & Unitree                  & SAPIEN
  & \tagDS\tspace\tagDex \\

\midrule

\rowcolor{gray!12}
\multicolumn{9}{l}{\textbf{Mobile Manipulation}} \\

HomeRobot~\cite{yenamandra2023homerobot}
  & 2023 & 7892  & 12  & -      & RGB, D    & Hello Robot Stretch        & AI Habitat
  & \tagG\tspace\tagS \\

ManipulaTHOR~\cite{ehsani2021manipulathor}
  & 2021 & 2.6K+ & 28  & 12.8K+ & RGB, D, N & Kinova Gen3 on Mobile Base & Unity/AI2-THOR
  & \tagG \\

BEHAVIOR-1K~\cite{li2023behavior1k}
  & 2023 & 5215  & 1K  & 230K+  & RGB, D    & Mobile Manipulator         & OmniGibson
  & \tagG\tspace\tagLH \\

\midrule

\rowcolor{gray!12}
\multicolumn{9}{l}{\textbf{Contact and Deformation Manipulation}} \\

SoftGym~\cite{lin2021softgym}
  & 2021 & 4 & 10 & 10K+ & RGB, D, P & Sawyer, Franka      & NVIDIA FleX
  & \tagDex \\

PlasticineLab~\cite{huang2021plasticinelab}
  & 2021 & 1 & 10 & 1.2K & RGB, D, P & Rigid End-Effector  & Taichi+DiffTaichi
  & \tagDex \\

DaXBench~\cite{chen2023daxbench}
  & 2023 & 4 & 9  & 9K+  & RGB, D, P & –                   & DaX
  & \tagDex \\

TacSL~\cite{akinola2024tacsl}
  & 2025 & - & 3  & -    & RGB, D, T & Franka Panda        & Isaac Gym
  & \tagS\tspace\tagDex \\

ManiFeel~\cite{xu2025manifeel}
  & 2025 & - & 6  & -    & RGB, D, T & Franka Panda        & Isaac Sim/TacSL
  & \tagS\tspace\tagDex \\

\midrule

\rowcolor{gray!12}
\multicolumn{9}{l}{\textbf{Real Device}} \\

RoboArena~\cite{atreya2025roboarena}
  & 2025 & - & -  & -     & RGB, D & Franka Panda                  & –
  & \tagG \\

RoboChallenge~\cite{yakefu2025robochallenge}
  & 2026 & - & 30 & -     & RGB, D & 4                             & –
  & \tagMT \\

Maniparena~\cite{sun2026maniparena}
  & 2026 & - & 20 & 10812 & RGB, D & X2Robot, Quanta X1            & –
  & \tagG\tspace\tagLH \\

\bottomrule
\end{tabular}
\end{threeparttable}
\end{adjustbox}
\end{table}

\subsubsection{Mobile Manipulation Benchmarks}
\label{sec:bench_mobile}
Mobile manipulation jointly considers navigation and manipulation capabilities, requiring
action policies to have perception, planning, and dynamic execution ability across scenes,
posing higher comprehensive challenges to policy performance.

\textbf{ManipulaTHOR}~\cite{ehsani2021manipulathor} uses Kinova Gen3 as the robotic arm
within the AI2-THOR environment, evaluating visually guided manipulation and scene
generalization on a mobile platform across 28 tasks.
\textbf{HomeRobot}~\cite{yenamandra2023homerobot} uses Hello Robot Stretch as the platform
and builds open-vocabulary mobile manipulation evaluation in household scenarios in AI
Habitat, focusing on navigation and grasping coordination under arbitrary objects and
arbitrary scenes, with dual evaluation capability in both simulation and real robots.
\textbf{BEHAVIOR-1K}~\cite{li2023behavior1k} covers 1,000 daily living tasks and explores
Sim-to-Real transfer in OmniGibson, supporting physical simulation of rigid bodies,
deformable objects, and liquids, making it one of the broadest and most realistic benchmarks
in the mobile manipulation domain to date in terms of task coverage.

\subsubsection{Contact and Deformation Manipulation Benchmarks}
\label{sec:bench_contact}
Contact and deformation manipulation breaks from the traditional rigid-body assumption, requiring Action Policies to perceive and precisely control the deformation process of compliant objects. It represents one of the most challenging directions in manipulation policy evaluation, placing the most stringent demands on physical modeling. Based on the physical characteristics of the manipulated objects and the sensory modalities employed, existing benchmarks in this category can be divided into two groups.

The first group focuses on macroscopic deformable object manipulation, where the central challenge lies in physically modeling and visually guiding the control of objects that undergo large-scale deformation, such as cloth, liquids, and plastically deformable materials. \textbf{SoftGym}~\cite{lin2021softgym} and \textbf{PlasticineLab}~\cite{huang2021plasticinelab}, built respectively on the NVIDIA FleX and Taichi physics engines, construct evaluation environments for tasks such as cloth folding, liquid pouring, and plasticine shaping, assessing dexterous control capability using purely visual inputs. Building on this, \textbf{DaXBench}~\cite{chen2023daxbench} provides 9 deformable object manipulation tasks within the DaX framework, further broadening the range of material types and deformation modes available for evaluation.

The second group shifts toward contact-aware fine manipulation, where the core challenge is no longer the large-scale deformation of objects, but rather the real-time perception and feedback control of complex contact states — precisely the blind spot of purely vision-based approaches. \textbf{TacSL}~\cite{akinola2024tacsl} and \textbf{ManiFeel}~\cite{xu2025manifeel} formally introduce tactile sensing as an evaluation modality, constructing tactile-driven manipulation benchmarks on Isaac Gym and Isaac Sim/TacSL respectively, with Sim-to-Real transfer of tactile signals as a central evaluation objective.This shift signals a broader paradigm transition in the field — from vision-dominated perception toward vision-tactile fusion

\subsubsection{Real-Robot Benchmarks}
\label{sec:bench_real}
Simulation benchmarks provide efficient and reproducible evaluation conditions, but the
inherent gap between simulation and real environments makes it difficult to accurately
predict policy performance during real-world deployment. For this reason, some works
directly build evaluation benchmarks on real robot platforms to provide evaluation
conclusions with greater practical significance.

\textbf{RoboArena}~\cite{atreya2025roboarena} designs open evaluation scenarios for real
devices, focusing on policy generalization ability in real environments.
\textbf{RoboChallenge}~\cite{yakefu2025robochallenge} covers 30 multi-task real
manipulation scenarios and supports horizontal comparison evaluation on 4 robot platforms.
\textbf{Maniparena}~\cite{sun2026maniparena} uses the X2Robot and Quanta X1 dual platforms
as carriers, collects 10,812 real robot manipulation trajectories, and covers three
evaluation dimensions of generalization, long-horizon tasks, and multi-task, making it
currently the largest real-device manipulation benchmark in terms of scale.

\section{Open challenges and Opportunities}
\label{sec:open}

\newcommand{\highlightpara}[1]{\textbf{\textcolor{openmossblue}{#1}}}

While the emergence of World Action Models (WAMs) marks a pivotal shift from token-level prediction to state-level prediction, the path toward general physical intelligence remains fragmented. We argue that addressing these issues requires more than incremental scaling; it necessitates a fundamental rethinking of how world models should be architectured, grounded, and verified. Consequently, we outline several challenges and opportunities that we believe will define the next phase of WAM research.

\highlightpara{Architectural Coupling.} The field has produced a remarkable diversity of structural strategies for coupling world prediction and action generation---cascaded pipelines, joint diffusion backbones, discrete tokenization schemes, and implicit representation alignment---yet no systematic, controlled study has evaluated these paradigms against one another under matched conditions of scale, data, and evaluation protocol. It remains empirically unclear whether explicit visual prediction is strictly necessary for physical grounding, when cascaded and joint architectures differ in ways that matter for downstream control, and what inductive biases each coupling mechanism actually provides. Rigorous ablation studies and theoretical analysis of the design space are urgently needed to move the field beyond architectural fashion toward principled design choices.

One potentially productive direction lies in questioning whether explicit pixel-space prediction is a necessary component of the coupling altogether. Recent evidence suggests that the primary utility of world modeling in certain WAMs may stem from the auxiliary gradients provided during training rather than the explicit generation of future frames during inference; accordingly, some frameworks demonstrate that removing the future prediction head at test time does not necessarily degrade downstream control performance. This observation opens the door to a more computationally efficient paradigm: instead of reconstructing high-dimensional future observations in pixel space, models could predict abstract representations of future states in a jointly learned latent space. Such a \textit{latent-predictive} approach—exemplified by, but not limited to, JEPA could allow the model to sidestep the pixel-prediction bottleneck by focusing on the causal invariants of the environment while ignoring perceptually irrelevant details. By shifting the objective from pixel-level reconstruction to action-conditioned latent state transitions, WAMs may achieve a more compact and physically grounded coupling between world understanding and action execution, though the extent to which these latent-only representations can maintain the same grounding quality as generative approaches remains a vital open question.

\highlightpara{Multimodal Physical State Representation.} Existing WAMs almost universally predict future world states in the RGB visual modality. However, much of the physical information most critical for contact-rich manipulation is largely invisible in pixel space, including tactile distributions, contact forces, acoustic signatures, and material compliance. A world model confined to visual prediction thus has a systematic blind spot precisely at the physical interactions it most needs to model. Extending WAMs to jointly predict and reason over tactile, force, and proprioceptive future states, while developing the architectures and datasets necessary to support such multimodal world modeling, constitutes an important and largely unexplored frontier.

More broadly, this challenge suggests a shift in how the
\textit{world state} itself is defined within WAMs.
If the state relevant to contact-rich manipulation is not a pixel array but a joint distribution over visual, tactile, and force modalities, then multimodal prediction is not a mere extension of the current paradigm. It demands a reconceptualization of what a world model predicts and for what purpose. A valuable capability for future architectures is, therefore, \textit{modality-adaptive prediction}: the capacity to generate physically grounded predictions when rich sensor streams are available, while degrading gracefully to visual-only inference when they are not, rather than treating full multimodality as a fixed architectural requirement.

\highlightpara{Data Utilization and Mixture Design.} Several works have demonstrated that incorporating egocentric human video and other non-robot data sources can substantially improve downstream manipulation performance, but the principles governing optimal data mixture design remain poorly understood. What is the marginal contribution of each data source as a function of its scale and domain gap? Are the benefits of human video pretraining primarily semantic or dynamical? How should training curricula transition from broad internet-scale priors to precise action-labeled robot demonstrations? Developing principled recipes to these questions is essential for scaling WAM training beyond the current reliance on expensive robot-centric teleoperation data.

We argue that the central challenge in data mixture design lies in disentangling the multiple, potentially overlapping roles that non-robot data plays in grounding. Beyond mere semantic augmentation, we can conceptualize the value of human video through a hierarchy of transferrable knowledge: (1) \textit{low-level physical priors}, such as object permanence and gravitational constraints; (2) \textit{mid-level causal dynamics}, which encode the causal relationship between specific interactions and their physical outcomes; and (3) \textit{high-level task logic}, which encode task-relevant temporal dependencies independent of specific embodiments. Shifting the focus from empirical hyperparameter tuning to a more principled information-theoretic perspective on data mixture could allow the field to identify which specific components of world modeling are best learned from internet-scale video versus precision robot demonstrations. A key research frontier is thus the development of embodiment-aware filtering mechanisms—architectures that can selectively distill universal physical laws from diverse sources while isolating and suppressing behaviors that are kinematically incompatible with the target robot.

\highlightpara{Long-Horizon Planning and Temporal Abstraction.}
WAMs are most commonly evaluated on short-horizon manipulation tasks within single interaction contexts, but genuine embodied generalism demands sustained reasoning over extended task horizons. Current architectures face compounding challenges in this regime: distributional drift in world model predictions accumulates over long rollouts, action errors compound in the absence of corrective replanning, and representing full long-horizon task trajectories as continuous generative outputs is both computationally prohibitive and architecturally unsupported. A principled framework for hierarchical world-action modeling---connecting high-level semantic task decomposition to low-level physical prediction within a unified, learnable architecture---remains a critical open challenge.

We envision three complementary pathways toward long-horizon reasoning. First is the \textit{modular hierarchy}, where WAMs function as low-level physical executors guided by high-level planners (e.g., Vision-Language Models) that decompose complex missions into semantic subgoals. Second is the \textit{intrinsic hierarchical WAM}, an architecture capable of generating multi-resolution future predictions—predicting coarse-grained state transitions for strategic planning while simultaneously synthesizing fine-grained physical details for reactive control. Third is the \textit{scaling of temporal context}, which involves architectural innovations to expand the "memory" of WAMs, allowing them to maintain rich historical state information without the quadratic overhead of standard attention. Whether these paths converge into a unified, end-to-end learnable hierarchy or remain as a composite system of specialized modules is a foundational question for the next generation of WAM research.

\highlightpara{Inference Latency and Computational Efficiency.}
The integration of world prediction imposes a severe ``latency tax'' that threatens the viability of closed-loop control. While early cascaded pipelines suffered from the cumulative delay of independent modules, modern joint WAMs still struggle to meet the millisecond-level response frequencies required for high-fidelity manipulation. DreamZero pushed WAM inference toward the 7Hz through algorithmic acceleration and low-level CUDA optimizations; however, this still lags significantly behind the 50Hz standards of contemporary non-generative VLA policies. The fundamental tension remains:  Can the rich physical foresight of a WAM be preserved at the temporal resolution of real-time motor control?

Underlying this challenge is a theoretical question that has received little direct study: how much predictive fidelity does downstream control actually need? If performance gains level off well before reaching the quality of full diffusion-based synthesis, then the goal should not be to make high-fidelity prediction faster. Instead, it should be to identify the minimal sufficient world model for a given task and act accordingly. This points toward task-adaptive predictive fidelity, where the model dynamically adjusts the depth and resolution of its predictions based on the task's demands and error tolerance， investing computation where precision matters most, and using coarser approximations elsewhere.

\highlightpara{Evaluation Methodology.} Current WAM evaluation suffers from a severe decoupling. World modeling is typically assessed via pixel-level metrics (e.g., PSNR, FVD) that capture visual plausibility but ignore physical correctness---allowing videos where unsupported objects hover or fluids defy gravity to score highly. Conversely, action generation is measured solely by downstream task success. This disjointed paradigm ignores the core premise of WAMs. A critical missing piece in the WAM ecosystem is the lack of joint evaluation metrics that quantify the causal consistency between imagined futures and generated actions. To bridge this gap, future benchmarks must introduce coupled metrics that probe the causal link between visual prediction and physical execution. For instance, \textit{Counterfactual Consistency} could quantify how actions adapt to perturbations in the imagined future, while \textit{Foresight-Conditioned Success} ensures executed trajectories strictly adhere to the generated visual plan rather than relying on spurious correlations. Evaluating the ``intent alignment'' between what the model imagines and what it physically executes is essential for ensuring that actions are genuinely grounded in accurate visual foresight rather than memorized dataset biases.

Beyond individual metrics, the community may benefit from a shared benchmarking framework designed around the core premise of WAMs: one in which world prediction quality and action quality are evaluated jointly and in relation to each other, rather than through separate leaderboards. Such a framework, analogous in ambition to what standard benchmarks have provided for visual recognition or language understanding, could help establish community consensus around what it means for a WAM to be ``working'' in the intended sense—not merely predicting plausible futures, nor merely succeeding at tasks, but generating actions that are demonstrably causally grounded in accurate physical foresight.

\highlightpara{Safety and Reliable Physical Deployment.} WAMs deployed in physical environments introduce safety considerations that go beyond the bounded reactive failures of conventional VLA policies. A model that confidently imagines an incorrect physical future may commit to extended action sequences whose real-world consequences are difficult to interrupt or recover from, potentially causing harm to objects, environments, or people. At the same time, the predictive capacity of WAMs offers a principled opportunity for safety enforcement: world predictions can, in principle, be checked against physical constraints or conservative uncertainty estimates before predicted actions are executed. Realizing this opportunity requires integrating safety verification into the inference pipeline in ways that are computationally tractable and robust to distributional shift, which remains a significant open challenge as WAM capabilities and deployment ambitions continue to expand.

More broadly, the safety challenge highlights an underexplored duality in the WAM paradigm: the same predictive capacity that makes these models potentially more capable than reactive policies also makes their failure modes potentially more consequential. Turning this duality into an advantage means using world predictions not only to guide actions but to verify them before execution. This points toward \textit{prediction-integrated safety}, where uncertainty estimates over imagined futures are treated as first-class inputs to a safety monitor rather than as incidental byproducts. Whether such monitors can be made both computationally tractable and sufficiently robust to distributional shift to be meaningful in open-world deployments remains to be established.

\section{Conclusions}
This survey provides the first systematic and critical analysis of the World Action Model (WAM) landscape, positioning it as a pivotal frontier in embodied AI. We establish a clear conceptual framework by formally defining WAMs, disambiguating them from related methodologies, and categorizing their diverse architectures into Cascaded and Joint paradigms. Within these paradigms, we systematically examine their generation modalities, conditioning mechanisms, and action decoding strategies. To support ongoing research, we further synthesize the efforts in scalable training datasets and summarize the emerging multidimensional evaluation protocols spanning visual fidelity, physical commonsense, and action plausibility. Finally, grounded in the current state of development, we outline critical open challenges and future trajectories to guide the next phase of progress. As generative world modeling and robotics continue to converge, WAM research holds immense promise. We hope this survey clarifies the field’s terminological boundaries, maps out its architectural design space, and contributes meaningfully to the pursuit of generalist embodied agents.

\clearpage
\bibliographystyle{unsrtnat} 
\bibliography{main}



\end{document}